\documentclass{article}
\usepackage{fancyhdr}

    \usepackage{neurips_2024}

\usepackage{soul}
\usepackage[utf8]{inputenc} 
\usepackage[T1]{fontenc}    
\usepackage{hyperref}       
\usepackage{url}            
\usepackage{booktabs}       
\usepackage{amsfonts}       
\usepackage{fancyhdr}       

\usepackage{nicefrac}       
\usepackage{microtype}      
\usepackage{graphicx}
\usepackage{pifont}
\usepackage{multirow}
\usepackage{CJKutf8}
\usepackage{makecell}
\definecolor{darkmagenta}{rgb}{0.56, 0.0, 1.0}
\definecolor{softyellow}{rgb}{1.0, 0.92, 0.3} 
\definecolor{LightAquamarine}{rgb}{0.75, 1.0, 0.8} 
\definecolor{FireBrick}{RGB}{178,34,34}
\definecolor{MediumPurple}{RGB}{147,112,219}

\definecolor{uclablue}{rgb}{0.15, 0.45, 0.68}
\hypersetup{
    breaklinks,
    colorlinks=true,
    citecolor={darkmagenta},
    linkcolor={uclablue},
    urlcolor={uclablue}
}
\usepackage{wrapfig}
\usepackage{float}
\usepackage{subcaption}
\usepackage{placeins}
\usepackage{lipsum} 
\usepackage{tcolorbox}
\usepackage{amsmath}
\usepackage{amssymb}
\usepackage{utfsym}
\usepackage{fontawesome}
\usepackage{xspace}
\usepackage{enumitem}
\usepackage{multirow} 
\tcbuselibrary{breakable}
\usepackage{enumitem}
\usepackage{colortbl}
\usepackage{fancyhdr}

\usepackage{tcolorbox}
\usepackage{transparent}
\usepackage{tabularx}

\newtcolorbox{abstractbox}{
    colback=blue!5!white,     
    frame empty,   
    boxrule=1pt,              
    arc=4mm,                  
    left=8pt,                 
    right=8pt,                
    top=8pt,                  
    bottom=8pt,                
    opacityback=0.9
}

\newcommand{\cosim}[2]{\operatorname{cos}\!\left(#1,#2\right)}

\title{%
  \texorpdfstring{%
    \raisebox{-0.15em}{%
      \includegraphics[height=1.3em]{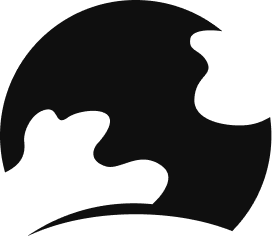}%
    }\hspace{0.4em}%
    HappyWorld-Bench%
  }{HappyWorld-Bench}%
}

\vspace{-0.5cm}

\author{
Zhiqi Bai$^{1}$, Junai Cai$^{1}$, Yixin Chen$^{2}$, Jingrun Du$^{1,3}$, Tao Feng$^{1}$, Wei Gong$^{1}$, Siyuan Huang$^{2}$, Xiao Lin$^{1}$, Jiaheng Liu$^{4}$, Jun Luo$^{2,5}$, Yongzhe Lyu$^{2,5}$, Liya Ma$^{1}$, Zenan Meng$^{1}$, Lin Qu$^{1}$, Wenbo Su$^{1}$, Jiaming Wang$^{4}$, Qinghe Wang$^{1}$, Shaofei Wang$^{2}$, Yanghai Wang$^{4}$, Zequn Wang$^{1}$, Ziming Wang$^{1}$, Hu Wei$^{1}$, Jiangtao Wu$^{4}$, Ruiqi Wu$^{1}$, Jiaxin Xie$^{1}$, Yuchi Xu$^{1}$, Ze Xu$^{1}$, Chengting Yu$^{1}$, Liangyu Yuan$^{1}$, Gang Zeng$^{5}$, Yawen Zeng$^{1}$, Xingyao Zhang$^{1}$, Zizheng Zhang$^{1}$, Bo Zheng$^{1}$, Jiancheng Zhu$^{1}$, Song-Chun Zhu$^{2,5}$ \\[6pt]
\textit{(Authors are listed in alphabetical order)}
}

\begin{document}

\maketitle
\let\oldthefootnote\thefootnote

\footnotetext[1]{Alibaba Token Hub, Alibaba Group}
\footnotetext[2]{State Key Laboratory of General Artificial Intelligence, BIGAI}
\footnotetext[3]{Tsinghua University}
\footnotetext[4]{Nanjing University}
\footnotetext[5]{State Key Laboratory of General Artificial Intelligence, Peking University}

\begin{abstractbox}
\begin{center}
\textbf{\Large Abstract}
\end{center}

Evaluating world models requires assessing both the quality of the worlds they generate and their consistency and responsiveness under exploration, interaction, and modification. We introduce \textbf{HappyWorld-Bench}, a comprehensive benchmark that evaluates whether generated worlds remain reliable as agents interact with them. Our design is built on a hierarchical capability framework of six world capabilities (W1–W6)—from generative construction to unified world modeling—instantiated across three independent evaluation tracks: video world models, spatial world models, and embodied world models. 
HappyWorld-Bench comprises 1,138 video prompts, 300 spatial scenes, and 254 embodied test cases. Across all three tracks, we build and operate HappyWorld-Arena to organize human A/B comparisons and derive model-level Elo ratings, which complement newly designed automated metrics that capture behavioral correctness. We evaluate 14 video world models, 9 spatial systems, and 8 embodied candidates under this unified framework. Results reveal remaining reliability gaps across all three tracks: video models exhibit reduced consistency during extended rollouts and revisits, spatial models achieve at best 70.14\% placement accuracy and 73.33\% edit execution, and embodied models struggle to preserve state across multi-step actions and respond precisely to altered action conditions and physical rules. These findings highlight the need to evaluate world models not only by visual quality, but also by state consistency and the correctness of their responses to actions and interventions.

\end{abstractbox}






\section{Introduction}

A world model is an internal representation that an agent uses to predict how its environment will respond to actions~\citep{ha2018worldmodels,lecun2022path}. In cognitive science, internal models and cognitive maps have long been associated with prediction, planning, and flexible behavior~\citep{tolman1948cognitive}. In artificial intelligence, world models have recently attracted renewed attention as a means for agents to \textit{imagine} future outcomes before acting~\citep{ha2018worldmodels,hafner2023dreamerv3}.

The premise is straightforward: if an agent can internally simulate the future, it can choose actions that lead to desirable states without trial-and-error in the real world. This premise rests on a critical requirement---the simulated world must be \emph{reliable}. A single photorealistic frame is insufficient; the world must remain geometrically consistent when the camera moves, physically plausible under interaction, temporally persistent over extended rollouts and revisits, and responsive to control and intervention~\citep{wang2026matrixgame30realtimestreaming,hong2025relicinteractivevideoworld,zhu2026sanawmefficientminutescaleworld,genie3,lingbotworld,alibabatokenhub2026happyoyster}. In short, a world model is useful only to the extent that its predictions \emph{hold up under interaction}.

Despite the centrality of this requirement, the evaluation of world models remains fragmented. Existing benchmarks have begun to evaluate controllability, physical plausibility, memory, interaction, and long-horizon consistency across different world-model settings~\citep{zheng2025vbench20advancingvideogeneration,duan2025worldscoreunifiedevaluationbenchmark,xu2026worldmarkunifiedbenchmarksuite,fang2026iworldbenchbenchmarkinteractiveworld,ying2026wbenchcomprehensivemultiturnbenchmark,zhang2026mbenchcomprehensivebenchmarkmemory,xu2026worldroambenchopenworldbenchmarklonghorizon,ding2026playworldbenchmarkingworldmodels}, but these capabilities are typically studied separately across model families and evaluation settings. Embodied world-model benchmarks increasingly evaluate action-conditioned visual behavior and downstream utility, but their evaluation interfaces and objectives remain distinct from those used for video and spatial world models~\citep{em_yue2025ewmbench,em_li2025worldmodelbench,em_yang2026mirabench,em_li2026robotrustbench,em_shang2026worldarena,em_jiang2026robowmbench}. Each community has developed sophisticated evaluation protocols, but existing benchmarks remain specialized to particular model forms or subsets of world capabilities~\citep{duan2025worldscoreunifiedevaluationbenchmark,xu2026worldmarkunifiedbenchmarksuite,ying2026wbenchcomprehensivemultiturnbenchmark,ding2026playworldbenchmarkingworldmodels,duggal2025eval3d,tam2025sceneeval}, making it difficult to evaluate them under a shared capability framework.

This fragmentation has concrete consequences. A video model that produces visually compelling sequences may nonetheless fail to preserve object identity or world state over extended interaction; a spatial model that produces visually complete scenes may still lack navigable or physically usable geometry; and an embodied world model that generates plausible action-conditioned videos may fail to preserve state or produce the intended action consequences~\citep{zhang2026mbenchcomprehensivebenchmarkmemory,xu2026worldroambenchopenworldbenchmarklonghorizon,ding2026playworldbenchmarkingworldmodels,tam2025sceneeval,duan2025worldscoreunifiedevaluationbenchmark}. Without a unified benchmark that systematically probes \emph{how a world behaves under action, memory, and intervention} across different model forms, it remains difficult to determine whether current systems support reliable world modeling beyond surface-level generation~\citep{duan2025worldscoreunifiedevaluationbenchmark,xu2026worldmarkunifiedbenchmarksuite,ying2026wbenchcomprehensivemultiturnbenchmark,xu2026worldroambenchopenworldbenchmarklonghorizon,ding2026playworldbenchmarkingworldmodels}.
\begin{figure}[h]
    \centering
    \includegraphics[width=1\linewidth]{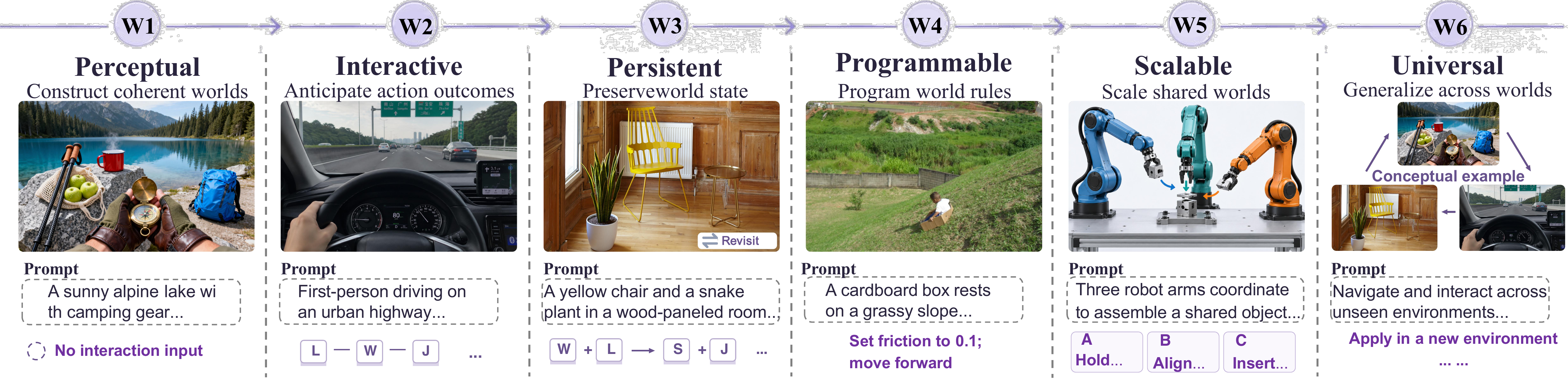}
    \caption{World levels of HappyWorld-Bench.}
\end{figure}

\paragraph{Contributions.}
\begin{table*}[tb!]
  \centering
  \small
  \setlength{\tabcolsep}{6pt}
  \renewcommand{\arraystretch}{1.15}

  \caption{
  World level W1--W6 capability hierarchy and definition.
  }
  \label{tab:w-hierarchy}

  \begin{tabularx}{\textwidth}{
    @{}
    p{0.24\textwidth}
    >{\raggedright\arraybackslash}X
    >{\raggedright\arraybackslash}X
    @{}
  }
  \toprule
  \textbf{Level}
  & \textbf{Scope Defintion} \\
  \midrule

  \textbf{W1: Perceptual World}
  & Construct a coherent world representation from visual or multimodal
    conditions, with correct semantics, spatial structure, and short-term
    temporal continuity.
  \\

  \addlinespace[2pt]

  \textbf{W2: Interactive World}
  & Simulate action-conditioned state transitions, predicting how agents,
    objects, and environments evolve in response to interaction while
    preserving local geometric, physical, and causal consistency.
  \\

  \addlinespace[2pt]

  \textbf{W3: Persistent World}
  & Maintain global spatial structure, object identity, and accumulated world
    states over long-horizon interaction, viewpoint changes, occlusion,
    and revisitation.
  \\

  \addlinespace[2pt]

  \textbf{W4: Programmable World}
  & Support explicit interventions on objects, events, behaviors, or
    world rules, with intended changes propagated causally while
    unaffected content remains consistent.
  \\

  \addlinespace[2pt]

  \textbf{W5: Scalable World}
  & 
  Generate infinitely extensible world states shared by multiple embodied or virtual agents, supporting agent communication, synchronization, cooperation, and conflict handling under partial observability.
  \\

  \addlinespace[2pt]

  \textbf{W6: Universal World}
  & Integrate generation, simulation, persistent state modeling,
    interaction, and planning into a unified system that fully replicates the real world and generalizes
    across environments, tasks, modalities, and embodiments.
  \\

  \bottomrule
  \end{tabularx}

  \vspace{-8pt}
\end{table*}

We introduce \textbf{HappyWorld-Bench}, a comprehensive benchmark that evaluates whether generated worlds remain reliable under generation, exploration, interaction, and intervention. Our design is organized around three principles.


\textbf{First}, we define a \emph{hierarchical capability framework} comprising six levels of world modeling (W1--W6). \textbf{Perceptual World} (W1) provides the perceptual foundation, organizing visual or multimodal inputs into semantically accurate and spatially coherent world representations with temporal continuity over short horizons. Building on this foundation, \textbf{Interactive World} (W2) advances to action-driven simulation, capturing how agents, objects, and environments respond to interactions in ways that respect local geometry, physics, and causal relationships. \textbf{Persistent World} (W3) extends these capabilities across longer interaction horizons, retaining global spatial organization, stable object identities, and accumulated state information as viewpoints shift, entities become occluded, and previously observed regions are revisited. \textbf{Programmable World} (W4) adds explicit control over objects, events, behaviors, and world rules, allowing targeted modifications whose consequences unfold causally without disrupting content outside their scope of influence. \textbf{Scalable World} (W5) broadens world generation to unbounded, shared environments in which multiple embodied or virtual agents communicate, synchronize, cooperate, and manage conflicts despite having only partial observations. Finally, \textbf{Universal World} (W6) brings generation, simulation, persistent state modeling, interaction, and planning together within a unified system, with the ultimate goal of fully replicating the real world and generalizing across diverse environments, tasks, modalities, and embodiments. Together, these levels delineate an increasingly comprehensive set of capabilities, progressing from perceptual coherence and responsive dynamics to enduring state, causal control, multi-agent scalability, and universal world modeling.

\textbf{Second}, we operationalize this shared capability framework across \emph{three complementary evaluation tracks} that correspond to the three fundamental functions of a world model: a \textbf{Video World Model} track that tests whether observations beyond the input are spatially and temporally coherent; a \textbf{Spatial World Model} track that tests whether exported scenes support valid physical operations; and an \textbf{Embodied World Model} track that tests the ability to predict, from an egocentric viewpoint, how the robot, its surrounding scene, and the manipulated objects evolve in response
to robot actions.
Crucially, all three tracks share the same W1--W6 capability taxonomy, while operationalizing different subsets of the hierarchy according to their model form and evaluation interface.


\begin{figure}
    \centering
        \includegraphics[width=1\linewidth]{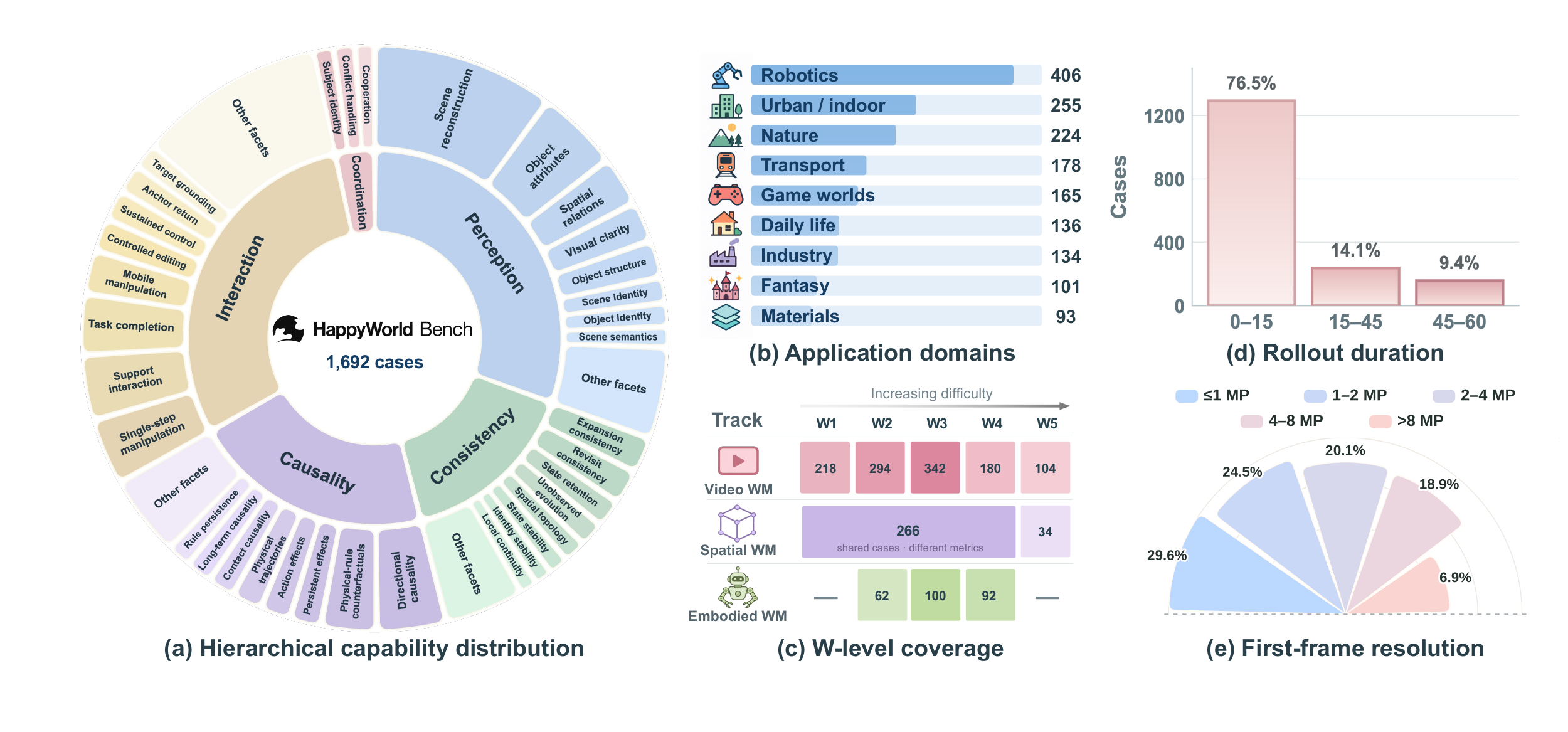}
    \caption{Data statistics of HappyWorld-Bench.}
    \label{fig:dataset-statistics}
\end{figure}

\textbf{Third}, we combine \emph{large-scale human evaluation} with \emph{newly designed automated metrics} to capture both
subjective quality and objective behavioral correctness. 
Across all three tracks, we build and operate \textbf{HappyWorld-Arena} to organize human A/B comparisons of model outputs and derive model-level Elo ratings within each track.
These ratings summarize overall human preference, while the automated metrics provide fine-grained assessments of specific world capabilities.
For the video track, we curate 1{,}138 prompts spanning capabilities
W1--W5 and evaluate perception, consistency, causality, and controllable interaction using automated metrics.
For the spatial track, we curate 300 scenes: 266 for evaluating visual quality, physical usability, consistency, and editing, and 34 for evaluating spatial expansion.
For the embodied track, we design 254 formal test cases spanning atomic
actions, multi-stage sequences, and  action pairs. 
Together, these tracks comprise \textbf{1{,}692 track-specific
instances}. Figure~\ref{fig:dataset-statistics} summarizes their hierarchical capability distribution, application domains, W-level coverage, duration distribution, and first-frame resolutions. 
The data span nine application domains, ranging from robotics, urban and indoor environments, and nature to transport, game worlds, daily life, industry, fantasy, and materials. Rollouts range from short interactions to 60-second sequences, while first-frame resolutions span from below 1~MP to above 8~MP, providing heterogeneous visual and temporal conditions for evaluation.

We evaluate 14 video world models, 9 spatial systems, and 8 embodied candidates under this unified protocol. Our results reveal a sobering landscape. 
In the video track, the two leading systems achieve Arena Elo ratings of 1263 and 1206, reflecting their relative standing in overall human preference. Capability-level evaluations reveal that even the strongest systems have limitations in maintaining consistent world states and producing the intended responses to actions and interventions.
In the spatial track, 
we found that existing spatial world models perform poorly in physical plausibility, editability, and scene expansion, highlighting remaining challenges in these aspects.
In the embodied track, models struggle to preserve state across multi-step actions and respond precisely to altered action conditions and physical rules.
These findings collectively suggest that current world models, despite impressive generative quality, still face substantial challenges in maintaining \emph{reliable state under repeated change}.

Our core contributions are selected as:
\definecolor{blueviolet}{RGB}{138,43,226}
\newtcolorbox{insightblock}{
  colback=blueviolet!5,   
  colframe=blueviolet!50!black!50!,    
  boxrule=0.5mm,       
  arc=2mm,             
  left=0pt,            
  right=8pt,           
  top=8pt,             
  bottom=8pt,          
}

\begin{insightblock}
\begin{enumerate}[leftmargin=1.5em]
    \item A \textbf{hierarchical capability framework} (W1--W6) that formalizes what it means for a generated world to be reliable, providing a common vocabulary for evaluating world models across video generation, 3D reconstruction, and embodied control.
    
    

\item \textbf{HappyWorld-Bench}, a large-scale benchmark comprising 1{,}138 video prompts, 300 spatial scenes, and 254 embodied test cases, with a dedicated suite of automated metrics for each of the three domains to support systematic and reproducible evaluation of world modeling capabilities.

\item \textbf{HappyWorld Arena}, an arena platform for comparative assessment of world models, providing a shared evaluation resource to foster community participation, facilitate the exchange of evaluation results, and support continued progress in world modeling.
    
\item A \textbf{comprehensive empirical study} of 14 video models, 9 spatial systems, and 8 embodied candidates, revealing that current world models exhibit significant gaps in interactive simulation, state persistence, and programmable dynamics, and identifying concrete directions for future research toward reliable world understanding.

\end{enumerate}
\end{insightblock}

\section{Related Work}
\label{sec:related-work}
\subsection{World Model}
\paragraph{Video world models}
Interactive video world models extend video generation from passive synthesis to controllable environments that evolve in response to user actions. Early studies explored learning latent actions from unlabeled videos and autoregressively simulating interactive game environments~\citep{genie,gamengen}. Subsequent systems introduced explicit control through keyboard and mouse inputs, camera trajectories, and history conditioning, while improving inference efficiency toward real-time and streaming interaction~\citep{matrixgame2,hunyuangamecraft,Mao_2026_CVPR}. Recent models further improve interaction duration and world persistence through long-context modeling and memory mechanisms: Matrix-Game 3.0 and RELIC explicitly maintain long-horizon visual history, while SANA-WM targets efficient minute-scale generation with precise camera control~\citep{wang2026matrixgame30realtimestreaming,hong2025relicinteractivevideoworld,zhu2026sanawmefficientminutescaleworld}. Genie 3, LingBot-World, and other recent systems further scale interactive generation across diverse environments and extended rollouts~\citep{genie3,lingbotworld}, while models such as HappyOyster and Echo-WM broaden the interaction space to scene manipulation, character and camera control, continuous motion control, and multimodal audio--visual generation~\citep{alibabatokenhub2026happyoyster,zhang2026echowmopenenterableomnimodal}. Overall, video world models are progressing toward increasingly responsive, persistent, and general interactive environments.

\paragraph{Spatial world models}
Spatial world models construct explicit, renderable environments from images or scene descriptions, extending scene content beyond observed views. Early approaches, including WonderJourney and WonderWorld, combine image synthesis, depth estimation, and incremental 3D construction to generate coherently connected scenes \citep{yu2023wonderjourney,yu2024wonderworld}. Panoramic representations broaden spatial coverage: WorldGen lifts panoramas into 3D environments \citep{worldgen2025ziyangxie}, HunyuanWorld 1.0 introduces semantic decomposition and layered mesh reconstruction \citep{hunyuanworld2025}, and Matrix-3D combines trajectory-conditioned panoramic video generation with feed-forward or optimization-based reconstruction \citep{yang2025matrix3d}. HY-World 2.0 further integrates panorama generation, camera-path planning, memory-conditioned view expansion, and 3D reconstruction \citep{hyworld2026}. A complementary direction transfers video generative priors into explicit geometry. Lyra trains a 3D Gaussian Splatting (3DGS) decoder through self-distillation from a video diffusion model \citep{bahmani2025lyra}, while Lyra 2.0 combines geometry-guided retrieval of historical observations with training on self-augmented histories to improve persistence during extended exploration \citep{shen2026lyra20explorablegenerative}. FlashWorld instead emphasizes efficiency, using cross-mode distillation to enable direct 3DGS generation with few denoising steps \citep{li2025flashworld}. Beyond generation, Marble supports multimodal authoring, editing, expansion, and composition, with Gaussian-splat and mesh exports that include collision geometry \citep{worldlabs2025marble}; Code2Worlds explores programmatic construction of executable 4D scenes through simulation code refined using visual and motion feedback \citep{zhang2026code2worlds}. Our evaluated set additionally includes GPT-6-Astra, whose submitted scenes follow the same spatial-world evaluation protocol. Together, these developments motivate assessing not only initial visual quality, but also whether generated environments support navigation and contact, maintain consistency across viewpoints, and preserve existing content during editing and expansion.

\paragraph{Embodied world models}
Embodied world models predict how an agent's observation evolves under its own
action. Latent-space variants learn compact predictive states, as recurrent
state-space models for imagination-based policy
learning~\citep{em_hafner2023dreamerv3}, discrete-token
sequences~\citep{em_micheli2023iris,em_wu2024ivideogpt}, or predicted video
features with an action-conditioned planning
head~\citep{em_bardes2024vjepa,em_assran2025vjepa2}. Pixel-space variants
instead treat action-conditioned video generation as the dominant interface,
ranging from general real-world simulators~\citep{em_yang2023unisim} and
playable environments with latent actions recovered from unlabeled
video~\citep{genie} to real-time interactive
engines~\citep{gamengen,em_zhang2025matrixgame}.
In manipulation, video prediction is coupled directly to action generation
through video pre-training~\citep{em_wu2023gr1,em_cheang2024gr2}, generated
frames converted into plans, edits, or
correspondences~\citep{em_du2023unipi,em_black2023susie,em_ko2023avdc},
compositional and action-unified embodied
futures~\citep{em_zhou2024robodreamer,em_cen2025worldvla,em_chi2024eva,em_huang2025enerverse},
and physical priors injected into the prediction
process~\citep{em_shang2025roboscape,em_zhang2024physdreamer,em_jiang2025phystwin}.
These directions are being integrated into embodied foundation
platforms~\citep{em_nvidia2025cosmos,em_nvidia2025cosmosreason1,em_nvidia2025gr00tn1,em_jang2025dreamgen,em_liao2025genieenvisioner,em_qiu2026gesim2,em_agibot2026geact2,em_zhang2026qwenrobotworld,em_zou2026pelicansim},
and similar paradigms extend to
driving~\citep{em_hu2023gaia1,em_russell2025gaia2,em_wang2023drivedreamer,em_gao2024vista,em_zheng2023occworld}
and navigation~\citep{em_bar2024navigationwm}, supported by generative task
substrates and large-scale
demonstrations~\citep{em_wang2023robogen,em_agibot2025agibotworld}. Community
evaluations further emphasize action-conditioned
realism~\citep{em_mereu2025onexwmchallenge}, and surveys organize the field
along representation, supervision, and downstream
use~\citep{em_ding2024worldmodelsurvey,em_li2025embodiedwmsurvey,%
em_lu2026worldactionsurvey,em_yao2026embodiedwm}.

\subsection{World Model Benchmarks}

\paragraph{Video world model benchmarks}
Evaluation has evolved from assessing generated videos to examining whether models can sustain coherent and controllable interactive worlds. Conventional video-generation benchmarks primarily evaluate visual quality, motion quality, temporal consistency, and semantic alignment~\citep{Huang_2024_CVPR,Liu_2024_CVPR}. VBench 2.0 extends this scope toward intrinsic faithfulness, including physical plausibility and commonsense consistency, while WorldScore evaluates world generation in terms of controllability, visual and three-dimensional consistency, and dynamics under explicit camera trajectories~\citep{zheng2025vbench20advancingvideogeneration,duan2025worldscoreunifiedevaluationbenchmark}. More recent benchmarks directly target interactive world models. WorldMark establishes standardized scenes, action sequences, and control mappings for cross-model comparison~\citep{xu2026worldmarkunifiedbenchmarksuite}, while iWorld-Bench introduces a unified action-generation framework to evaluate visual generation, trajectory following, and memory~\citep{xu2026worldroambenchopenworldbenchmarklonghorizon}. WBench further introduces multi-turn interactions spanning navigation, subject actions, event editing, and perspective switching, evaluating video quality, setting and interaction adherence, consistency, and physical compliance~\citep{ying2026wbenchcomprehensivemultiturnbenchmark}. Beyond local controllability, recent benchmarks increasingly examine persistence over extended interaction. MBench focuses on entity, environment, and causal consistency as complementary aspects of memory, while WorldRoamBench evaluates action following, visual drift, interaction physics, and scene and subject memory under continuous interaction~\citep{zhang2026mbenchcomprehensivebenchmarkmemory,xu2026worldroambenchopenworldbenchmarklonghorizon}. PlayWorld further moves beyond fixed action sequences by introducing closed-loop agent interaction toward specified long-horizon objectives, evaluating geometry consistency, interaction fidelity, and state evolution both within and outside the current view~\citep{ding2026playworldbenchmarkingworldmodels}. Collectively, these efforts broaden world-model evaluation from perceptual quality and short-term controllability toward consistency, physics, memory, interaction, navigation, and sustained goal-directed behavior.


\paragraph{Spatial world model benchmarks}
Existing benchmarks assess complementary aspects of generated spatial worlds, ranging from asset quality to scene plausibility and coherent exploration. Eval3D uses foundation models and specialized tools as probes for fine-grained assessment of generated assets, including geometric and semantic consistency, text alignment, and visual quality \citep{duggal2025eval3d}. At the scene level, SceneEval measures compliance with specified object counts, attributes, and spatial relations, together with physical and functional plausibility through support, collision, and navigability checks \citep{tam2025sceneeval}. WorldScore formulates world generation as successive next-scene generation tasks under prescribed camera trajectories, evaluating controllability, quality, and dynamics across 3D, 4D, and video generation methods \citep{duan2025worldscoreunifiedevaluationbenchmark}. Building on these complementary perspectives, our spatial track combines rendered observations, exported geometry, and paired operations within a common protocol for explicit generated environments. It evaluates construction quality (W1), navigation and stable placement (W2), scene- and object-level consistency (W3), editing with preservation of non-target content (W4), and expansion with retention of existing regions and connecting paths (W5). This organization tests whether a generated environment remains usable and coherent as it is explored and modified.

\paragraph{Embodied world model benchmarks}
Evaluation of embodied world models is gradually extending from
appearance-oriented scoring towards action fidelity, physical plausibility,
and downstream utility, and existing protocols can be broadly grouped into
three directions along this shift. The first scores the generated video
itself~\citep{em_yue2025ewmbench,em_li2025worldmodelbench,em_qin2024worldsimbench,duan2025worldscoreunifiedevaluationbenchmark};
the second examines behavioral reliability under explicit action
conditioning~\citep{em_yang2026mirabench,em_li2026robotrustbench,em_rong2026h2rbench,em_chen2026harnessevalw,fang2026iworldbenchbenchmarkinteractiveworld};
and the third measures functional utility, treating the world model as a data
engine, a policy evaluator, or an in-model environment for policy
evaluation~\citep{em_shang2026worldarena,em_shang2026worldarena2,em_jiang2026robowmbench,em_quevedo2025worldgym}.
The object of evaluation thus moves from visual quality, to action-conditioned
behavior, to downstream utility.
Our embodied track addresses the first two of these objects, visual quality and
action-conditioned behavior, and keeps the interface purely generative: each
candidate is conditioned only on a single egocentric reference frame and an
action prompt, a requirement that any model generating an action-conditioned
rollout can meet, so no simulator, action decoder, or robot is needed. The
evidence is organized along a scoring axis of perception, consistency,
causality, and controllability, and a capability axis whose interaction
requirement rises across W2--W4: atomic action response at W2, state
persistence under ordered multi-step interaction at W3, and the response to an
edited action or physical condition at W4, measured through matched-pair
intervention from a shared initial state.

\section{Video World Model Track}

\subsection{Data Construction}
\label{sec:video-data}
\begin{figure}
    \centering
    \includegraphics[width=1.0\linewidth]{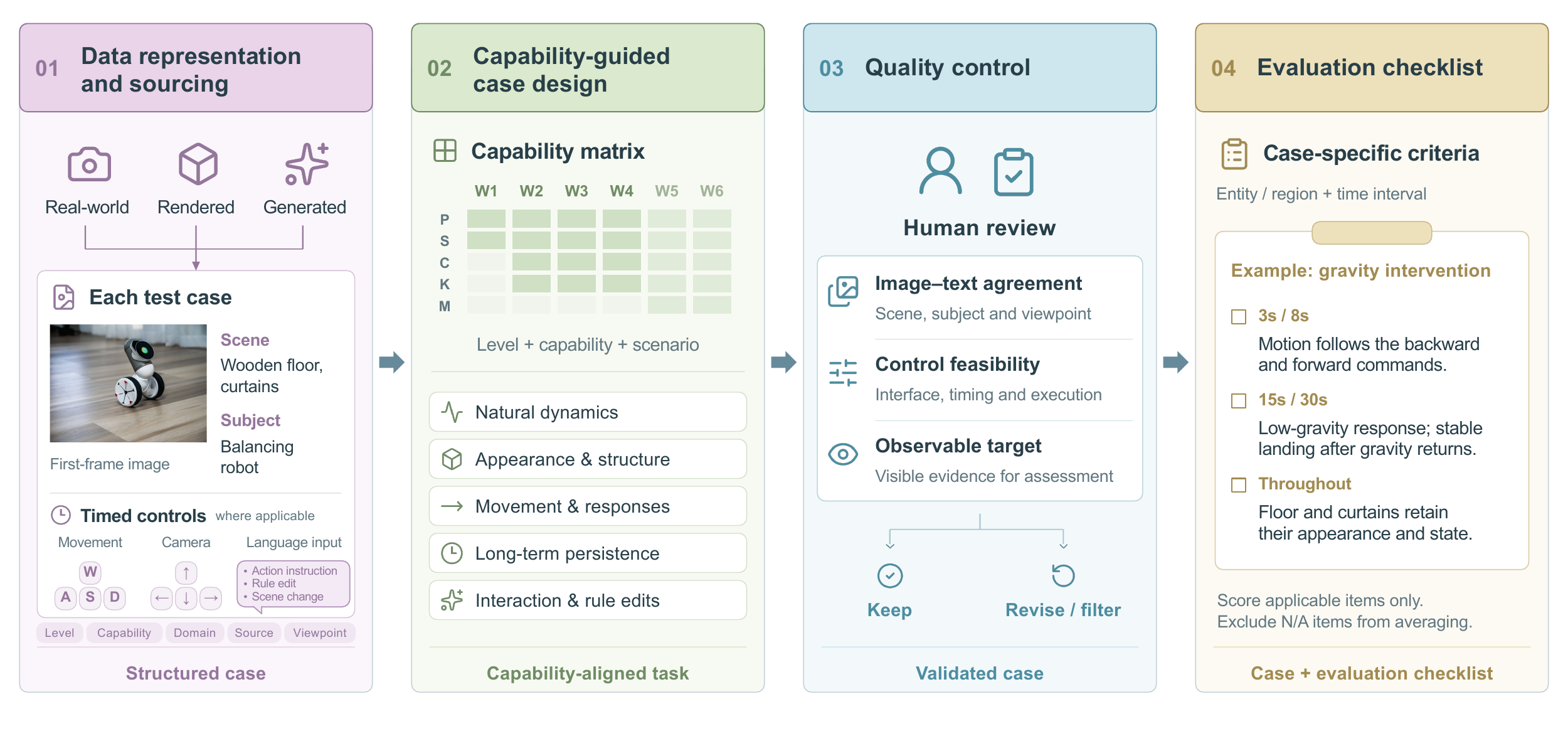}
    \caption{Data pipeline of HappyWorld-Bench.}
\end{figure}
\paragraph{Data representation and sourcing.} We construct structured test cases for video generation. Each case comprises a first-frame image and a textual description of the scene and any selected subject, together with a time-indexed control sequence where applicable.

Cases are annotated by target capability, W level, application domain, image source, and viewpoint. Depending on the W level, control sequences specify movement or camera rotation and may include textual instructions for subject actions, interactions, or environmental rule edits.

Task assignments follow a capability matrix that specifies the target W level and evaluation dimension, along with the application domain and viewpoint. First frames are collected from real-world imagery and rendered environments, with generated images used to supplement the dataset.

Control specifications are tailored to the target capability and W level. W1 cases require no actions, as they assess natural scene evolution and visual continuity without external control inputs.
For W2 and W3, action sequences specify control directions and time intervals rather than target positions or orientations, since the same control input may result in different amounts of movement or rotation across models.
W4 cases additionally support natural-language instructions for subject actions, interactions with objects or the environment, and modifications to environmental rules.
For W5 and W6, controls can be assigned to specific subjects, allowing multiple subjects to be controlled within the same scene.

\paragraph{Capability-guided case design.}
Scene selection and task design are guided by the target capabilities.
Cases cover the following complementary aspects of world modeling,
with variation in scene content, dynamic elements, and control patterns:
\begin{itemize}[leftmargin=*]
    \item \textbf{Natural dynamics:}
    We select scenes with elements that naturally evolve over time,
    such as moving animals, passing vehicles, and vegetation swaying
    in the wind. These cases examine whether scenes develop naturally
    while maintaining coherent appearance and structure without
    action inputs.

    \item \textbf{Appearance and spatial structure:}
We select scenes with distinct foreground and background elements,
varied textures, and clear depth relationships. Viewpoint changes
expose different surfaces and occlusion patterns, allowing assessment
of visual stability, geometric coherence, and preservation of scene
details.


    \item \textbf{Movement and environmental responses:}
We combine timed movement and camera controls with varied spatial
layouts, using both individual and combined inputs with changes
in direction and duration. These cases examine whether scene
responses follow the requested controls in direction and timing,
including environmental effects such as a subject entering water.

    \item \textbf{Long-term persistence:}
    We select scenes with recognizable objects and landmarks and design
    extended control sequences involving translation, rotation, and
    revisits to previously observed regions. Paths and intervals between
    observations are varied to examine the persistence of object identity,
    spatial layout, and scene state over time.

    \item \textbf{Interaction and rule editing:}
    We specify subject actions, interactions with objects or the
    environment, and changes to environmental rules. These cases examine
    how instructed changes affect subsequent scene evolution while
    preserving content that should remain unchanged.
\end{itemize}

\paragraph{Quality control.} Each case undergoes human review to ensure that it defines a clear, executable, and assessable task. Reviewers first verify that the image and descriptions agree on the scene, viewpoint, entities, and initial state. Control requirements vary by W level. Where controls are required, reviewers check that they use the corresponding control interface and are clear, temporally coherent, and feasible in the scene. Reviewers also check that the task specifies a clear evaluation target grounded in observable scene content. We filter out cases with ambiguous evaluation targets, infeasible
controls, or insufficient observable evidence for assessing the
intended capability.



\paragraph{Dataset Statistics}
The benchmark is built on a shared pool of \textbf{1,138 unique cases} spanning W1--W5, with 218, 294, 342, 180, and 104 cases at each level, respectively. Across this pool, we define 74 fine-grained evaluation facets organized into five capability dimensions: perception and representation (342 cases), consistency and state retention (228 cases), causality and causal rollout (299 cases), controllable interaction and counterfactuals (212 cases), and multi-subject coordination (57 cases). Figure~\ref{fig:dataset-statistics}(a) summarizes this hierarchical capability coverage, with capability dimensions in the inner ring and individual facets in the outer ring. Representative facets include action--response alignment, scene revisitation, topology preservation, and accumulated causal effects, reflecting the benchmark's emphasis on interactive behavior, persistent world states, and long-horizon evolution.



\subsection{Evaluation Metrics}

Following Table~\ref{tab:w-hierarchy}, we define four evaluation
dimensions with the following sub-metrics.

\begin{itemize}[leftmargin=*]
\setlength{\itemsep}{3pt}
\setlength{\parsep}{0pt}

\item \textbf{Perception and Representation.}
Video quality (\textbf{VQ}) and perceptual preference (\textbf{HPS})
use MUSIQ~\citep{ke2021musiq} and HPSv3~\citep{ma2025hpsv3widespectrumhumanpreference},
respectively. Imaging stability (\textbf{IS}) aggregates brightness consistency,
color-temperature consistency, and sharpness retention.
Dynamic degree (\textbf{Dyn}) uses RAFT optical flow to measure visible
motion. Instruction following (\textbf{IF}) evaluates atomic prompt
requirements; for W4, it measures only the fidelity of the world before the
first intervention.

\item \textbf{Consistency and State Retention.}
Background consistency (\textbf{BC}) uses masked CLIP similarity to measure
environmental stability. Geometric consistency (\textbf{GC}) and texture
consistency (\textbf{TC}) use DA3-based geometric reprojection and photometric
alignment to assess spatial structure and surface appearance, respectively.
State consistency (\textbf{SC}) evaluates the persistence of object states
and spatial relations, while subject consistency (\textbf{SuC}) uses
DINOv2/CLIP features on SAM2.1 masks to assess identity and appearance
preservation.

\item \textbf{Causality and Causal Rollout.}
Physical causality (\textbf{PC}) and content causality (\textbf{CC}) use
Gemini-based checklists to assess physically plausible processes and
action-to-consequence event progression, respectively.
Interpenetration (\textbf{IP}) evaluates penetration, embedding, unsupported
floating, and invalid contact between relevant objects through visual
judgment.

\item \textbf{Controllable Interaction and Counterfactuals.}
For W1--W3, trajectory accuracy (\textbf{TA}) compares DA3-estimated camera
motion with time-aligned translation commands, while action execution
(\textbf{AE}) evaluates whether commanded actions are visibly completed in
their corresponding temporal segments.
For W4, interaction validity (\textbf{IV}) evaluates whether interventions
follow physically and causally admissible processes, while state fidelity
(\textbf{SF}) measures target-state achievement, persistence, and preservation
of unaffected content. 

\end{itemize}


\definecolor{WAHeader}{HTML}{F3F3FE}
\definecolor{WAAlternate}{HTML}{FAFAFE}
\definecolor{WAAccent}{HTML}{5F588F}
\definecolor{WARule}{HTML}{B8B4D2}
\providecommand{\WAcell}[1]{\makebox[20pt][r]{#1}}
\providecommand{\WAgroup}[1]{{\color{WAAccent}\fontsize{7.0}{7.7}\selectfont\shortstack{#1}}}
\subsection{Experimental Results}
\label{sec:video-experimental-results}

\paragraph{Experimental setup.}
We evaluate 14 interactive world models: Genie 3~\citep{genie3}, HappyOyster~\citep{alibabatokenhub2026happyoyster}, JoyAI-Echo~\citep{zhang2026echowmopenenterableomnimodal}, Alaya-EVOKE~\citep{yin2026alayaevokelinearscalingsupervisionendless}, LingBot-World-v2~\citep{gao2026infinite}, NVIDIA Cosmos3~\citep{agarwal2026cosmos}, Yume-1.5~\citep{Mao_2026_CVPR}, Lyra 2.0~\citep{shen2026lyra20explorablegenerative}, DreamX-World~\citep{dreamxteam2026dreamxworld10generalpurposeinteractive}, Matrix-Game 3.0~\citep{wang2026matrixgame30realtimestreaming}, SANA-WM~\citep{zhu2026sanawmefficientminutescaleworld}, Matrix-Game 2.0~\citep{matrixgame2}, ABot-World~\citep{jiang2026abotworld0infiniteinteractiveworld}, and Open-Oasis~\citep{oasis2024}. For W1, we additionally evaluate five video-generation baselines: HappyHorse 1.1~\citep{happyhorse2026i2v}, Kling 3.0~\cite{kuaishou2026kling30}, MiniMax-H3~\citep{minimax2026h3}, Seedance 2.5~\citep{seed2026seedance25}, and Wan 3.0~\citep{alibaba2026wan30}. The W4 comparison is restricted to HappyOyster and LingBot-World-v2, which support the required intervention setting. Model-specific inference configurations and input adaptations are detailed in Appendix~\ref{app:video-inference}.

We assess overall human preference through pairwise comparisons in the Arena and report capability scores separately for each W level.
Table~\ref{tab:arena-overall} summarizes the Arena Elo ratings together with the level-specific summary scores.
Tables~\ref{tab:video-w1}--\ref{tab:video-w4} provide the corresponding metric breakdowns.
Table~\ref{tab:video-w1} evaluates perception and consistency for W1;
Table~\ref{tab:video-w2} additionally includes causality and interaction metrics for action responsiveness;
Table~\ref{tab:video-w3} uses the same four evaluation dimensions to assess state persistence over extended interaction and revisitation;
and Table~\ref{tab:video-w4} evaluates rule programmability, covering intervention execution, preservation, causal validity, and sustained effects.

\noindent\textbf{Overall results reveal level-specific capability differences.}
HappyOyster achieves the highest Arena Elo of 1263, followed by Genie 3 at 1206
(Table~\ref{tab:arena-overall}). Across the W-level evaluations, however, the
relative performance varies by capability level. Genie 3 obtains the highest
W1 score of 82.8, slightly above HappyOyster at 82.4, whereas HappyOyster achieves
the highest reported W2 and W3 scores, with 78.0 and 76.8, respectively.
For W4, results are currently available only for HappyOyster and
Lingbot-World-v2, with scores of 66.7 and 59.3. These results motivate a
level-wise analysis rather than relying solely on overall Arena preference.

\begin{table}[H]
  \centering
  \small
  \setlength{\tabcolsep}{10pt}
  \renewcommand{\arraystretch}{1.1}
  \caption{Overall evaluation across Arena preference and W1--W4 capability levels. Arena Elo reflects human preference, while W1--W4 report level-specific benchmark performance. Higher is better for all reported scores.}
  \label{tab:arena-overall}
  \arrayrulecolor{WARule}
  \begin{tabular}{lrrrrr}
    \toprule
    \rowcolor{WAHeader}
    \textbf{Model} & \textbf{Arena Elo} $\uparrow$ & \textbf{W1} $\uparrow$ &
    \textbf{W2} $\uparrow$ & \textbf{W3} $\uparrow$ & \textbf{W4} $\uparrow$ \\
    \midrule
    \rowcolor{WAHeader}
    HappyOyster & 1263 & 82.4 & 78.0 & 76.8 & 66.7 \\
    \rowcolor{WAAlternate}
    Genie 3 & 1206 & 82.8 & 74.0 & 75.0 & - \\
    JoyAI-Echo & 1147 & 78.2 & 74.5 & 74.1 & - \\
    \rowcolor{WAAlternate}
    Alaya-EVOKE & 1132 & 79.2 & 73.9 & 74.6 & - \\
    Lingbot-World-v2 & 1115 & 81.9 & 72.6 & 71.0 & 59.3 \\
    \rowcolor{WAAlternate}
    Cosmos3 & 1092 & 81.3 & 70.2 & 64.8 & - \\
    Yume-1.5 & 1063 & 79.6 & 70.2 & 67.7 & - \\
    \rowcolor{WAAlternate}
    Lyra 2.0 & 1059 & 81.0 & 73.5 & 71.6 & - \\
    DreamX-World & 1041 & 80.3 & 72.7 & 72.4 & - \\
    \rowcolor{WAAlternate}
    MatrixGame-3.0 & 943 & 78.8 & 64.6 & 66.1 & - \\
    SANA-WM & 917 & 78.0 & 64.0 & 62.4 & - \\
    \rowcolor{WAAlternate}
    MatrixGame-2.0 & 856 & 63.4 & 61.1 & 59.6 & - \\
    ABot-World & 843 & 69.0 & 64.6 & 60.4 & - \\
    \rowcolor{WAAlternate}
    Open-Oasis & 324 & 53.1 & 43.3 & 39.4 & - \\
    \bottomrule
  \end{tabular}
  \arrayrulecolor{black}
\end{table}


\begin{table}[H]
\centering
\caption{W1: Perceptual World. Evaluation of perceptual quality, visual consistency, and instruction fidelity in world generation. Higher values indicate better performance for all metrics.}
\label{tab:video-w1}
\begingroup
\fontsize{9.0}{10.4}\selectfont
\setlength{\tabcolsep}{1.0pt}
\renewcommand{\arraystretch}{1.12}
\arrayrulecolor{WARule}
\begin{tabular*}{\textwidth}{@{\extracolsep{\fill}}l*{10}{r}@{}}
\toprule
\rowcolor{WAHeader}
 & \multicolumn{5}{c}{\cellcolor{WAHeader}\WAgroup{Perception}}
 & \multicolumn{5}{c}{\cellcolor{WAHeader}\WAgroup{Consistency}} \\
\cmidrule(lr){2-6}\cmidrule(lr){7-11}
\rowcolor{WAHeader}
\textbf{Model}
& \textbf{VQ}
& \textbf{HPS}
& \textbf{IS}
& \textbf{Dyn}
& \textbf{IF}
& \textbf{BC}
& \textbf{GC}
& \textbf{TC}
& \textbf{SC}
& \textbf{SuC} \\
\midrule
ABot-World
& \WAcell{60.89} & \WAcell{53.94} & \WAcell{59.86} & \WAcell{10.58} & \WAcell{88.78}
& \WAcell{92.28} & \WAcell{97.92} & \WAcell{70.11} & \WAcell{63.94} & \WAcell{92.13} \\

\rowcolor{WAAlternate}
Alaya-EVOKE
& \WAcell{66.41} & \WAcell{68.43} & \WAcell{82.98} & \WAcell{51.77} & \WAcell{95.03}
& \WAcell{94.50} & \WAcell{98.42} & \WAcell{74.62} & \WAcell{64.86} & \WAcell{94.90} \\

DreamX-World
& \WAcell{67.40} & \WAcell{70.20} & \WAcell{75.25} & \WAcell{99.63} & \WAcell{93.59}
& \WAcell{93.59} & \WAcell{96.95} & \WAcell{46.26} & \WAcell{66.88} & \WAcell{93.74} \\

\rowcolor{WAAlternate}
Genie 3
& \WAcell{70.46} & \WAcell{70.79} & \WAcell{83.27} & \WAcell{85.14} & \WAcell{90.16}
& \WAcell{94.86} & \WAcell{97.91} & \WAcell{60.85} & \WAcell{82.30} & \WAcell{92.50} \\

HappyHorse 1.1
& \WAcell{70.51} & \WAcell{75.74} & \WAcell{84.32} & \WAcell{62.93} & \WAcell{96.06}
& \WAcell{96.67} & \WAcell{98.05} & \WAcell{54.41} & \WAcell{71.47} & \WAcell{95.28} \\

\rowcolor{WAAlternate}
HappyOyster
& \WAcell{68.32} & \WAcell{71.85} & \WAcell{89.76} & \WAcell{60.01} & \WAcell{95.83}
& \WAcell{97.43} & \WAcell{98.43} & \WAcell{69.01} & \WAcell{77.80} & \WAcell{95.44} \\

JoyAI-Echo
& \WAcell{66.06} & \WAcell{64.31} & \WAcell{76.49} & \WAcell{70.13} & \WAcell{90.58}
& \WAcell{94.26} & \WAcell{98.11} & \WAcell{60.11} & \WAcell{68.90} & \WAcell{93.30} \\

\rowcolor{WAAlternate}
Kling 3.0
& \WAcell{69.20} & \WAcell{73.24} & \WAcell{85.35} & \WAcell{52.29} & \WAcell{95.89}
& \WAcell{97.18} & \WAcell{98.23} & \WAcell{63.16} & \WAcell{71.10} & \WAcell{96.10} \\

Lingbot-World-v2
& \WAcell{68.32} & \WAcell{73.55} & \WAcell{85.24} & \WAcell{52.20} & \WAcell{95.20}
& \WAcell{97.00} & \WAcell{98.76} & \WAcell{75.46} & \WAcell{76.97} & \WAcell{96.47} \\

\rowcolor{WAAlternate}
Lyra 2.0
& \WAcell{68.06} & \WAcell{69.38} & \WAcell{86.26} & \WAcell{4.50} & \WAcell{95.83}
& \WAcell{97.98} & \WAcell{99.44} & \WAcell{94.19} & \WAcell{96.42} & \WAcell{98.18} \\

MatrixGame-2.0
& \WAcell{63.63} & \WAcell{46.34} & \WAcell{75.22} & \WAcell{0.00} & \WAcell{82.65}
& \WAcell{90.25} & \WAcell{98.07} & \WAcell{56.99} & \WAcell{27.98} & \WAcell{92.38} \\

\rowcolor{WAAlternate}
MatrixGame-3.0
& \WAcell{68.75} & \WAcell{71.05} & \WAcell{75.20} & \WAcell{13.18} & \WAcell{92.02}
& \WAcell{95.55} & \WAcell{99.35} & \WAcell{82.49} & \WAcell{95.41} & \WAcell{94.87} \\

MiniMax-H3
& \WAcell{68.66} & \WAcell{69.90} & \WAcell{88.49} & \WAcell{46.16} & \WAcell{96.45}
& \WAcell{97.27} & \WAcell{98.48} & \WAcell{71.36} & \WAcell{83.39} & \WAcell{96.00} \\

\rowcolor{WAAlternate}
Cosmos3
& \WAcell{67.77} & \WAcell{69.62} & \WAcell{94.43} & \WAcell{11.65} & \WAcell{95.02}
& \WAcell{97.26} & \WAcell{99.35} & \WAcell{92.12} & \WAcell{89.08} & \WAcell{97.05} \\

Open-Oasis
& \WAcell{47.88} & \WAcell{14.92} & \WAcell{45.83} & \WAcell{1.90} & \WAcell{81.55}
& \WAcell{83.84} & \WAcell{97.06} & \WAcell{44.85} & \WAcell{22.29} & \WAcell{90.79} \\

\rowcolor{WAAlternate}
SANA-WM
& \WAcell{66.74} & \WAcell{59.07} & \WAcell{84.88} & \WAcell{65.79} & \WAcell{91.10}
& \WAcell{95.42} & \WAcell{98.37} & \WAcell{61.35} & \WAcell{64.13} & \WAcell{93.32} \\

Seedance2.5
& \WAcell{67.42} & \WAcell{73.10} & \WAcell{90.28} & \WAcell{36.82} & \WAcell{95.06}
& \WAcell{97.53} & \WAcell{98.57} & \WAcell{76.76} & \WAcell{83.79} & \WAcell{96.79} \\

\rowcolor{WAAlternate}
Wan 3.0
& \WAcell{69.60} & \WAcell{77.97} & \WAcell{84.51} & \WAcell{62.39} & \WAcell{95.81}
& \WAcell{96.63} & \WAcell{97.88} & \WAcell{58.78} & \WAcell{74.22} & \WAcell{94.69} \\

Yume-1.5
& \WAcell{69.35} & \WAcell{67.61} & \WAcell{66.52} & \WAcell{86.88} & \WAcell{93.92}
& \WAcell{94.62} & \WAcell{97.42} & \WAcell{61.02} & \WAcell{64.50} & \WAcell{93.75} \\
\bottomrule
\end{tabular*}
\arrayrulecolor{black}
\endgroup
\end{table}

\noindent\textbf{W1: appearance quality and consistency.} W1 reveals a clear separation between perceptual quality and world consistency. While many models achieve strong visual quality and instruction following, these strengths do not necessarily translate into stable geometry, appearance, or world state over time. In particular, models with competitive perceptual scores can still exhibit substantially weaker temporal or state consistency, indicating that visually appealing generation alone is insufficient to maintain a coherent world representation. These results support evaluating perception and consistency as complementary rather than interchangeable capabilities.

\begin{table}[H]
\centering
\caption{W2: Interactive World. Evaluation of action responsiveness, local state transitions, physical and causal consistency, and interaction fidelity. Higher values indicate better performance for all metrics.}
\label{tab:video-w2}
\begingroup
\fontsize{7.6}{8.9}\selectfont
\setlength{\tabcolsep}{1.0pt}
\renewcommand{\arraystretch}{1.12}
\arrayrulecolor{WARule}
\begin{tabular*}{\textwidth}{@{\extracolsep{\fill}}l*{15}{r}@{}}
\toprule
\rowcolor{WAHeader}
 & \multicolumn{5}{c}{\cellcolor{WAHeader}\WAgroup{Perception}}
 & \multicolumn{5}{c}{\cellcolor{WAHeader}\WAgroup{Consistency}}
 & \multicolumn{3}{c}{\cellcolor{WAHeader}\WAgroup{Causality}}
 & \multicolumn{2}{c}{\cellcolor{WAHeader}\WAgroup{Interaction}} \\
\cmidrule(lr){2-6}\cmidrule(lr){7-11}\cmidrule(lr){12-14}\cmidrule(lr){15-16}
\rowcolor{WAHeader}
\textbf{Model}
& \textbf{VQ} & \textbf{HPS} & \textbf{IS} & \textbf{Dyn} & \textbf{IF}
& \textbf{BC} & \textbf{GC} & \textbf{TC} & \textbf{SC} & \textbf{SuC}
& \textbf{PC} & \textbf{CC} & \textbf{IP}
& \textbf{TA} & \textbf{AE} \\
\midrule

ABot-World
& \WAcell{59.33} & \WAcell{48.09} & \WAcell{49.50} & \WAcell{48.32} & \WAcell{88.46}
& \WAcell{87.68} & \WAcell{90.93} & \WAcell{51.45} & \WAcell{40.61} & \WAcell{87.52}
& \WAcell{56.45} & \WAcell{53.54} & \WAcell{86.74}
& \WAcell{59.38} & \WAcell{65.91} \\

\rowcolor{WAAlternate}
Alaya-EVOKE
& \WAcell{66.61} & \WAcell{64.87} & \WAcell{68.81} & \WAcell{88.84} & \WAcell{91.81}
& \WAcell{89.77} & \WAcell{92.62} & \WAcell{41.42} & \WAcell{57.89} & \WAcell{88.89}
& \WAcell{73.68} & \WAcell{62.80} & \WAcell{96.32}
& \WAcell{70.61} & \WAcell{64.83} \\

DreamX-World
& \WAcell{64.64} & \WAcell{62.52} & \WAcell{70.15} & \WAcell{99.82} & \WAcell{91.52}
& \WAcell{90.30} & \WAcell{89.88} & \WAcell{32.98} & \WAcell{52.15} & \WAcell{89.43}
& \WAcell{69.30} & \WAcell{60.20} & \WAcell{91.68}
& \WAcell{72.09} & \WAcell{64.84} \\

\rowcolor{WAAlternate}
Genie 3
& \WAcell{68.55} & \WAcell{66.84} & \WAcell{67.76} & \WAcell{88.87} & \WAcell{91.71}
& \WAcell{90.48} & \WAcell{93.93} & \WAcell{47.32} & \WAcell{67.99} & \WAcell{87.60}
& \WAcell{78.64} & \WAcell{59.80} & \WAcell{97.07}
& \WAcell{67.89} & \WAcell{58.62} \\

HappyOyster
& \WAcell{67.25} & \WAcell{67.82} & \WAcell{73.98} & \WAcell{81.25} & \WAcell{94.14}
& \WAcell{93.79} & \WAcell{91.36} & \WAcell{35.94} & \WAcell{65.60} & \WAcell{90.35}
& \WAcell{84.90} & \WAcell{77.51} & \WAcell{95.96}
& \WAcell{67.31} & \WAcell{80.20} \\

\rowcolor{WAAlternate}
JoyAI-Echo
& \WAcell{65.84} & \WAcell{59.92} & \WAcell{67.02} & \WAcell{87.73} & \WAcell{90.87}
& \WAcell{90.33} & \WAcell{90.15} & \WAcell{38.15} & \WAcell{62.04} & \WAcell{87.80}
& \WAcell{75.76} & \WAcell{67.33} & \WAcell{94.03}
& \WAcell{72.58} & \WAcell{69.53} \\

Lingbot-World-v2
& \WAcell{65.98} & \WAcell{64.21} & \WAcell{65.12} & \WAcell{81.68} & \WAcell{91.66}
& \WAcell{90.68} & \WAcell{88.90} & \WAcell{40.77} & \WAcell{52.18} & \WAcell{87.71}
& \WAcell{68.92} & \WAcell{60.48} & \WAcell{94.38}
& \WAcell{73.48} & \WAcell{66.61} \\

\rowcolor{WAAlternate}
Lyra 2.0
& \WAcell{66.51} & \WAcell{59.04} & \WAcell{69.98} & \WAcell{56.05} & \WAcell{92.89}
& \WAcell{91.60} & \WAcell{91.71} & \WAcell{52.35} & \WAcell{71.77} & \WAcell{90.66}
& \WAcell{67.90} & \WAcell{59.46} & \WAcell{93.53}
& \WAcell{76.28} & \WAcell{67.28} \\

MatrixGame-2.0
& \WAcell{63.22} & \WAcell{43.90} & \WAcell{63.90} & \WAcell{45.29} & \WAcell{83.54}
& \WAcell{87.78} & \WAcell{91.85} & \WAcell{43.57} & \WAcell{32.38} & \WAcell{87.79}
& \WAcell{40.46} & \WAcell{46.63} & \WAcell{64.43}
& \WAcell{70.45} & \WAcell{60.08} \\

\rowcolor{WAAlternate}
MatrixGame-3.0
& \WAcell{64.22} & \WAcell{50.54} & \WAcell{57.36} & \WAcell{56.54} & \WAcell{84.25}
& \WAcell{86.99} & \WAcell{89.11} & \WAcell{46.30} & \WAcell{43.47} & \WAcell{86.33}
& \WAcell{45.08} & \WAcell{47.78} & \WAcell{82.83}
& \WAcell{72.79} & \WAcell{60.48} \\

Cosmos3
& \WAcell{67.49} & \WAcell{69.07} & \WAcell{89.91} & \WAcell{51.22} & \WAcell{96.04}
& \WAcell{95.56} & \WAcell{98.19} & \WAcell{67.58} & \WAcell{80.54} & \WAcell{95.00}
& \WAcell{69.20} & \WAcell{33.18} & \WAcell{96.65}
& \WAcell{61.90} & \WAcell{42.65} \\

\rowcolor{WAAlternate}
Open-Oasis
& \WAcell{44.81} & \WAcell{11.71} & \WAcell{39.12} & \WAcell{3.54} & \WAcell{81.43}
& \WAcell{85.23} & \WAcell{94.65} & \WAcell{35.31} & \WAcell{21.16} & \WAcell{90.83}
& \WAcell{19.72} & \WAcell{9.98} & \WAcell{75.33}
& \WAcell{48.72} & \WAcell{24.89} \\

SANA-WM
& \WAcell{65.12} & \WAcell{51.77} & \WAcell{72.01} & \WAcell{87.03} & \WAcell{87.40}
& \WAcell{90.41} & \WAcell{93.60} & \WAcell{33.56} & \WAcell{34.08} & \WAcell{87.84}
& \WAcell{41.52} & \WAcell{47.95} & \WAcell{62.36}
& \WAcell{65.61} & \WAcell{64.07} \\

\rowcolor{WAAlternate}
Yume-1.5
& \WAcell{69.10} & \WAcell{65.74} & \WAcell{63.80} & \WAcell{95.46} & \WAcell{92.88}
& \WAcell{92.43} & \WAcell{94.00} & \WAcell{48.12} & \WAcell{61.70} & \WAcell{91.19}
& \WAcell{70.50} & \WAcell{42.83} & \WAcell{93.54}
& \WAcell{72.34} & \WAcell{41.87} \\
\bottomrule
\end{tabular*}
\arrayrulecolor{black}
\endgroup
\end{table}

\noindent\textbf{W2: control response and its consequences.}
W2 distinguishes successful action execution from the correctness of the resulting world transition. High TA or AE scores indicate that a model can follow the requested control or action, but do not guarantee that the subsequent state remains coherent. Variations in SC and CC further show that models with similar control-following performance can differ substantially in whether the resulting state is preserved and whether the observed transition is causally consistent. Together with the perception metrics, these results separate visually plausible action responses from interactions that maintain coherent state evolution and causal structure.

\noindent\textbf{State persistence and rule programmability.}
W3 and W4 extend the evaluation beyond immediate action response along two complementary directions. W3 examines whether previously established structure, state, and identity remain stable over extended interaction and revisitation. W4 instead evaluates whether natural-language interventions can modify the world as intended, preserve unrelated content, and consistently govern subsequent evolution.

\begin{table}[H]
\centering
\caption{W3: Persistent World. Evaluation of long-horizon world persistence, including spatial, visual, and state consistency under extended interaction and revisitation. Higher values indicate better performance for all metrics.}
\label{tab:video-w3}
\begingroup
\fontsize{7.6}{8.9}\selectfont
\setlength{\tabcolsep}{1.0pt}
\renewcommand{\arraystretch}{1.12}
\arrayrulecolor{WARule}
\begin{tabular*}{\textwidth}{@{\extracolsep{\fill}}l*{15}{r}@{}}
\toprule
\rowcolor{WAHeader}
 & \multicolumn{5}{c}{\cellcolor{WAHeader}\WAgroup{Perception}}
 & \multicolumn{5}{c}{\cellcolor{WAHeader}\WAgroup{Consistency}}
 & \multicolumn{3}{c}{\cellcolor{WAHeader}\WAgroup{Causality}}
 & \multicolumn{2}{c}{\cellcolor{WAHeader}\WAgroup{Interaction}} \\
\cmidrule(lr){2-6}\cmidrule(lr){7-11}\cmidrule(lr){12-14}\cmidrule(lr){15-16}
\rowcolor{WAHeader}
\textbf{Model}
& \textbf{VQ} & \textbf{HPS} & \textbf{IS} & \textbf{Dyn} & \textbf{IF}
& \textbf{BC} & \textbf{GC} & \textbf{TC} & \textbf{SC} & \textbf{SuC}
& \textbf{PC} & \textbf{CC} & \textbf{IP}
& \textbf{TA} & \textbf{AE} \\
\midrule

ABot-World
& \WAcell{55.25} & \WAcell{31.08} & \WAcell{37.19} & \WAcell{70.63} & \WAcell{83.00}
& \WAcell{81.87} & \WAcell{79.41} & \WAcell{39.77} & \WAcell{26.08} & \WAcell{84.31}
& \WAcell{47.05} & \WAcell{52.87} & \WAcell{78.25}
& \WAcell{54.05} & \WAcell{75.08} \\

\rowcolor{WAAlternate}
Alaya-EVOKE
& \WAcell{68.46} & \WAcell{67.95} & \WAcell{62.11} & \WAcell{93.57} & \WAcell{92.62}
& \WAcell{88.92} & \WAcell{88.81} & \WAcell{34.06} & \WAcell{52.05} & \WAcell{89.58}
& \WAcell{73.74} & \WAcell{71.77} & \WAcell{96.03}
& \WAcell{63.46} & \WAcell{77.07} \\

DreamX-World
& \WAcell{64.18} & \WAcell{61.82} & \WAcell{60.77} & \WAcell{99.87} & \WAcell{89.44}
& \WAcell{88.73} & \WAcell{79.87} & \WAcell{24.76} & \WAcell{45.96} & \WAcell{88.81}
& \WAcell{67.08} & \WAcell{70.12} & \WAcell{90.84}
& \WAcell{61.47} & \WAcell{84.06} \\

\rowcolor{WAAlternate}
Genie 3
& \WAcell{65.88} & \WAcell{62.02} & \WAcell{54.99} & \WAcell{91.63} & \WAcell{91.54}
& \WAcell{88.52} & \WAcell{88.95} & \WAcell{41.77} & \WAcell{60.29} & \WAcell{86.65}
& \WAcell{81.63} & \WAcell{74.60} & \WAcell{97.22}
& \WAcell{62.18} & \WAcell{76.22} \\


\rowcolor{WAAlternate}
HappyOyster
& \WAcell{66.73} & \WAcell{62.53} & \WAcell{63.06} & \WAcell{86.99} & \WAcell{93.91}
& \WAcell{92.22} & \WAcell{84.04} & \WAcell{27.65} & \WAcell{58.24} & \WAcell{89.78}
& \WAcell{87.67} & \WAcell{83.78} & \WAcell{95.28}
& \WAcell{60.79} & \WAcell{86.06} \\

JoyAI-Echo
& \WAcell{66.20} & \WAcell{57.92} & \WAcell{63.04} & \WAcell{93.04} & \WAcell{90.01}
& \WAcell{89.54} & \WAcell{84.16} & \WAcell{30.44} & \WAcell{50.06} & \WAcell{87.67}
& \WAcell{73.48} & \WAcell{76.05} & \WAcell{91.69}
& \WAcell{63.01} & \WAcell{84.15} \\

\rowcolor{WAAlternate}
Lingbot-World-v2
& \WAcell{66.36} & \WAcell{64.34} & \WAcell{52.17} & \WAcell{89.65} & \WAcell{89.10}
& \WAcell{88.49} & \WAcell{81.01} & \WAcell{26.29} & \WAcell{41.87} & \WAcell{85.49}
& \WAcell{69.47} & \WAcell{69.81} & \WAcell{94.20}
& \WAcell{57.67} & \WAcell{80.37} \\

Lyra 2.0
& \WAcell{67.95} & \WAcell{57.56} & \WAcell{60.44} & \WAcell{77.99} & \WAcell{91.59}
& \WAcell{89.69} & \WAcell{86.91} & \WAcell{40.11} & \WAcell{64.56} & \WAcell{89.20}
& \WAcell{64.33} & \WAcell{62.89} & \WAcell{89.74}
& \WAcell{62.12} & \WAcell{75.40} \\

\rowcolor{WAAlternate}
MatrixGame-2.0
& \WAcell{63.10} & \WAcell{39.75} & \WAcell{48.11} & \WAcell{69.51} & \WAcell{79.58}
& \WAcell{83.98} & \WAcell{80.58} & \WAcell{26.42} & \WAcell{25.73} & \WAcell{86.06}
& \WAcell{36.84} & \WAcell{50.19} & \WAcell{73.29}
& \WAcell{58.08} & \WAcell{70.66} \\

MatrixGame-3.0
& \WAcell{66.20} & \WAcell{48.76} & \WAcell{49.61} & \WAcell{79.41} & \WAcell{84.90}
& \WAcell{84.77} & \WAcell{84.22} & \WAcell{31.91} & \WAcell{37.49} & \WAcell{84.21}
& \WAcell{49.90} & \WAcell{61.68} & \WAcell{85.51}
& \WAcell{58.04} & \WAcell{78.47} \\

\rowcolor{WAAlternate}
Cosmos3
& \WAcell{68.09} & \WAcell{66.40} & \WAcell{82.21} & \WAcell{56.75} & \WAcell{92.50}
& \WAcell{94.16} & \WAcell{97.42} & \WAcell{55.72} & \WAcell{63.10} & \WAcell{94.60}
& \WAcell{61.45} & \WAcell{25.82} & \WAcell{85.13}
& \WAcell{57.06} & \WAcell{37.94} \\

Open-Oasis
& \WAcell{41.21} & \WAcell{7.30} & \WAcell{29.48} & \WAcell{4.52} & \WAcell{75.21}
& \WAcell{81.51} & \WAcell{92.12} & \WAcell{31.82} & \WAcell{21.17} & \WAcell{88.59}
& \WAcell{16.45} & \WAcell{7.33} & \WAcell{71.24}
& \WAcell{47.66} & \WAcell{14.98} \\

\rowcolor{WAAlternate}
SANA-WM
& \WAcell{63.20} & \WAcell{41.87} & \WAcell{59.58} & \WAcell{90.97} & \WAcell{84.32}
& \WAcell{87.22} & \WAcell{88.76} & \WAcell{24.81} & \WAcell{29.12} & \WAcell{86.48}
& \WAcell{42.35} & \WAcell{55.84} & \WAcell{56.12}
& \WAcell{59.47} & \WAcell{74.72} \\

Yume-1.5
& \WAcell{69.36} & \WAcell{61.78} & \WAcell{49.24} & \WAcell{96.65} & \WAcell{89.69}
& \WAcell{90.11} & \WAcell{86.11} & \WAcell{41.90} & \WAcell{44.44} & \WAcell{90.12}
& \WAcell{63.75} & \WAcell{52.64} & \WAcell{87.54}
& \WAcell{60.75} & \WAcell{57.03} \\
\bottomrule
\end{tabular*}
\arrayrulecolor{black}
\endgroup
\end{table}

\noindent\textbf{W3: persistence across revisits.}
W3 highlights the difficulty of preserving a coherent world over extended interaction and revisitation. Consistency metrics, including BC, GC, TC, SC, and SuC, assess whether established layout, appearance, temporal continuity, world state, and subject identity remain stable as interaction proceeds and previously observed regions are revisited. Strong perceptual quality alone is therefore insufficient evidence of world persistence, since visually plausible frames may still exhibit geometric drift, appearance changes, or altered states over time. Because W2 and W3 use different case pools, differences in their aggregate scores should not be interpreted as a pure duration effect.

\begin{table}[H]
\centering
\caption{W4: Programmable World. Evaluation of explicit world interventions, including intervention execution, preservation of unaffected content, causal validity, and sustained intervention effects. Higher values indicate better performance for all metrics.}
\label{tab:video-w4}
\begingroup
\fontsize{7.6}{8.9}\selectfont
\setlength{\tabcolsep}{1.0pt}
\renewcommand{\arraystretch}{1.12}
\arrayrulecolor{WARule}
\begin{tabular*}{\textwidth}{@{\extracolsep{\fill}}l*{15}{r}@{}}
\toprule
\rowcolor{WAHeader}
 & \multicolumn{5}{c}{\cellcolor{WAHeader}\WAgroup{Perception}} & \multicolumn{5}{c}{\cellcolor{WAHeader}\WAgroup{Consistency}} & \multicolumn{3}{c}{\cellcolor{WAHeader}\WAgroup{Causality}} & \multicolumn{2}{c}{\cellcolor{WAHeader}\WAgroup{Interaction}} \\
\cmidrule(lr){2-6}\cmidrule(lr){7-11}\cmidrule(lr){12-14}\cmidrule(lr){15-16}
\rowcolor{WAHeader}
\textbf{Model} & \textbf{VQ} & \textbf{HPS} & \textbf{IS} & \textbf{Dyn} & \textbf{IF} & \textbf{BC} & \textbf{GC} & \textbf{TC} & \textbf{SC} & \textbf{SuC} & \textbf{PC} & \textbf{CC} & \textbf{IP} & \textbf{IV} & \textbf{SF} \\
\midrule
HappyOyster & \WAcell{68.24} & \WAcell{67.12} & \WAcell{85.85} & \WAcell{81.06} & \WAcell{94.25} & \WAcell{95.49} & \WAcell{96.64} & \WAcell{49.32} & \WAcell{59.44} & \WAcell{91.98} & \WAcell{53.35} & \WAcell{23.51} & \WAcell{88.75} & \WAcell{68.55} & \WAcell{38.61} \\
\rowcolor{WAAlternate}
Lingbot-World-v2 & \WAcell{69.14} & \WAcell{65.37} & \WAcell{71.71} & \WAcell{80.18} & \WAcell{94.80} & \WAcell{93.79} & \WAcell{95.79} & \WAcell{35.20} & \WAcell{42.11} & \WAcell{91.12} & \WAcell{46.01} & \WAcell{20.02} & \WAcell{81.81} & \WAcell{51.27} & \WAcell{28.58} \\
\bottomrule
\end{tabular*}
\arrayrulecolor{black}
\endgroup
\end{table}

\noindent\textbf{W4: intervention execution and preservation.}
W4 evaluates whether a model can execute a natural-language intervention while preserving the parts of the world that should remain unchanged. The interaction metrics IV and SF capture whether the intended intervention is correctly realized and sustained, while consistency and causality metrics assess whether unrelated scene content remains coherent and the resulting world evolution remains compatible with the intervention. The current results show that successful intervention execution alone does not guarantee stable post-edit behavior, motivating joint evaluation of edit success, preservation, and subsequent consistency.

\noindent\textbf{Rule edits must govern subsequent evolution.}
A valid rule edit should alter not only the immediate output but also the subsequent dynamics of the generated world. Causal evaluation therefore uses the edited rule, rather than the original world dynamics, as the reference for later behavior. For example, after a gravity intervention, later motion should follow the modified rule while unrelated scene state remains preserved. W4 thus evaluates whether an intervention becomes a persistent part of the world dynamics rather than a transient visual change.

\begin{figure}[t]
  \centering
  \includegraphics[width=\textwidth]{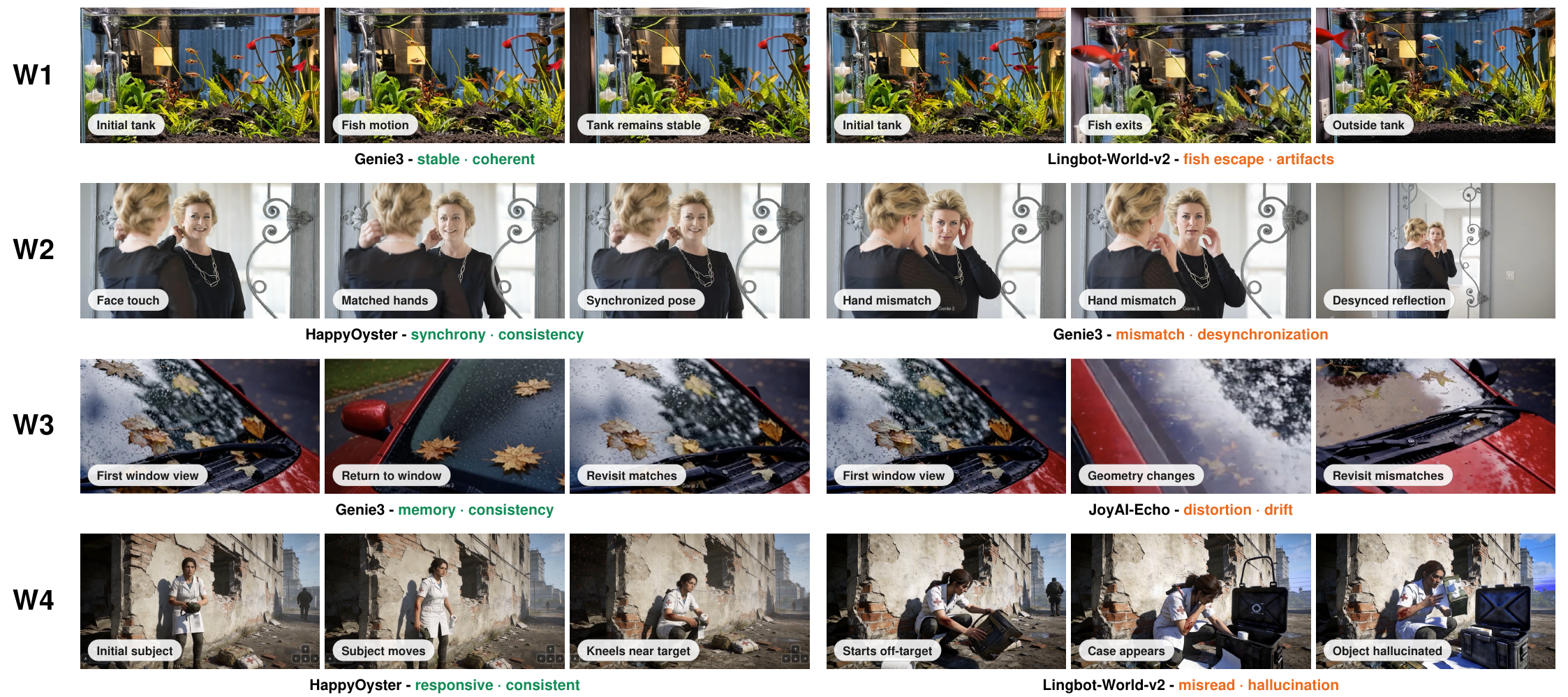}
  \caption{Qualitative examples of video world models in HappyWorld-Bench.}
  \label{fig:embodied-w2-failures}
\end{figure}
\providecolor{WAHeader}{HTML}{F3F3FE}
\providecolor{WAAlternate}{HTML}{FAFAFE}
\providecolor{WAAccent}{HTML}{5F588F}
\providecolor{WARule}{HTML}{B8B4D2}
\providecommand{\WAcell}[1]{\makebox[20pt][r]{#1}}
\providecommand{\WAgroup}[1]{{\color{WAAccent}\fontsize{7.0}{7.7}\selectfont\shortstack{#1}}}

\section{Spatial World Model Track}
\label{sec:reconstruction}
\suppressfloats[t]

The Spatial World Model Track evaluates explicit, renderable environments constructed from images and/or scene descriptions, including generation of regions not observed in the input. Outputs may be gaussian scenes, meshes, or other spatial representations. Beyond visual quality, the track assesses whether these environments enable exploration and physical support, and whether editing or expansion preserves existing content. These capabilities are environment-level prerequisites for future general and embodied AI applications.

\subsection{Data Construction}
\label{sec:recon-data}
We build the spatial-world dataset from filtered video-track inputs and additional cases collected for specific spatial tasks. Note that no existing method can achieve spatiotemporal world modeling, so we select only static scenes.

\paragraph{Scene Selection and Manual Review.}
We select candidate scenes from the video track and manually review them for scene evaluation. For the present static-scene setting, review excludes initial images containing people or visible body parts, including hands and legs. And descriptions from video track are rewritten to remove first-/third-person role phrasing.

\begin{figure}[htbp]
  \centering
  \begin{subfigure}[t]{0.24\linewidth}
      \centering
      \includegraphics[width=\linewidth]{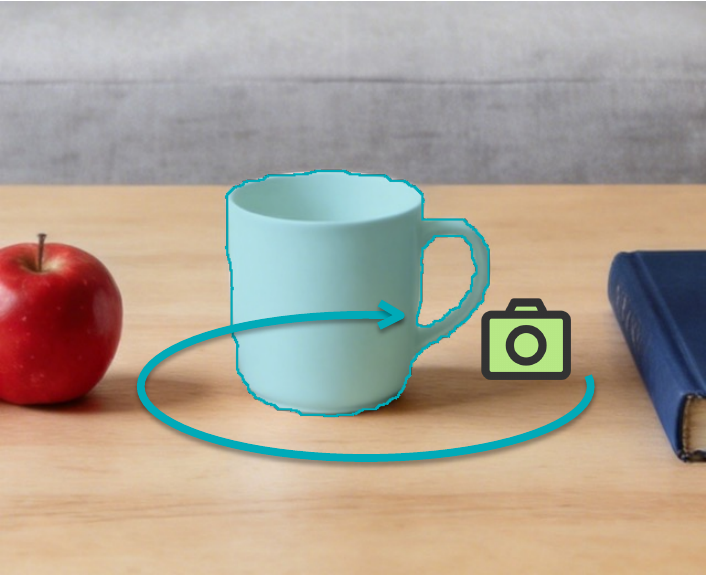}
      \caption{An object target for orbit-based consistency evaluation.}
      \label{fig:recon-data-a}
  \end{subfigure}\hfill
  \begin{subfigure}[t]{0.24\linewidth}
      \centering
      \includegraphics[width=\linewidth]{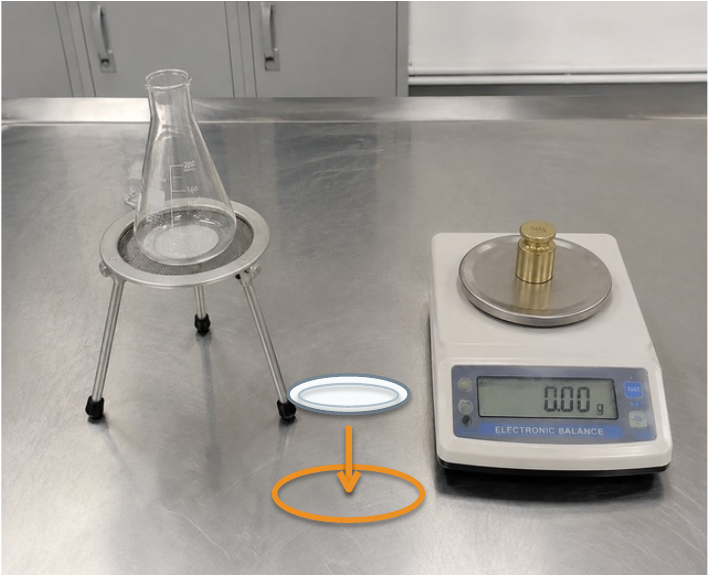}
      \caption{A support surface for plate-placement trials.}
      \label{fig:recon-data-b}
  \end{subfigure}\hfill
  \begin{subfigure}[t]{0.24\linewidth}
      \centering
      \includegraphics[width=\linewidth]{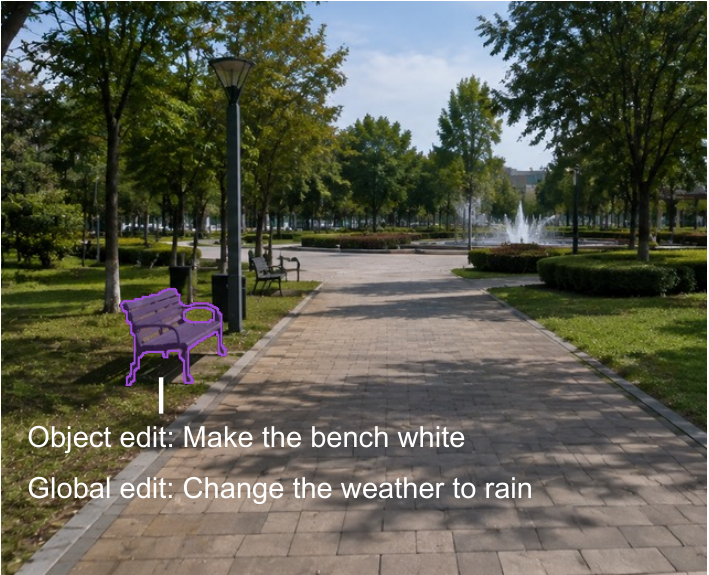}
      \caption{Example instructions for object-level and global editing.}
      \label{fig:recon-data-c}
  \end{subfigure}\hfill
  \begin{subfigure}[t]{0.24\linewidth}
      \centering
      \includegraphics[width=\linewidth]{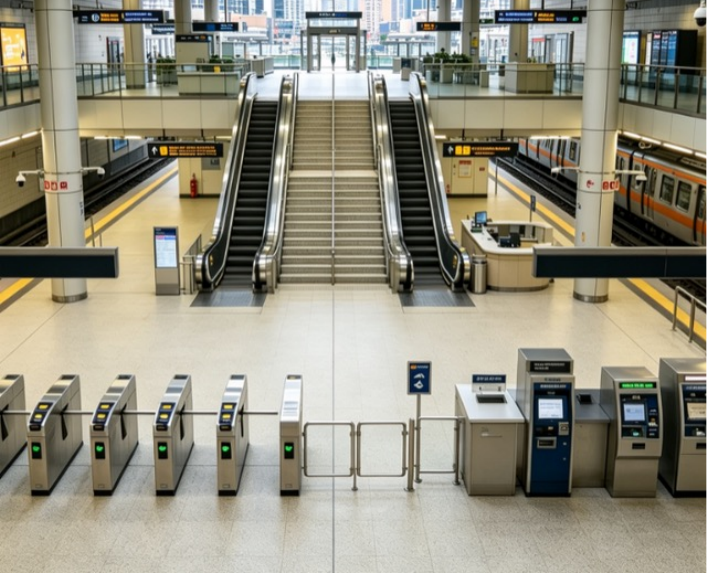}
      \caption{A subway-station scene selected for spatial expansion.}
      \label{fig:recon-data-d}
  \end{subfigure}

  \caption{Task-specific data and annotation examples.}
  \label{fig:recon-data}
\end{figure}

\paragraph{Task-Specific Collection and Annotation.}
Generic video conditions do not necessarily provide support surfaces, identifiable edit targets, or meaningful expansion boundaries. Selected cases from \S\ref{sec:video-data} are therefore supplemented with newly collected or purpose-selected conditions and task-specific annotations:
\begin{itemize}[leftmargin=*,nosep]
    \item \textbf{Object consistency:} We annotate up to two visible, unobstructed objects per scene for orbit-based evaluation.
    \item \textbf{Physical support:} We select scenes with identifiable support surfaces, such as tabletops, and annotate the target surface for six plate-placement trials.
    \item \textbf{Editing:} We construct scene--instruction pairs with explicit targets and permitted changes. The editing set covers six object-level categories (addition, deletion, replacement, color, state, and material) and four global categories (weather, lighting/time, atmospheric effects, and terrain).
    \item \textbf{Expansion:} We collect scenes that allow further spatial expansion, including open doorways, corridors, paths, and open landscapes.
\end{itemize}
Scenes are selected separately for each task according to its requirements. For example, object removal or modification requires an identifiable target object, while physical support requires a surface for plate placement. Figure~\ref{fig:recon-data} shows examples of the task-specific scenes and annotations.

\paragraph{Dataset Statistics.} The final dataset contains 300 scenes: 266 snapshot scenes and 34 separate spatial expansion cases. For object consistency, we annotate 424 target objects within the snapshot scenes. For physical support, we identify and annotate support surfaces in 72 scenes. For editing, we provide 30 scene--instruction pairs covering 18 object-level and 12 global edits. For expansion, we select 34 scenes that allow further spatial expansion.

\FloatBarrier

\subsection{Evaluation Metrics}
\label{sec:recon-metrics}

We evaluate spatial world models through observable quality (W1), physical usability (W2), scene- and object-level consistency (W3), controlled editing (W4), and expansion with preservation (W5). W6 is unscored. Appearance-based evaluation primarily uses videos rendered along predefined camera trajectories, including 360-degree views, scene traversal, and object orbits. Geometry-based measurements and physical simulation are used to assess navigable area, spatial expansion, and placement stability.

\paragraph{W1: Generative Construction.}
Observable quality uses TOPIQ-KonIQ~\citep{spatial_topiq}, LAION aesthetics~\citep{spatial_laion}, perceptual index~\citep{spatial_pi} (PI), and Q-Align~\citep{spatial_qalign}. Perceptual preference uses HPSv3~\citep{ma2025hpsv3widespectrumhumanpreference}, while Qwen3-VL~\citep{spatial_qwen3vl} assesses visual artifact, sharpness, and layout plausibility. Condition alignment uses CLIP~\citep{radford2021clip} text--image and image--image similarity. Hole-free ratio measures rendered coverage, and named-object coverage checks the presence of requested categories.

\paragraph{W2: Interactive Simulation.}
Physical usability is assessed through navigation and support on exported geometry. Navigable ratio uses the largest connected navigation-mesh component to measure space available for continuous exploration. Placement rate measures stable plate placement on annotated support surfaces in a simulator. Together, these tests assess whether the generated environment enables movement and physical support.

\paragraph{W3: State Persistence.}
\textit{Scene-level consistency.}
In-bounds ratio measures spatial validity along prescribed camera trajectories. Brightness and hue consistency assess appearance stability across views. Location HPS F1 combines usable-location coverage with HPSv3 consistency across sampled places.
\noindent\textit{Object-level consistency.}
Object HPS F1 combines SAM3~\citep{carion2025sam3segmentconcepts} visibility with cross-view HPSv3 consistency. Track F1 uses TAPIP3D~\citep{spatial_tapip3d} to assess return correspondence, accounting for visibility. VLM F1 combines valid-view coverage with visual artifact and object completeness. These orbit-based scores assess whether entities remain observable, intact, and identifiable across viewpoints.

\paragraph{W4: Programmable Dynamics.}
Controlled editing pairs instruction satisfaction with preservation of non-target content. Edit success rate measures the fraction of editing cases whose resulting scenes satisfy the requested instruction: each case scores 1 if Qwen3-VL judges that at least one evaluated post-edit view satisfies the instruction, and 0 otherwise. Background RGB retention, LPIPS~\citep{spatial_lpips} similarity, and DINOv2~\citep{spatial_dinov2} similarity assess object-edit locality; depth agreement assesses geometric preservation under global edits. Object and global edit scores combine per-case edit success with the corresponding preservation measures.

\paragraph{W5: Scalable Shared World.}
Expansion is evaluated through spatial growth, boundary coverage, and regional preservation. Area score uses before/after footprints; junction score measures improved rendered coverage at the expansion interface. New-region quality uses HPSv3, while old-region and path preservation use paired HPSv3 observations in the existing world and along its connecting path. These measures assess whether additional space retains the quality of previously available observations.

\paragraph{Reporting.}
All scores are normalized to $[0,1]$ and multiplied by 100 for reporting in the tables, yielding a $[0,100]$ scale, with higher values better. World scores average the available metrics, retaining only F1 within precision/recall families. Untested groups are marked by a dash. Appendix~\ref{supp:recon-specification} provides the full metric inventory, calculation formulas, sampling procedures, normalization conventions, and validity rules.

\FloatBarrier

\subsection{Experimental Results}
\label{sec:recon-results}

\begin{figure}[t]
  \centering
  \includegraphics[width=\textwidth]{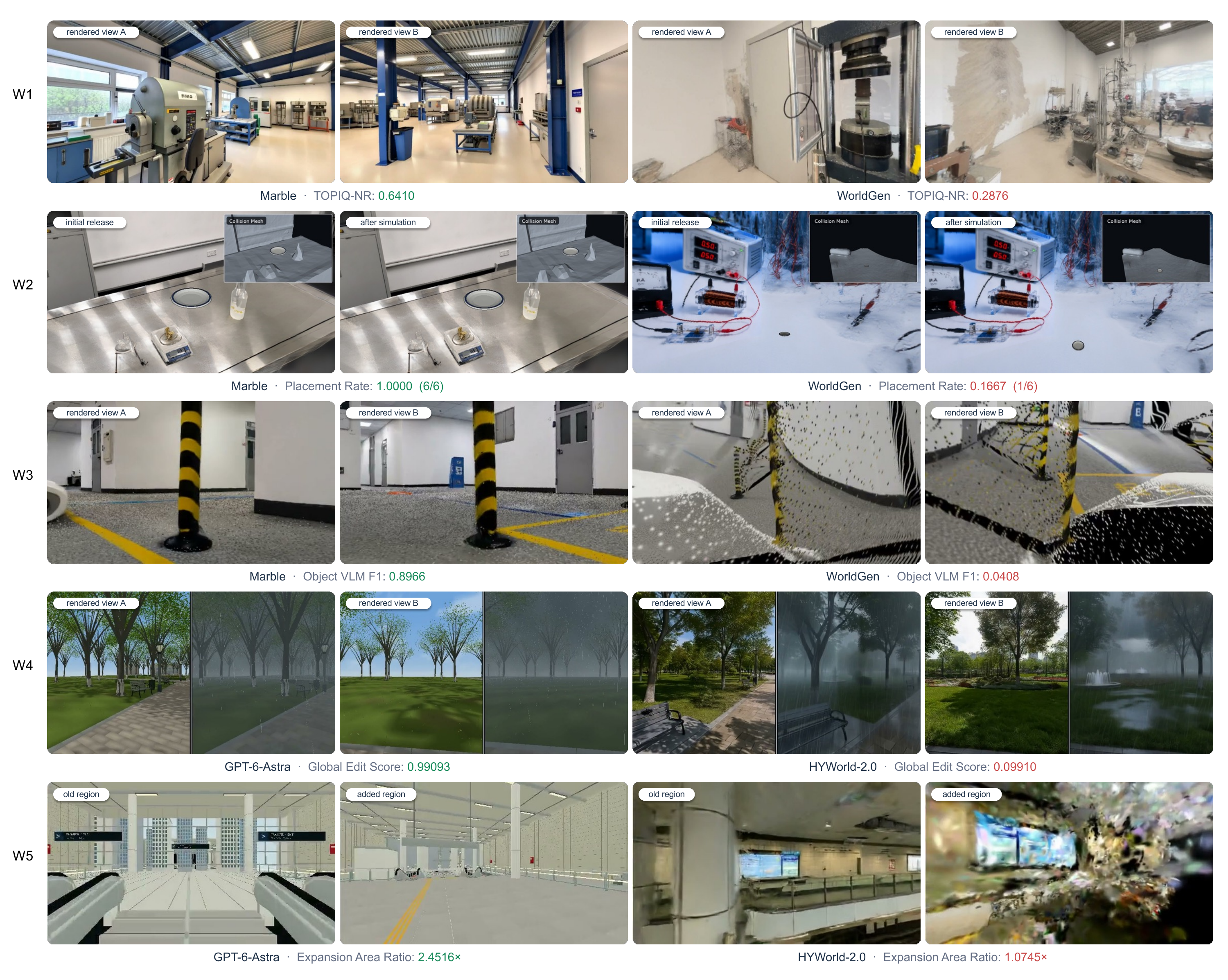}
  \caption{Qualitative examples of spatial world models in HappyWorld-Bench.}
  \label{fig:embodied-w2-failures}
\end{figure}

We first compare overall performance across W1--W5, then analyze
the individual metrics within each group. The analysis follows the progression from observable construction to physical use, persistent observation, controlled modification, and spatial extension. All tables use the same fixed baseline order, which is not a ranking on every metric.

\begin{table}[H]
\centering
\caption{Capability-adjusted Arena Elo and benchmark performance in the spatial world model track.}
\label{tab:recon-world-summary}
\begingroup
\small
\providecolor{WAHeader}{HTML}{F3F3FE}
\providecolor{WAAlternate}{HTML}{FAFAFE}
\providecolor{WARule}{HTML}{B8B4D2}
\setlength{\tabcolsep}{10pt}
\renewcommand{\arraystretch}{1.1}
\arrayrulecolor{WARule}
\begin{tabular}{lrrrrrr}
\toprule
\rowcolor{WAHeader}
\textbf{Model} & \textbf{Arena Elo} $\uparrow$ & \textbf{W1} $\uparrow$ & \textbf{W2} $\uparrow$ & \textbf{W3} $\uparrow$ & \textbf{W4} $\uparrow$ & \textbf{W5} $\uparrow$ \\
\midrule
GPT-6-Astra & 1252 & 63.72 & 65.43 & 68.80 & 82.09 & 63.10 \\
\rowcolor{WAAlternate}
Marble & 1308 & 79.10 & 64.99 & 59.89 & 53.51 & 45.44 \\
HYWorld-2.0 & 1230 & 76.30 & 67.71 & 57.99 & 59.97 & 43.16 \\
\rowcolor{WAAlternate}
HYWorld-1.0 & 1029 & 78.78 & 59.40 & 55.19 & 66.16 & - \\
Matrix-3D & 1019 & 46.35 & 56.35 & 52.73 & 49.76 & 35.15 \\
\rowcolor{WAAlternate}
WorldGen & 962 & 69.17 & 60.11 & 54.16 & 57.56 & - \\
FlashWorld & 743 & 49.39 & 0.00 & 38.13 & - & - \\
\rowcolor{WAAlternate}
Lyra 2.0 & 729 & 57.58 & 0.90 & 40.52 & - & - \\
Lyra 1.0 & 728 & 41.10 & 1.36 & 37.07 & - & - \\
\bottomrule
\end{tabular}
\arrayrulecolor{black}
\endgroup
\end{table}

\paragraph{Overall performance across W1--W5.}
Table~\ref{tab:recon-world-summary} shows that different models lead
different capability groups.
After accounting for unsupported capabilities, Marble achieves the
highest Arena Elo of 1308, followed by GPT-6-Astra at 1252 and
HYWorld-2.0 at 1230; details of this adjustment are provided in appendix \S\ref{app:elo-calculate}.
Across individual capability levels, Marble and HYWorld-1.0 achieve
the highest W1 scores, while HYWorld-2.0 leads W2.
GPT-6-Astra leads W3--W5 despite lower W1 performance,
showing that strong visual quality does not necessarily translate
into reliable navigation, consistency, editing, or expansion.
FlashWorld and the Lyra variants further illustrate this gap:
they produce renderable scenes but obtain W2 scores of only 0.00--1.36.
These results highlight the importance of evaluating
both visual quality and the ability to navigate, maintain, edit, and expand a world.

\begin{table}[H]
\centering
\caption{Observable quality, input alignment, and structural coverage (W1).}
\label{tab:recon-quality-results}
\begingroup
\fontsize{8.2}{9.6}\selectfont
\setlength{\tabcolsep}{1.2pt}
\renewcommand{\arraystretch}{1.12}
\arrayrulecolor{WARule}
\begin{tabular*}{\textwidth}{@{\extracolsep{\fill}}l*{10}{r}@{}}
\toprule
\rowcolor{WAHeader}
 & \multicolumn{6}{c}{\cellcolor{WAHeader}\WAgroup{Appearance quality}} & \multicolumn{4}{c}{\cellcolor{WAHeader}\WAgroup{Coverage and alignment}} \\
\cmidrule(lr){2-7}\cmidrule(lr){8-11}
\rowcolor{WAHeader}
\textbf{Model} & \makecell{\textbf{TOPIQ} $\uparrow$} & \makecell{\textbf{AES} $\uparrow$} & \makecell{\textbf{PI} $\uparrow$} & \makecell{\textbf{Q-Align} $\uparrow$} & \makecell{\textbf{HPS} $\uparrow$} & \makecell{\textbf{VLM} $\uparrow$} & \makecell{\textbf{Hole-free}\\\textbf{ratio} $\uparrow$} & \makecell{\textbf{CLIP-T} $\uparrow$} & \makecell{\textbf{CLIP-I} $\uparrow$} & \makecell{\textbf{Object}\\\textbf{coverage} $\uparrow$} \\
\midrule
GPT-6-Astra & \WAcell{\textbf{57.74}} & \WAcell{40.26} & \WAcell{18.33} & \WAcell{60.29} & \WAcell{48.48} & \WAcell{88.48} & \WAcell{95.24} & \WAcell{61.19} & \WAcell{85.99} & \WAcell{81.20} \\
\rowcolor{WAAlternate}
Marble & \WAcell{56.08} & \WAcell{48.30} & \WAcell{81.36} & \WAcell{\textbf{82.77}} & \WAcell{\textbf{88.22}} & \WAcell{\textbf{96.67}} & \WAcell{\textbf{100.00}} & \WAcell{62.05} & \WAcell{92.04} & \WAcell{83.46} \\
HYWorld-2.0 & \WAcell{42.82} & \WAcell{46.76} & \WAcell{73.70} & \WAcell{78.73} & \WAcell{80.81} & \WAcell{95.47} & \WAcell{99.95} & \WAcell{62.23} & \WAcell{93.85} & \WAcell{88.72} \\
\rowcolor{WAAlternate}
HYWorld-1.0 & \WAcell{53.78} & \WAcell{\textbf{48.71}} & \WAcell{\textbf{83.88}} & \WAcell{82.28} & \WAcell{78.89} & \WAcell{92.79} & \WAcell{\textbf{100.00}} & \WAcell{62.16} & \WAcell{\textbf{95.63}} & \WAcell{89.66} \\
Matrix-3D & \WAcell{16.04} & \WAcell{33.77} & \WAcell{20.18} & \WAcell{22.12} & \WAcell{24.55} & \WAcell{22.93} & \WAcell{99.76} & \WAcell{61.14} & \WAcell{87.03} & \WAcell{75.94} \\
\rowcolor{WAAlternate}
WorldGen & \WAcell{32.29} & \WAcell{43.96} & \WAcell{71.32} & \WAcell{57.94} & \WAcell{66.90} & \WAcell{75.78} & \WAcell{99.58} & \WAcell{\textbf{62.29}} & \WAcell{94.42} & \WAcell{87.22} \\
FlashWorld & \WAcell{44.84} & \WAcell{40.84} & \WAcell{\textemdash} & \WAcell{38.50} & \WAcell{12.15} & \WAcell{37.97} & \WAcell{25.61} & \WAcell{59.68} & \WAcell{94.66} & \WAcell{\textbf{90.23}} \\
\rowcolor{WAAlternate}
Lyra 2.0 & \WAcell{36.11} & \WAcell{40.45} & \WAcell{69.13} & \WAcell{45.22} & \WAcell{50.31} & \WAcell{30.80} & \WAcell{71.74} & \WAcell{61.39} & \WAcell{89.06} & \WAcell{81.58} \\
Lyra 1.0 & \WAcell{39.92} & \WAcell{41.12} & \WAcell{0.00} & \WAcell{34.35} & \WAcell{6.23} & \WAcell{34.58} & \WAcell{18.32} & \WAcell{59.21} & \WAcell{91.71} & \WAcell{85.53} \\
\bottomrule
\end{tabular*}
\arrayrulecolor{black}
\endgroup
\end{table}

\paragraph{W1: Observable quality.}
Table~\ref{tab:recon-quality-results} compares visual quality using
360-degree videos rendered at the initial camera position. This
setting provides a favorable view of rendering quality, as the camera
does not move into newly exposed regions.
Marble and HYWorld-1.0 achieve similar leading W1 scores.
HYWorld-1.0 lifts a generated panorama into a mesh, allowing it to
retain the panorama's visual detail when viewed from its center.
Marble uses a 3D Gaussian representation and leads Q-Align, HPS,
and VLM, indicating strong perceptual quality in the rendered views.
GPT-6-Astra instead constructs meshes directly through Blender.
It achieves the highest TOPIQ score but lower PI and HPS scores.
A possible explanation is that its scenes resemble synthetic environments, with clean surfaces and well-defined shapes but limited geometric detail and material realism.

\begin{table}[H]
\centering
\caption{Physical usability for navigation and support (W2).}
\label{tab:recon-physical-results}
\begingroup
\fontsize{8.2}{9.6}\selectfont
\setlength{\tabcolsep}{1.2pt}
\renewcommand{\arraystretch}{1.12}
\arrayrulecolor{WARule}
\begin{tabular*}{0.6\textwidth}{@{\extracolsep{\fill}}l*{2}{r}@{}}
\toprule
\rowcolor{WAHeader}
 & \multicolumn{2}{c}{\cellcolor{WAHeader}\WAgroup{Physical usability}} \\
\cmidrule(lr){2-3}
\rowcolor{WAHeader}
\textbf{Model} & \makecell{\textbf{Navigable ratio} $\uparrow$} & \makecell{\textbf{Placement rate} $\uparrow$} \\
\midrule
GPT-6-Astra & \WAcell{\textbf{69.05}} & \WAcell{61.81} \\
\rowcolor{WAAlternate}
Marble & \WAcell{65.17} & \WAcell{64.81} \\
HYWorld-2.0 & \WAcell{65.27} & \WAcell{\textbf{70.14}} \\
\rowcolor{WAAlternate}
HYWorld-1.0 & \WAcell{51.90} & \WAcell{66.90} \\
Matrix-3D & \WAcell{64.09} & \WAcell{48.61} \\
\rowcolor{WAAlternate}
WorldGen & \WAcell{68.83} & \WAcell{51.39} \\
FlashWorld & \WAcell{0.00} & \WAcell{0.00} \\
\rowcolor{WAAlternate}
Lyra 2.0 & \WAcell{1.79} & \WAcell{0.00} \\
Lyra 1.0 & \WAcell{2.72} & \WAcell{0.00} \\
\bottomrule
\end{tabular*}
\arrayrulecolor{black}
\endgroup
\end{table}

\paragraph{W2: Physical usability.}
Table~\ref{tab:recon-physical-results} evaluates exported geometry through navigation and plate placement, moving beyond the rendered appearance assessed in W1. FlashWorld and Lyra variants produce renderable scenes but provide almost no navigable area and no successful placements under this protocol, as their geometry is incomplete and the hole rate is high. Among the remaining systems, navigation and support do not improve uniformly: WorldGen approaches the highest navigable ratio but falls substantially behind HYWorld-2.0 in placement rate. Thus, geometry that supports traversal does not necessarily provide dependable object-support surfaces. HYWorld-2.0 achieves the highest placement rate at 70.14\%, but the remaining failures indicate that even the strongest evaluated system does not yet provide consistently reliable support.

\begin{table}[H]
\centering
\caption{State persistence (W3).}
\label{tab:recon-consistency-results}
\label{tab:recon-exploration-results}
\label{tab:recon-persistence-results}
\begingroup
\fontsize{8.2}{9.6}\selectfont
\setlength{\tabcolsep}{1.2pt}
\renewcommand{\arraystretch}{1.12}
\arrayrulecolor{WARule}
\begin{tabular*}{\textwidth}{@{\extracolsep{\fill}}l*{7}{r}@{}}
\toprule
\rowcolor{WAHeader}
 & \multicolumn{4}{c}{\cellcolor{WAHeader}\WAgroup{Scene-level}} & \multicolumn{3}{c}{\cellcolor{WAHeader}\WAgroup{Object-level}} \\
\cmidrule(lr){2-5}\cmidrule(lr){6-8}
\rowcolor{WAHeader}
\textbf{Model} & \makecell{\textbf{In-bounds}\\\textbf{ratio} $\uparrow$} & \makecell{\textbf{Brightness}\\\textbf{consistency} $\uparrow$} & \makecell{\textbf{Hue}\\\textbf{consistency} $\uparrow$} & \makecell{\textbf{Location}\\\textbf{HPS F1} $\uparrow$} & \makecell{\textbf{HPS F1} $\uparrow$} & \makecell{\textbf{Track F1} $\uparrow$} & \makecell{\textbf{VLM F1} $\uparrow$} \\
\midrule
GPT-6-Astra & \WAcell{\textbf{98.05}} & \WAcell{\textbf{90.01}} & \WAcell{\textbf{90.70}} & \WAcell{\textbf{62.04}} & \WAcell{\textbf{28.16}} & \WAcell{\textbf{47.34}} & \WAcell{\textbf{65.27}} \\
\rowcolor{WAAlternate}
Marble & \WAcell{96.62} & \WAcell{86.58} & \WAcell{88.70} & \WAcell{54.25} & \WAcell{24.83} & \WAcell{39.27} & \WAcell{28.96} \\
HYWorld-2.0 & \WAcell{94.84} & \WAcell{86.17} & \WAcell{88.21} & \WAcell{46.38} & \WAcell{26.60} & \WAcell{40.73} & \WAcell{22.97} \\
\rowcolor{WAAlternate}
HYWorld-1.0 & \WAcell{96.01} & \WAcell{80.04} & \WAcell{82.35} & \WAcell{39.79} & \WAcell{23.45} & \WAcell{42.40} & \WAcell{22.28} \\
Matrix-3D & \WAcell{94.55} & \WAcell{87.02} & \WAcell{87.94} & \WAcell{56.61} & \WAcell{18.17} & \WAcell{18.38} & \WAcell{6.44} \\
\rowcolor{WAAlternate}
WorldGen & \WAcell{87.70} & \WAcell{80.80} & \WAcell{83.82} & \WAcell{46.76} & \WAcell{21.59} & \WAcell{33.17} & \WAcell{25.30} \\
FlashWorld & \WAcell{13.30} & \WAcell{64.61} & \WAcell{68.18} & \WAcell{34.90} & \WAcell{21.60} & \WAcell{36.86} & \WAcell{27.47} \\
\rowcolor{WAAlternate}
Lyra 2.0 & \WAcell{29.58} & \WAcell{78.28} & \WAcell{78.58} & \WAcell{38.70} & \WAcell{17.25} & \WAcell{22.62} & \WAcell{18.66} \\
Lyra 1.0 & \WAcell{15.82} & \WAcell{68.38} & \WAcell{71.77} & \WAcell{20.47} & \WAcell{20.84} & \WAcell{36.60} & \WAcell{25.63} \\
\bottomrule
\end{tabular*}
\arrayrulecolor{black}
\endgroup
\end{table}

\paragraph{W3: Scene- and object-level persistence.}
Table~\ref{tab:recon-consistency-results} examines whether scenes and objects remain consistently observable across viewpoint changes. Matrix-3D maintains relatively stable scene-level brightness and hue, yet has the lowest Track F1 and VLM F1 scores. Its stable overall appearance therefore masks weaknesses in object correspondence and visible integrity, which matter when objects serve as landmarks during exploration. GPT-6-Astra presents a contrasting profile: it leads both scene-level metrics and all three object composites, despite its lower visual-quality scores in W1. Together, these results distinguish visual appeal, scene-level appearance stability, and object-level persistence: an attractive or stable-looking scene does not necessarily retain intact, identifiable objects across views.

\begin{table}[H]
\centering
\caption{Controlled editing, preservation, and joint edit quality (W4).}
\label{tab:recon-edit-results}
\begingroup
\fontsize{8.2}{9.6}\selectfont
\setlength{\tabcolsep}{1.2pt}
\renewcommand{\arraystretch}{1.12}
\arrayrulecolor{WARule}
\begin{tabular*}{\textwidth}{@{\extracolsep{\fill}}l*{7}{r}@{}}
\toprule
\rowcolor{WAHeader}
 & \multicolumn{1}{c}{\cellcolor{WAHeader}\WAgroup{Execution}} & \multicolumn{4}{c}{\cellcolor{WAHeader}\WAgroup{Object editing}} & \multicolumn{2}{c}{\cellcolor{WAHeader}\WAgroup{Global editing}} \\
\cmidrule(lr){2-2}\cmidrule(lr){3-6}\cmidrule(lr){7-8}
\rowcolor{WAHeader}
\textbf{Model} & \makecell{\textbf{Succ. rate} $\uparrow$} & \makecell{\textbf{BG RGB}\\\textbf{retention} $\uparrow$} & \makecell{\textbf{BG LPIPS}\\\textbf{similarity} $\uparrow$} & \makecell{\textbf{BG DINO}\\\textbf{similarity} $\uparrow$} & \makecell{\textbf{Object}\\\textbf{score} $\uparrow$} & \makecell{\textbf{Depth}\\\textbf{agreement} $\uparrow$} & \makecell{\textbf{Global}\\\textbf{score} $\uparrow$} \\
\midrule
GPT-6-Astra & \WAcell{60.00} & \WAcell{\textbf{100.00}} & \WAcell{\textbf{99.12}} & \WAcell{\textbf{99.82}} & \WAcell{\textbf{66.55}} & \WAcell{\textbf{99.57}} & \WAcell{\textbf{49.57}} \\
\rowcolor{WAAlternate}
Marble & \WAcell{56.67} & \WAcell{90.08} & \WAcell{7.86} & \WAcell{89.29} & \WAcell{40.50} & \WAcell{55.08} & \WAcell{35.08} \\
HYWorld-2.0 & \WAcell{\textbf{73.33}} & \WAcell{97.05} & \WAcell{55.42} & \WAcell{93.04} & \WAcell{50.86} & \WAcell{25.04} & \WAcell{25.04} \\
\rowcolor{WAAlternate}
HYWorld-1.0 & \WAcell{66.67} & \WAcell{99.61} & \WAcell{94.86} & \WAcell{99.27} & \WAcell{54.98} & \WAcell{24.34} & \WAcell{23.36} \\
Matrix-3D & \WAcell{40.00} & \WAcell{97.89} & \WAcell{43.73} & \WAcell{93.41} & \WAcell{19.90} & \WAcell{32.46} & \WAcell{20.93} \\
\rowcolor{WAAlternate}
WorldGen & \WAcell{56.67} & \WAcell{98.96} & \WAcell{89.30} & \WAcell{98.46} & \WAcell{32.65} & \WAcell{15.83} & \WAcell{11.03} \\
\bottomrule
\end{tabular*}
\arrayrulecolor{black}
\endgroup
\end{table}

\paragraph{W4: Controlled editing.}
W4 evaluates whether a requested edit changes the intended content while retaining the surrounding world. HYWorld-2.0 has the highest edit success rate, but GPT-6-Astra leads preservation and both joint edit scores. This difference shows why editability cannot be identified with instruction satisfaction alone: realizing a change and restricting its side effects are distinct requirements (Table~\ref{tab:recon-edit-results}).

\begin{table}[H]
\centering
\caption{Expansion and preservation scores (W5).}
\label{tab:recon-operation-results}
\begingroup
\fontsize{8.2}{9.6}\selectfont
\setlength{\tabcolsep}{1.2pt}
\renewcommand{\arraystretch}{1.12}
\arrayrulecolor{WARule}
\begin{tabular*}{\textwidth}{@{\extracolsep{\fill}}l*{5}{r}@{}}
\toprule
\rowcolor{WAHeader}
 & \multicolumn{2}{c}{\cellcolor{WAHeader}\WAgroup{Growth and coverage}} & \multicolumn{1}{c}{\cellcolor{WAHeader}\WAgroup{New region}} & \multicolumn{2}{c}{\cellcolor{WAHeader}\WAgroup{Preservation}} \\
\cmidrule(lr){2-3}\cmidrule(lr){4-4}\cmidrule(lr){5-6}
\rowcolor{WAHeader}
\textbf{Model} & \makecell{\textbf{Area}\\\textbf{score} $\uparrow$} & \makecell{\textbf{Junction}\\\textbf{score} $\uparrow$} & \makecell{\textbf{New-region}\\\textbf{quality} $\uparrow$} & \makecell{\textbf{Old-region}\\\textbf{preservation} $\uparrow$} & \makecell{\textbf{Path}\\\textbf{preservation} $\uparrow$} \\
\midrule
GPT-6-Astra & \WAcell{\textbf{39.71}} & \WAcell{\textbf{67.48}} & \WAcell{72.56} & \WAcell{\textbf{52.54}} & \WAcell{\textbf{83.20}} \\
\rowcolor{WAAlternate}
Marble & \WAcell{17.48} & \WAcell{55.33} & \WAcell{\textbf{74.00}} & \WAcell{28.28} & \WAcell{52.12} \\
HYWorld-2.0 & \WAcell{5.70} & \WAcell{52.64} & \WAcell{55.97} & \WAcell{49.17} & \WAcell{52.30} \\
\rowcolor{WAAlternate}
Matrix-3D & \WAcell{29.68} & \WAcell{52.05} & \WAcell{22.78} & \WAcell{22.57} & \WAcell{48.65} \\
\bottomrule
\end{tabular*}
\arrayrulecolor{black}
\endgroup
\end{table}

\paragraph{W5: Expansion with preservation.}
Table~\ref{tab:recon-operation-results} evaluates whether expansion adds new space while retaining the quality of existing regions. Matrix-3D achieves a higher area score than HYWorld-2.0 and Marble, but has the lowest new-region quality and old-region preservation scores. Its larger expansion therefore does not translate into a more convincing continuation. GPT-6-Astra provides a contrasting profile, leading in spatial growth, junction improvement, and both preservation scores, although Marble produces the highest-quality new views. These results show that expansion quality depends not simply on how much space is added, but on whether new regions are convincing and previously available regions retain their quality.

\FloatBarrier

\section{Embodied World Model Track}
\label{sec:embodied}

We propose an evaluation framework for embodied world models that predict, from
an egocentric robot viewpoint, how the robot, its surrounding scene, and the
manipulated objects evolve in response to an action. The framework has two
orthogonal axes: a \textbf{capability axis} W1--W6, where each level represents a
qualitative capability transition, and a \textbf{scoring axis} comprising
perception, consistency, causality, and controllability. We instantiate the
embodied track with video models and evaluate their action-conditioned visual
predictions on W2--W4. The current action interface uses natural-language
prompts and is compatible with a future extension to discrete robot actions.

Concretely, each embodied case is formulated as an image-conditioned,
prompt-controlled video generation problem. Let $I_0$ denote the prescribed
reference image, an egocentric observation of the scene before the action. Let
$p$ denote the action prompt, and let $Y_{1:T}$ denote the generated video. The
prompt $p$ specifies the complete action condition, including the acting
subject, target, action type, direction, and, for multi-step tasks, the ordered
stages and their corresponding time intervals. The model then generates the
video according to
\begin{equation}
    Y_{1:T} = G_{\theta}(I_0, p).
    \label{eq:embodied-video-generation}
\end{equation}

Within the embodied track, we evaluate three progressively challenging levels
of the shared W1--W6 world capability axis. W2 evaluates the immediate response
to a single atomic robot action, including movement and manipulation. W3
evaluates ordered multi-step rollouts with persistent state, covering
movement-only, manipulation-only, and combined movement--manipulation
sequences. W4 evaluates paired rollouts from a shared initial state under
either a changed action condition or a changed physical rule, thereby testing
whether the model responds to the edited condition while preserving
unaffected content.

The scoring axis quantifies every level along four complementary dimensions:
\emph{perception}, whether the scene, robot, and relevant objects remain
recognizable under the prescribed input; \emph{consistency}, whether
appearance, geometry, state, and temporal continuity are maintained across the
rollout; \emph{causality}, whether contact, event order, and state changes
follow plausible physical and semantic causes; and \emph{controllability},
whether the specified action is faithfully followed, including the required
action process and resulting state changes. Keeping the four separate 
allows the framework to discriminate failure modes that a holistic judgment
would conflate: a video can reach a plausible terminal state while omitting the
grasp, displacing the object in the wrong direction, or altering the scene
without a visible trigger.

\subsection{Data Construction}
\label{sec:embodied-data}

The present data construction and formal evaluation focus on W2--W4.
W1, W5, and W6 remain shared framework-level capability definitions and are
not operationalized in the current embodied benchmark.

\paragraph{Task Representation and Annotations}

Each case is represented as a structured task record identified by a unique
sample ID and containing the task level, any required initial-frame condition,
and an action prompt. The action prompt combines the observation viewpoint from
the robot's head-mounted camera with the complete action condition. Each record
also includes annotated environment and subject descriptions, which provide
structured context for auditing and for compiling the VLM assertions used in
the subsequent evaluation. W4 records additionally contain a shared
paired-condition group, a branch ID, and the branch-specific action or
physical-rule condition.

\paragraph{Capability-Guided Case Design}

The W2--W4 data use an image-to-video format with an explicit initial-frame
condition. Each case provides a reference image that fixes the initial state,
followed by a visual rollout in which the action-conditioned state transition
is assessed.

W2 isolates the response to a single atomic robot action. Each case specifies
the acting subject, target, direction, and expected post-action state. The
evaluation separately verifies whether the prescribed action occurs and whether
it produces the intended visible state change, while the atomic action scope
keeps the source of that change well defined.

W3 extends the same interface to longer-horizon action sequences. Each case
specifies an ordered set of actions or stages with their temporal placement,
enabling the evaluation to jointly assess action order, the persistence of
robot, object, and scene states, and the accumulation of effects over time.

W4 uses paired condition tests from a shared initial state and covers two
intervention types. Action-condition variants change properties such as the
motion direction, displacement magnitude, target, or action type. Physical-rule
variants change properties such as gravity, friction, collision constraints, or
material rigidity. Both branches in a group share the same reference image and
are packaged as a synchronized pair, holding the initial scene constant so that
the paired design isolates the effect of the intervention.

\paragraph{Dataset Statistics}
\label{sec:embodied-statistics}

The formal evaluation set contains 254 task instances spanning W2--W4. Each
instance is paired with a reference image and evaluated under a common rollout
protocol. The composition of the evaluation set is
summarized in Table~\ref{tab:embodied-dataset}.

All generated videos are converted to a common evaluation format before
scoring. W2 and W4 use a 5.0\,s rollout, while W3 uses a 12.0\,s rollout;
videos are standardized to 24 FPS, corresponding to 120 frames for W2/W4 and
288 frames for W3. Both automatic metrics and assertion-based VLM evaluation
operate on a 4-FPS sampling of these rollouts, yielding 20 frames for W2/W4
and 48 frames for W3. Automatic metrics are computed over the corresponding
frame sequence, while VLM assertions are evaluated at fixed checkpoints within
it. Spatial dimensions and encoding are normalized to the evaluation
specification, so differences in native output formats do not affect metric
computation. This normalization preserves the prescribed visual evolution
while making outputs comparable.

The action vocabulary covers a range of embodied skills, including grasping,
lifting, opening, moving, pushing, placing, releasing, pressing, pulling,
folding, inserting, wiping, sliding, turning, carrying, tapping, touching,
stacking, and cutting. Together, these actions cover contact-based manipulation,
object displacement, and observable changes in object state.

\begin{table}[t]
    \centering\small
    \setlength{\tabcolsep}{4pt}
    \renewcommand{\arraystretch}{1.1}
    \caption{Composition of the formal evaluation set in the Embodied World
    Model Track. W4 cases are organized as matched branches from a shared
    initial state.}
    \label{tab:embodied-dataset}
    \arrayrulecolor{WARule}
    \begin{tabularx}{\linewidth}{>{\raggedright\arraybackslash}p{0.08\linewidth}c>{\raggedright\arraybackslash}p{0.14\linewidth}>{\raggedright\arraybackslash}X}
        \toprule
        \rowcolor{WAHeader}
        Level & Cases & Rollout & Task composition \\
        \midrule
        W2 & 62 & 5\,s
        & Single-step robot actions with specified subjects, targets, directions, and post-action states. \\
        W3 & 100 & 12\,s
        & Ordered multi-step rollouts with explicit temporal stages and persistent robot, object, and scene states. \\
        W4 & 92 & 5\,s per branch
        & 40 action-condition pairs and 52 physical-rule pairs, each generated from one shared initial frame. \\
        \midrule
        Total & 254 & ---
        & 254 reference images and 2{,}768 rollout records. \\
        \bottomrule
    \end{tabularx}
    \arrayrulecolor{black}
\end{table}
\subsection{Evaluation Metrics}
\label{sec:embodied-metrics}

The embodied evaluation score is organized along four complementary dimensions:
\emph{Perception}, \emph{Consistency}, \emph{Causality}, and
\emph{Controllability}. Perception assesses whether the scene and robot remain
recognizable; consistency assesses whether appearance, geometry, state, and
temporal continuity persist over the rollout; causality assesses whether
interactions and state transitions follow plausible physical and temporal
causes; and controllability assesses whether the specified action and its
consequences are realized. Separating these dimensions exposes distinct failure
modes that would otherwise be conflated by a single holistic score.

The evaluation combines automatic frame-based metrics with assertion-based VLM
metrics. Automatic metrics produce normalized continuous scores from sampled
frames, whereas VLM metrics evaluate frozen task-specific assertions using a
fixed protocol. Binary assertion judgments are combined with predefined
importance weights, and inapplicable assertions are excluded. The scoring axis
contains 17 metrics: four for Perception, five for Consistency, three for
Causality, and five for Controllability. Perception, Consistency, and Causality
are evaluated at all three levels, whereas the applicable Controllability
metrics vary by level. Their evidence and applicability across W2--W4 are summarized in
Table~\ref{tab:embodied-metric-levels}, which lists the metrics contributing
to each level's score and the evidence used by each metric. Each case is evaluated from the
reference image $I_0$ and generated video $Y_{1:T}$.

\paragraph{Perception.} Scene Fidelity (SF) checks whether persistent scene
facts remain visible and recognizable. Subject Fidelity (SuF) checks whether
the robot retains its specified identity, appearance, and structural attributes.
Perceptual Quality (PQ) measures frame-level image quality using MUSIQ-SPAQ,
and Sharpness Retention (SR) measures the preservation of visual detail relative
to the first generated frame.

\paragraph{Consistency.} Subject Appearance Consistency (SAC) measures the
stability of the robot's visual identity using features extracted from tracked
subject masks. Geometric Consistency (GC) measures the stability of static scene
geometry. Scene--State Consistency (SSC) checks scene and object states at fixed
temporal checkpoints. Temporal Coherence (TC) measures residual frame changes
after motion compensation, while Motion Smoothness (MS) measures motion
continuity through intermediate-frame prediction.

\paragraph{Causality.} Physical Plausibility (PP) checks whether motion and
interaction satisfy task-relevant physical constraints. Temporal and Causal
Order (TCO) checks whether specified causes precede their effects, and Causal
Trigger Validity (CTV) checks whether visible state changes are supported by
their prescribed triggers.

\paragraph{Controllability.} Goal-State Achievement (GSA) checks whether the
requested endpoint is reached. Atomic Action Compliance (AAC) checks whether
the specified action is performed by the correct subject on the correct target.
Functional Interaction Success (FIS) checks the required task-specific effect.
Multi-Action Ordering (MAO) checks the order of prescribed multi-step actions,
and Condition-Branch Fidelity (CBF) checks whether paired W4 branches produce
the required difference under changed action conditions or physical rules.

Each metric is first aggregated across the applicable task instances, except for CBF, which is evaluated at the matched-pair level in W4, with each matched pair contributing one group score. Within each dimension, the scores of all applicable metrics are then averaged, with inapplicable metrics excluded rather than treated as zero. The four dimension scores are finally combined using fixed weights of 20\% for Perception, 20\% for Consistency, 30\% for Causality, and 30\% for Controllability.

\subsection{Experimental Results}
\label{sec:embodied-results}

\paragraph{Experimental setup.} We evaluate eight candidate video models under
a common W2--W4 protocol. All models are evaluated on the same task cases, using
the same reference images and action prompts. Their generated rollouts are
evaluated with the same metrics and frozen assertions specified by the
evaluation protocol. The evaluated set comprises eight models:
MiniMax-H3~\citep{minimax2026h3},
Cosmos 3~\citep{em_nvidia2025cosmos}, Seedance 2.5~\citep{seed2026seedance25},
Grok Imagine Video 1.5~\citep{xai2026grokvideo15}, HappyHorse 1.1~\citep{happyhorse2026i2v},
Kling 3~\citep{kuaishou2026kling30}, Sora 2~\citep{openai2025sora2systemcard}, and
Wan 3.0~\citep{alibaba2026wan30}. All metric and capability scores are
normalized to $[0,100]$, with higher values indicating better performance.
Alongside this protocol, the same eight models are compared in a blind pairwise
human-preference arena: for a shared case, raters view two anonymized rollouts
and select the better one, and the collected votes are aggregated into an Arena
ELO rating that is reported together with the level scores.

\paragraph{Overall capability evaluation and human preference.}
Table~\ref{tab:embodied-summary} summarizes the overall capability score
of each model at W2--W4 together with its Arena ELO. The per-level results
in Tables~\ref{tab:embodied-w2-scores}--\ref{tab:embodied-w4-scores}
use the same model and metric definitions throughout. MiniMax-H3 achieves
the highest overall capability score at all three levels and also ranks
first by Arena ELO. In contrast, Cosmos 3 and Sora 2 remain below 75 in overall
capability at all three levels. The Arena ELO broadly follows the ordering
established by the automatic evaluation, with differences mainly limited
to local ordering among closely performing models, and provides a
complementary assessment based on holistic human preference.

\begin{table}[H]
  \centering
  \small
  \setlength{\tabcolsep}{10pt}
  \renewcommand{\arraystretch}{1.1}
  \caption{Overall capability and human preference. Overall capability scores
at W2--W4 and human-preference Arena ELO ratings for the eight evaluated
models. The W2, W3, and W4 columns correspond to the Overall scores in
Tables~\ref{tab:embodied-w2-scores}--\ref{tab:embodied-w4-scores},
respectively. Arena ELO is estimated from 62,592 blind pairwise votes.
Rows are ordered by Arena ELO.}
  \label{tab:embodied-summary}
  \arrayrulecolor{WARule}
  \begin{tabular}{lrrrr}
    \toprule
    \rowcolor{WAHeader}
    \textbf{Model} & \textbf{Arena ELO $\uparrow$} & \textbf{W2 $\uparrow$} & \textbf{W3 $\uparrow$} & \textbf{W4 $\uparrow$} \\
    \midrule
    MiniMax-H3 & 1060 & 93.60 & 91.42 & 88.62 \\
    \rowcolor{WAAlternate}
    Grok Imagine Video 1.5 & 1051 & 93.02 & 88.51 & 83.48 \\
    Wan 3.0 & 1051 & 90.53 & 87.96 & 84.53 \\
    \rowcolor{WAAlternate}
    Seedance 2.5 & 1031 & 90.91 & 86.73 & 84.11 \\
    HappyHorse 1.1 & 1026 & 91.59 & 88.82 & 84.60 \\
    \rowcolor{WAAlternate}
    Kling 3 & 978 & 85.94 & 84.26 & 78.70 \\
    Cosmos 3 & 931 & 69.16 & 74.47 & 68.33 \\
    \rowcolor{WAAlternate}
    Sora 2 & 872 & 60.91 & 65.35 & 64.12 \\
    \bottomrule
  \end{tabular}
  \arrayrulecolor{black}
\end{table}

\paragraph{W2: single-action execution.} W2 evaluates whether a
model can execute a single atomic action and produce the prescribed contact
event, object response, and terminal state. At this level, perception is
generally not the primary bottleneck: scene fidelity, subject fidelity, and
frame-level perceptual quality remain high across models. The dominant
failures instead arise from a breakdown in the causal coupling between the
commanded action, the contact event, and the resulting object response, which
we summarize
through three representative phenomena, illustrated in order by the rows of
Fig.~\ref{fig:embodied-w2-failures}. First, the prescribed contact is never
established: the manipulator approaches the target but only repeats grasp-like
gestures in empty space without touching the prescribed contact points, so the
rollout shows an action-like gesture without either the prescribed contact or
its effect (Fig.~\ref{fig:embodied-w2-failures}(a)). Second, the interaction is
misdirected to another object: the manipulator performs the commanded action
on an entity other than the prescribed target, so the prescribed target never
responds (Fig.~\ref{fig:embodied-w2-failures}(b)). Third, the prescribed
interaction is disrupted by uncaused changes to the scene or the observation:
objects absent from the initial observation appear mid-rollout, or the
viewpoint departs from the prescribed fixed camera, so the world state and
observation conditions established by the initial frame are not preserved
(Fig.~\ref{fig:embodied-w2-failures}(c)). Beyond these explicit
failures, some rollouts exhibit subtler action--effect decoupling: the motion
and object response look plausible overall, yet the prescribed terminal
condition is violated, as when a lifted object is released outside its target
container or a slid object comes to rest in an unstable pose. Taken together, these
observations indicate that W2 requires more than plausible action appearance:
a successful rollout should establish a complete
action--contact--effect--terminal-state chain, in which the commanded action
corresponds to a visible contact event, the interaction produces a consistent
object response, and the target ultimately reaches the prescribed terminal
state.

\begin{figure}[t]
  \centering
  \includegraphics[width=\textwidth]{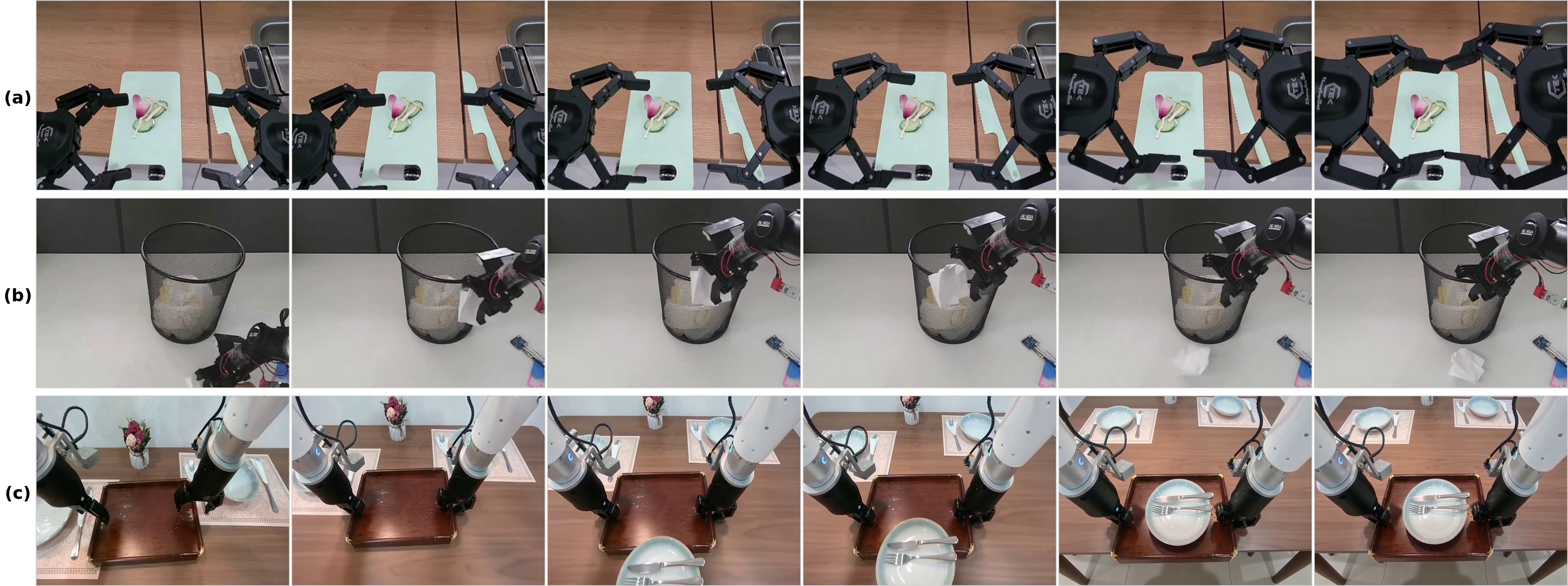}
  \caption{Representative W2 failure rollouts; each row pairs one action
  prompt with six key frames of the corresponding generated video.
  (a) Prescribed contact never established. Prompt: both grippers grasp the
  side edges of the cutting board and lift it vertically while keeping it
  level. Generated: the grippers repeatedly open and close in the air beside
  the board without ever grasping its side edges, and the board never leaves
  the table. (b) Interaction misdirected to an unintended object. Prompt: the right gripper grasps the crumpled
  paper inside the mesh bin on the floor and lifts it out of the bin.
  Generated: the gripper instead grasps a tissue lying on the floor beside
  the bin, lifts it above the bin, and releases it back onto the floor, while
  the prescribed paper never leaves the bin.
  (c) Uncaused scene and viewpoint changes. Prompt: push the tray and the items on it forward across the table.
  Generated: the initially empty tray acquires a plate and cutlery that were
  never present, and the viewpoint jumps away from the fixed camera
.}
  \label{fig:embodied-w2-failures}
\end{figure}

\begin{table}[H]
\centering
\caption{W2 embodied scores: single-action execution. Scores are averages
over the 62 formal W2 manipulation cases.}
\label{tab:embodied-w2-scores}
\begingroup
\fontsize{7.6}{8.9}\selectfont
\setlength{\tabcolsep}{1.0pt}
\renewcommand{\arraystretch}{1.12}
\arrayrulecolor{WARule}
\begin{tabular*}{\textwidth}{@{\extracolsep{\fill}}l*{15}{r}@{}}
\toprule
\rowcolor{WAHeader}
\textbf{Model}
& \multicolumn{4}{c}{\cellcolor{WAHeader}\WAgroup{Perception}}
& \multicolumn{5}{c}{\cellcolor{WAHeader}\WAgroup{Consistency}}
& \multicolumn{3}{c}{\cellcolor{WAHeader}\WAgroup{Causality}}
& \multicolumn{3}{c}{\cellcolor{WAHeader}\WAgroup{Controllability}} \\
\cmidrule(lr){2-5}
\cmidrule(lr){6-10}
\cmidrule(lr){11-13}
\cmidrule(lr){14-16}
\rowcolor{WAHeader}
&
\textbf{SF} & \textbf{SuF} & \textbf{PQ} & \textbf{SR}
&
\textbf{SAC} & \textbf{GC} & \textbf{SSC} & \textbf{TC} & \textbf{MS}
&
\textbf{PP} & \textbf{TCO} & \textbf{CTV}
&
\textbf{GSA} & \textbf{AAC} & \textbf{FIS} \\
\midrule

MiniMax-H3
& \WAcell{98.19} & \WAcell{98.39} & \WAcell{71.71} & \WAcell{97.62}
& \WAcell{84.16} & \WAcell{71.57} & \WAcell{98.26} & \WAcell{84.33} & \WAcell{84.55}
& \WAcell{100.00} & \WAcell{96.77} & \WAcell{96.77}
& \WAcell{95.16} & \WAcell{98.39} & \WAcell{96.77} \\

\rowcolor{WAAlternate}
Grok Imagine Video 1.5
& \WAcell{97.96} & \WAcell{98.79} & \WAcell{73.32} & \WAcell{98.37}
& \WAcell{89.53} & \WAcell{70.52} & \WAcell{98.42} & \WAcell{79.54} & \WAcell{80.77}
& \WAcell{94.62} & \WAcell{96.77} & \WAcell{96.77}
& \WAcell{95.16} & \WAcell{98.39} & \WAcell{96.77} \\

HappyHorse 1.1
& \WAcell{98.19} & \WAcell{98.59} & \WAcell{70.45} & \WAcell{94.72}
& \WAcell{88.38} & \WAcell{69.57} & \WAcell{98.28} & \WAcell{77.43} & \WAcell{79.02}
& \WAcell{98.92} & \WAcell{95.16} & \WAcell{95.16}
& \WAcell{91.94} & \WAcell{93.55} & \WAcell{95.16} \\

\rowcolor{WAAlternate}
Seedance 2.5
& \WAcell{98.19} & \WAcell{98.79} & \WAcell{70.22} & \WAcell{95.96}
& \WAcell{90.03} & \WAcell{74.93} & \WAcell{98.40} & \WAcell{88.72} & \WAcell{89.09}
& \WAcell{97.85} & \WAcell{90.32} & \WAcell{90.32}
& \WAcell{88.71} & \WAcell{91.94} & \WAcell{91.94} \\

Wan 3.0
& \WAcell{98.19} & \WAcell{97.58} & \WAcell{72.13} & \WAcell{98.82}
& \WAcell{82.24} & \WAcell{64.17} & \WAcell{98.25} & \WAcell{77.36} & \WAcell{77.52}
& \WAcell{98.39} & \WAcell{95.16} & \WAcell{92.74}
& \WAcell{90.32} & \WAcell{91.94} & \WAcell{93.55} \\

\rowcolor{WAAlternate}
Kling 3
& \WAcell{97.96} & \WAcell{98.79} & \WAcell{71.40} & \WAcell{99.10}
& \WAcell{90.02} & \WAcell{71.34} & \WAcell{97.46} & \WAcell{79.39} & \WAcell{76.43}
& \WAcell{97.85} & \WAcell{85.48} & \WAcell{84.68}
& \WAcell{75.81} & \WAcell{79.03} & \WAcell{87.10} \\

Cosmos 3
& \WAcell{98.19} & \WAcell{98.99} & \WAcell{69.68} & \WAcell{97.91}
& \WAcell{84.78} & \WAcell{78.27} & \WAcell{98.64} & \WAcell{85.48} & \WAcell{86.39}
& \WAcell{99.46} & \WAcell{45.16} & \WAcell{45.97}
& \WAcell{50.00} & \WAcell{46.77} & \WAcell{48.39} \\

\rowcolor{WAAlternate}
Sora 2
& \WAcell{89.56} & \WAcell{89.52} & \WAcell{64.97} & \WAcell{95.06}
& \WAcell{88.40} & \WAcell{66.82} & \WAcell{88.23} & \WAcell{83.81} & \WAcell{84.89}
& \WAcell{90.86} & \WAcell{40.32} & \WAcell{37.10}
& \WAcell{37.10} & \WAcell{30.65} & \WAcell{38.71} \\

\bottomrule
\end{tabular*}
\arrayrulecolor{black}
\endgroup
\end{table}

\begin{figure}[t]
  \centering
  \includegraphics[width=\textwidth]{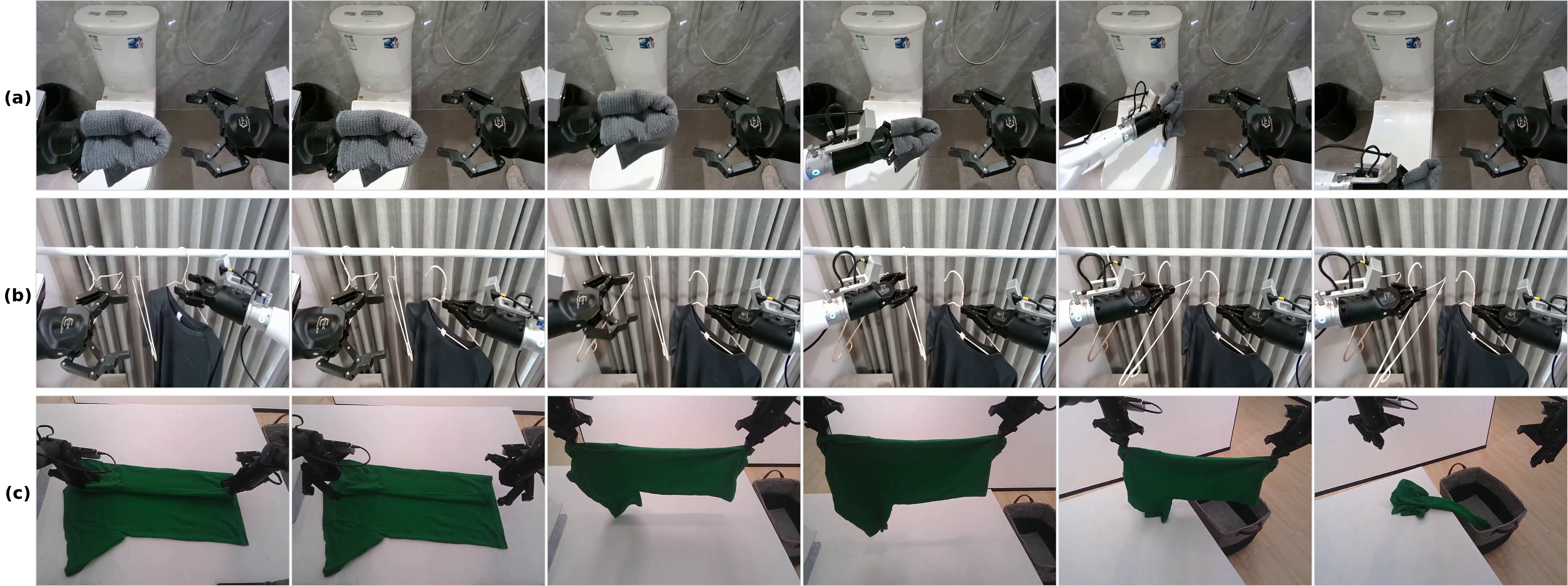}
  \caption{Representative W3 failure rollouts; each row pairs one action
  prompt with six key frames of the corresponding generated video.
  (a) Initial state never established. Prompt: open the toilet lid, grasp the towel, wipe the seat back and
  forth, and place the towel aside. Generated: the lid is never opened; the left arm nevertheless grasps the towel
  and performs the wiping motion over the closed lid, and the rollout ends with
  the towel still held rather than placed aside. (b) Intermediate state never transferred. Prompt: take the hanger with the shirt off the rod,
  translate it to the empty rod position on the left, and re-hang it before
  releasing. Generated: the right gripper takes the shirt hanger off the rod
  but keeps holding it instead of translating it to the empty rod position on
  the left, while the other gripper takes an empty wire hanger off the rod and
  holds it in mid-air; the prescribed re-hanging never occurs. (c) Terminal state never achieved. Prompt: both grippers grasp the
  two sides of the green T-shirt, lift and shake it open, then place it into the
  striped basket on the right of the table. Generated: the shirt is grasped,
  lifted, and shaken open, but the final release drops it onto the basket rim,
  and the rollout ends with the shirt bunched on the table and only one end
  trailing into the otherwise empty basket.}
  \label{fig:embodied-w3-failures}
\end{figure}

\paragraph{W3: multi-action sequences with persistent state.}
W3 evaluates whether a model can execute an ordered multi-stage action
sequence while keeping object identity and scene state persistent throughout
the generated rollout. At this level, perception remains largely intact: scene
fidelity, subject fidelity, and frame-level perceptual quality stay close to
their W2 values, and the global layout is preserved as the sequence unfolds.
Consistency, in contrast, degrades as the horizon lengthens, indicating that
longer action chains place higher demands on the persistence
and transfer of state across consecutive stages. The dominant failures concentrate on a
breakdown of state transfer between consecutive stages: a stage fails to
establish or preserve the state that the next stage presupposes, so later
actions lose their execution precondition. We observe three representative
phenomena, illustrated in order by the rows of
Fig.~\ref{fig:embodied-w3-failures}. First, the initial state fails to be
established, and the downstream chain loses its precondition: when the key state
change prescribed by the first stage does not occur, the later stages either are
skipped or proceed against the wrong state, so their effects no longer
correspond to the task (Fig.~\ref{fig:embodied-w3-failures}(a)). Second, an intermediate state fails
to be transferred to the following stages: even when a stage brings its object
into the expected state, the subsequent stages do not act on that state but
continue against an empty state or an unintended object, so the sequence remains
continuous in form while producing no accumulated object-state change
(Fig.~\ref{fig:embodied-w3-failures}(b)). Third, the terminal state prescribed
by the chain is never achieved: individual stages appear to be executed, yet the
object state they accumulate diverges from the prescribed one; the object may be
dropped during transport, released onto the rim of its container instead of into
it, deposited outside the prescribed container, or, for deformable targets, left
bunched rather than spread, hung, or inserted, so the rollout ends in a state
that violates the task condition
(Fig.~\ref{fig:embodied-w3-failures}(c)). Taken together, these phenomena
indicate that W3
requires more than a sequence of locally plausible actions: a successful
rollout should maintain a complete order--transfer--accumulation chain, in which
the stages occur in the prescribed order, the state established by each stage is
carried over as the precondition of the next, and the accumulated state
converges to the terminal condition prescribed by the task.

\begin{table}[H]
\centering
\caption{W3 embodied scores: multi-action sequences with persistent state. Scores are averages over the 100 W3 cases.}
\label{tab:embodied-w3-scores}
\begingroup
\fontsize{7.6}{8.9}\selectfont
\setlength{\tabcolsep}{1.0pt}
\renewcommand{\arraystretch}{1.12}
\arrayrulecolor{WARule}
\begin{tabular*}{\textwidth}{@{\extracolsep{\fill}}l*{16}{r}@{}}
\toprule
\rowcolor{WAHeader}
\textbf{Model}
& \multicolumn{4}{c}{\cellcolor{WAHeader}\WAgroup{Perception}}
& \multicolumn{5}{c}{\cellcolor{WAHeader}\WAgroup{Consistency}}
& \multicolumn{3}{c}{\cellcolor{WAHeader}\WAgroup{Causality}}
& \multicolumn{4}{c}{\cellcolor{WAHeader}\WAgroup{Controllability}} \\
\cmidrule(lr){2-5}
\cmidrule(lr){6-10}
\cmidrule(lr){11-13}
\cmidrule(lr){14-17}
\rowcolor{WAHeader}
&
\textbf{SF} & \textbf{SuF} & \textbf{PQ} & \textbf{SR}
&
\textbf{SAC} & \textbf{GC} & \textbf{SSC} & \textbf{TC} & \textbf{MS}
&
\textbf{PP} & \textbf{TCO} & \textbf{CTV}
&
\textbf{GSA} & \textbf{AAC} & \textbf{FIS} & \textbf{MAO} \\
\midrule

MiniMax-H3
& \WAcell{99.17} & \WAcell{98.42} & \WAcell{70.17} & \WAcell{91.90}
& \WAcell{88.14} & \WAcell{51.30} & \WAcell{98.72} & \WAcell{72.85} & \WAcell{72.30}
& \WAcell{99.67} & \WAcell{97.00} & \WAcell{94.38}
& \WAcell{97.00} & \WAcell{96.33} & \WAcell{96.33} & \WAcell{97.00} \\

\rowcolor{WAAlternate}
HappyHorse 1.1
& \WAcell{98.91} & \WAcell{98.92} & \WAcell{68.80} & \WAcell{85.32}
& \WAcell{84.25} & \WAcell{46.65} & \WAcell{97.26} & \WAcell{66.62} & \WAcell{68.04}
& \WAcell{98.08} & \WAcell{95.33} & \WAcell{93.12}
& \WAcell{92.00} & \WAcell{94.08} & \WAcell{92.67} & \WAcell{95.33} \\

Grok Imagine Video 1.5
& \WAcell{98.98} & \WAcell{98.75} & \WAcell{73.53} & \WAcell{93.19}
& \WAcell{84.77} & \WAcell{40.24} & \WAcell{95.92} & \WAcell{56.94} & \WAcell{56.58}
& \WAcell{95.92} & \WAcell{95.50} & \WAcell{95.00}
& \WAcell{94.00} & \WAcell{93.42} & \WAcell{94.00} & \WAcell{95.50} \\

\rowcolor{WAAlternate}
Wan 3.0
& \WAcell{99.32} & \WAcell{98.40} & \WAcell{71.28} & \WAcell{93.95}
& \WAcell{86.84} & \WAcell{43.54} & \WAcell{96.66} & \WAcell{69.49} & \WAcell{70.31}
& \WAcell{96.33} & \WAcell{93.00} & \WAcell{89.38}
& \WAcell{89.00} & \WAcell{90.25} & \WAcell{91.83} & \WAcell{92.50} \\

Seedance 2.5
& \WAcell{99.62} & \WAcell{98.50} & \WAcell{68.07} & \WAcell{84.48}
& \WAcell{86.09} & \WAcell{52.74} & \WAcell{98.73} & \WAcell{76.73} & \WAcell{77.07}
& \WAcell{97.00} & \WAcell{88.00} & \WAcell{88.12}
& \WAcell{84.50} & \WAcell{87.83} & \WAcell{88.33} & \WAcell{89.00} \\

\rowcolor{WAAlternate}
Kling 3
& \WAcell{99.05} & \WAcell{98.66} & \WAcell{71.73} & \WAcell{98.22}
& \WAcell{87.09} & \WAcell{56.71} & \WAcell{98.06} & \WAcell{73.64} & \WAcell{71.52}
& \WAcell{94.83} & \WAcell{82.33} & \WAcell{86.25}
& \WAcell{81.00} & \WAcell{78.42} & \WAcell{79.67} & \WAcell{81.67} \\

Cosmos 3
& \WAcell{99.34} & \WAcell{98.67} & \WAcell{70.36} & \WAcell{97.39}
& \WAcell{86.74} & \WAcell{63.10} & \WAcell{98.24} & \WAcell{81.69} & \WAcell{82.35}
& \WAcell{96.58} & \WAcell{56.67} & \WAcell{68.12}
& \WAcell{55.50} & \WAcell{62.42} & \WAcell{60.00} & \WAcell{56.17} \\

\rowcolor{WAAlternate}
Sora 2
& \WAcell{93.30} & \WAcell{95.75} & \WAcell{63.92} & \WAcell{87.06}
& \WAcell{83.73} & \WAcell{56.62} & \WAcell{93.22} & \WAcell{75.06} & \WAcell{77.23}
& \WAcell{90.83} & \WAcell{49.83} & \WAcell{46.25}
& \WAcell{38.00} & \WAcell{52.17} & \WAcell{49.33} & \WAcell{50.17} \\

\bottomrule
\end{tabular*}
\arrayrulecolor{black}
\endgroup
\end{table}

\paragraph{W4: rule editing under a shared initial state.}
W4 evaluates paired rollouts generated from one initial state under either a
changed action condition or a changed physical rule. Because both branches
share the same first frame and differ only in the edited condition, differences between the two branches can therefore be used to assess the model's response to that condition.

\begin{figure}[t]
  \centering
  \includegraphics[width=\textwidth]{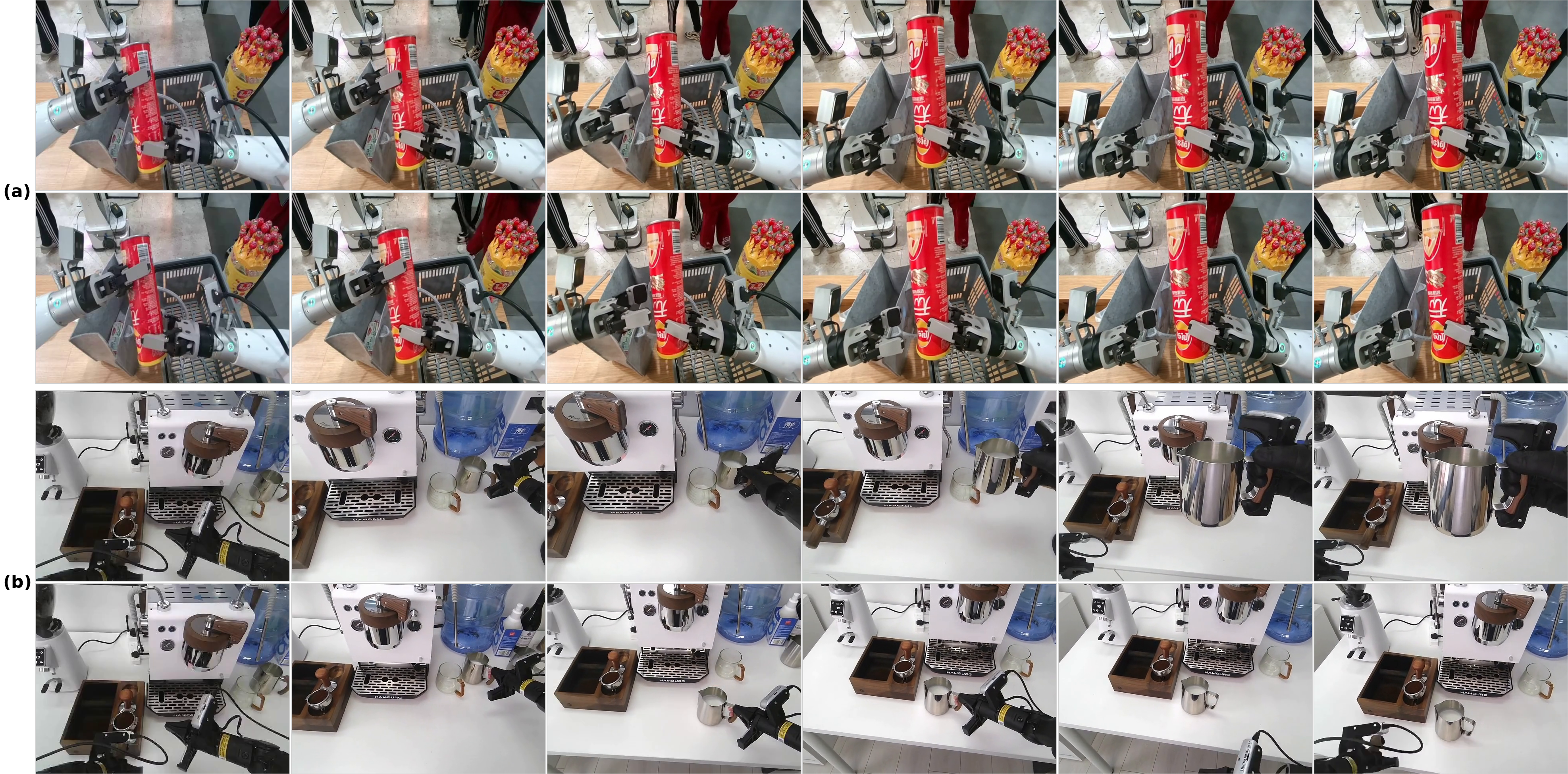}
  \caption{Representative W4 failure pairs from a shared initial state; each
  pair shows the two rollouts of one matched group under its two prescribed
  conditions, an edited action condition or an edited physical rule, with the
  upper and lower rows corresponding to the two conditions in the order stated.
  (a) Edited physical rule without observable consequence. Prompt: both rollouts
  grasp and lift the can, and the lower one additionally prescribes a
  soft-rubber material. Generated: both rollouts lift the can with identical
  rigid deformation, so the changed material rule leaves no trace in the
  dynamics and the pair remains visually indistinguishable.
  (b) Imprecise realization of the edited action condition. Prompt: the upper
  rollout is prescribed to grasp the silver milk pitcher at the right of the
  coffee machine and lift it upward, while the lower one is prescribed to
  contact the same pitcher and push it leftward along the table. Generated: the
  upper rollout lifts the pitcher as prescribed; in the lower one the pitcher
  does translate leftward, but the gripper produces this displacement by
  grasping its handle and dragging it across the table rather than by the
  prescribed pushing contact, so the condition is honored in direction but not
  in manner.}
  \label{fig:embodied-w4-failures}
\end{figure}

The dominant failure at this level is that the model does not respond to the
edit. For edited physical rules the response is close to absent: a changed
material, gravity, or friction setting leaves no identifiable trace in the
generated dynamics
(Fig.~\ref{fig:embodied-w4-failures}(a)). For edited action conditions the
response is partial rather than absent: the two branches do differentiate, and
the edited direction or target is roughly reflected in the outcome, but the
manner of interaction deviates from the prescription; a prescribed push, for
instance, is executed as a grasp-and-drag, so the object reaches the requested
side of the table without the requested pushing contact
(Fig.~\ref{fig:embodied-w4-failures}(b)); in more severe cases the prescribed
contact never occurs, the interaction is misdirected to another object, or the
two branches remain visually interchangeable, which is the same non-response
that the physical-rule cases exhibit. In both cases the causal coupling between
the edited condition and the generated dynamics breaks down: the condition is
stated in the prompt but does not constrain how the world evolves. A successful pair
should  establish a complete
condition--dynamics--contrast chain, in which the edited condition acts as a
constraint on the generated dynamics, the two branches differ exactly where
the condition differs, and the shared initial state and unaffected content
remain stable elsewhere.

\begin{table}[H]
\centering
\caption{W4 embodied scores: rule editing under a shared initial state. Scores are averages over the 92 formal W4 matched groups, pooling the
two intervention types of the level (40 action-condition groups and
52 physical-rule groups).}
\label{tab:embodied-w4-scores}
\begingroup
\fontsize{7.6}{8.9}\selectfont
\setlength{\tabcolsep}{1.0pt}
\renewcommand{\arraystretch}{1.12}
\arrayrulecolor{WARule}
\begin{tabular*}{\textwidth}{@{\extracolsep{\fill}}l*{15}{r}@{}}
\toprule
\rowcolor{WAHeader}
\textbf{Model}
& \multicolumn{4}{c}{\cellcolor{WAHeader}\WAgroup{Perception}}
& \multicolumn{5}{c}{\cellcolor{WAHeader}\WAgroup{Consistency}}
& \multicolumn{3}{c}{\cellcolor{WAHeader}\WAgroup{Causality}}
& \multicolumn{3}{c}{\cellcolor{WAHeader}\WAgroup{Controllability}} \\
\cmidrule(lr){2-5}
\cmidrule(lr){6-10}
\cmidrule(lr){11-13}
\cmidrule(lr){14-16}
\rowcolor{WAHeader}
&
\textbf{SF} & \textbf{SuF} & \textbf{PQ} & \textbf{SR}
&
\textbf{SAC} & \textbf{GC} & \textbf{SSC} & \textbf{TC} & \textbf{MS}
&
\textbf{PP} & \textbf{TCO} & \textbf{CTV}
&
\textbf{GSA} & \textbf{FIS} & \textbf{CBF} \\
\midrule
MiniMax-H3
& \WAcell{98.91} & \WAcell{97.69} & \WAcell{71.44} & \WAcell{96.40}
& \WAcell{88.81} & \WAcell{69.73} & \WAcell{98.17} & \WAcell{78.09} & \WAcell{76.11}
& \WAcell{97.15} & \WAcell{89.13} & \WAcell{89.67}
& \WAcell{87.50} & \WAcell{88.75} & \WAcell{87.39} \\
\rowcolor{WAAlternate}
HappyHorse 1.1
& \WAcell{98.97} & \WAcell{98.06} & \WAcell{69.31} & \WAcell{85.69}
& \WAcell{84.42} & \WAcell{48.98} & \WAcell{95.05} & \WAcell{59.54} & \WAcell{60.80}
& \WAcell{95.29} & \WAcell{86.41} & \WAcell{85.87}
& \WAcell{86.41} & \WAcell{85.00} & \WAcell{91.52} \\
Wan 3.0
& \WAcell{98.56} & \WAcell{98.25} & \WAcell{71.87} & \WAcell{95.38}
& \WAcell{86.74} & \WAcell{55.37} & \WAcell{96.09} & \WAcell{65.61} & \WAcell{64.98}
& \WAcell{92.03} & \WAcell{83.70} & \WAcell{85.33}
& \WAcell{83.15} & \WAcell{90.00} & \WAcell{81.52} \\
\rowcolor{WAAlternate}
Seedance 2.5
& \WAcell{98.83} & \WAcell{98.19} & \WAcell{69.62} & \WAcell{90.91}
& \WAcell{87.65} & \WAcell{64.37} & \WAcell{97.91} & \WAcell{77.00} & \WAcell{75.92}
& \WAcell{94.11} & \WAcell{82.61} & \WAcell{80.43}
& \WAcell{79.35} & \WAcell{82.50} & \WAcell{82.17} \\
Grok Imagine Video 1.5
& \WAcell{98.88} & \WAcell{97.28} & \WAcell{73.39} & \WAcell{95.57}
& \WAcell{84.82} & \WAcell{48.30} & \WAcell{92.98} & \WAcell{48.75} & \WAcell{47.18}
& \WAcell{88.99} & \WAcell{86.96} & \WAcell{85.87}
& \WAcell{86.41} & \WAcell{90.00} & \WAcell{85.22} \\
\rowcolor{WAAlternate}
Kling 3
& \WAcell{98.77} & \WAcell{96.58} & \WAcell{71.21} & \WAcell{97.63}
& \WAcell{88.65} & \WAcell{66.01} & \WAcell{96.47} & \WAcell{70.97} & \WAcell{67.07}
& \WAcell{87.27} & \WAcell{70.11} & \WAcell{74.46}
& \WAcell{65.22} & \WAcell{75.00} & \WAcell{77.17} \\
Cosmos 3
& \WAcell{99.12} & \WAcell{97.26} & \WAcell{70.54} & \WAcell{97.48}
& \WAcell{87.80} & \WAcell{74.35} & \WAcell{98.01} & \WAcell{82.12} & \WAcell{82.37}
& \WAcell{89.72} & \WAcell{45.11} & \WAcell{48.37}
& \WAcell{44.02} & \WAcell{47.50} & \WAcell{56.52} \\
\rowcolor{WAAlternate}
Sora 2
& \WAcell{93.95} & \WAcell{94.03} & \WAcell{64.34} & \WAcell{91.59}
& \WAcell{87.47} & \WAcell{65.54} & \WAcell{93.12} & \WAcell{75.30} & \WAcell{76.76}
& \WAcell{84.01} & \WAcell{42.93} & \WAcell{42.39}
& \WAcell{40.22} & \WAcell{41.25} & \WAcell{59.13} \\
\bottomrule
\end{tabular*}
\arrayrulecolor{black}
\endgroup
\end{table}
\paragraph{Cross-level capability analysis.}
Across W2--W4, the results reveal a progressive gap between visual
fidelity and faithful modeling of action-conditioned world dynamics.
Perceptual quality, scene appearance, and overall visual structure remain
generally strong across the three levels, indicating that current video
models can generate realistic robot-centric rollouts from real-world
first-person observations. The main limitations instead emerge in
grounding actions in the world and modeling their consequences, and become
more pronounced as the interaction requirements increase.

At W2, models can often generate motions that resemble the prescribed
action, but still struggle to ground these motions in the intended
interaction, including the prescribed contact, object response, and
terminal state. At W3, this challenge extends to state persistence:
states established by earlier actions are not always reliably transferred
to subsequent stages, leading to failures in long-horizon action
execution and accumulated scene changes. At W4, models further struggle
to make the generated dynamics respond specifically to edited action or
physical conditions. Action edits may be reflected only partially in the
manner of interaction, while changes to physical rules such as material,
gravity, or friction often produce little observable change.

Overall, current video models achieve relatively strong visual fidelity
in robot-centric rollouts, but remain limited in action grounding,
persistent state evolution, and condition-dependent physical response.
These capabilities distinguish embodied world modeling from conventional
video generation and remain key challenges for reliable action-conditioned
simulation.
\section{Conclusion}
We introduced HappyWorld-Bench, a benchmark that brings video, spatial, and embodied world models into a shared W1–W6 capability framework. Its central premise is that world models should be evaluated by both the quality of their generated worlds and their reliability under interaction, persistent state changes, and intervention. Across all three tracks, HappyWorld-Arena complements automated capability metrics with model-level Elo ratings derived from human A/B comparisons. By combining a common capability vocabulary with automated evaluation and human preference assessments, this work helps these research directions converge on shared requirements for useful world models. We hope this framework will guide the integration of advances in generation, simulation, and embodied prediction toward systems that support dependable exploration, planning, and action in complex environments.

\bibliographystyle{unsrtnat}
\bibliography{references,embodied_wm_assets/em_rf}

\addtocontents{toc}{\protect\setcounter{tocdepth}{-1}}
\clearpage
\onecolumn
\appendix
\onecolumn

\section{Video World Model Track}

\subsection{Inference Protocols and Model Configurations}
\label{app:video-inference}

This section describes the inference procedures used to generate model outputs for the Video World Model Track. The evaluated models differ in their conditioning inputs, control interfaces, and supported rollout lengths. We therefore document how benchmark cases are translated into each model's input format, including reference images, textual descriptions, and timed controls where supported. For each model, we specify the inference configuration, rollout procedure, and any interface limitations or adaptations that affect task execution. These details clarify how the prescribed tasks are instantiated across models and provide context for interpreting their evaluation results.

\paragraph{Genie 3 via Project Genie.}
We evaluate Genie 3~\citep{genie3} through the hosted Project Genie web interface~\citep{projectgenie2026}, using browser automation implemented with Playwright and Google Chrome. Project Genie integrates Genie 3 with additional components for world creation and sketching. Accordingly, our results characterize the deployed Genie 3-based system rather than an isolated model checkpoint.

\emph{Input conditioning.}
For each benchmark case, we upload the reference image and populate the Environment and Character fields with the provided environment and role descriptions, respectively. The first- or third-person perspective is configured according to the case specification. Descriptions are transferred without semantic rewriting; when a description exceeds the interface's character limit, we retain its leading characters up to the supported length. In the standardized preprocessing pipeline, reference images are decoded, orientation-corrected, and converted to RGB PNG. Images below the interface's minimum input dimensions of $1280 \times 704$ pixels are enlarged using Lanczos interpolation. For an image of size $W \times H$, the scaling factor is $s=\max(1,1280/W,704/H)$, preserving the aspect ratio up to integer rounding without cropping. This preprocessing does not use learned or generative super-resolution.

\emph{World initialization and controlled rollout.}
Each case is initialized from a fresh creation-page state to prevent reuse of a preceding case's image or world. The automation completes the sketch-creation step when required, invokes world creation, and waits until a playable world is detected before executing the prescribed controls. Translational actions are mapped to the \texttt{W/A/S/D} keys, whereas camera rotations are mapped to the arrow keys. In particular, the benchmark labels \texttt{Mouse\_Up}, \texttt{Mouse\_Down}, \texttt{Mouse\_Left}, and \texttt{Mouse\_Right} are implemented as keyboard camera controls rather than physical mouse motion. Each action segment is scheduled using its specified start time and duration; translation and rotation inputs within a segment are applied concurrently and released at its end. No-input intervals are implemented by leaving all control keys released. Cases without an action sequence use a 15-second no-input rollout. For action-controlled cases, the intended rollout horizon is $T=\max_i(t_i+\Delta t_i)$, where $t_i$ and $\Delta t_i$ denote the start time and duration of segment $i$. Controls are scheduled in wall-clock time through the browser, rather than synchronized to individual generated frames.

\emph{Configuration, export, and operational limitations.}
We use the hosted service's default generation settings. Our interface does not expose the underlying checkpoint identifier, random seed, sampling steps, or guidance scale; these quantities are therefore not controlled in our experiments. After the action sequence, the automation exits exploration and retrieves the video through the application's built-in download function, rather than recording the desktop display. Videos are retained at their exported spatial resolution, predominantly $1280 \times 704$ pixels. Transient generation and download failures trigger retries with the same case inputs. The automated duration check uses a tolerance of $\pm 5$ seconds relative to the intended rollout horizon, and out-of-tolerance cases are queued for rerunning. Retry decisions are operational rather than based on benchmark evaluation scores. Because browser interaction, service latency, and export boundaries can introduce timing discrepancies, the prescribed control duration should not be interpreted as an exact encoded video duration. We evaluate the W1--W3 subset supported by the available control interface; cases requiring unsupported intervention or event controls are recorded as unsupported rather than approximated with no-input rollouts.

\paragraph{HappyOyster.}
We evaluate HappyOyster~\citep{alibabatokenhub2026happyoyster} under a standardized inference protocol. A single model configuration and a common set of decoding parameters are used across all benchmark cases, without case-specific adaptation or tuning.

\emph{Input conditioning.}
Each case is initialized independently from its reference image and the corresponding textual description. The specified first- or third-person perspective is provided as part of the conditioning when available. The inference canvas is adapted to the aspect ratio of each reference image under a fixed pixel budget approximately equivalent to $832 \times 480$ pixels. Both spatial dimensions are rounded to model-compatible multiples of 16. Consequently, generated videos approximately preserve the reference image's aspect ratio, but do not necessarily retain its original pixel dimensions. No cropping or learned super-resolution is applied during this preprocessing. 

\emph{Action-conditioned rollout.}
Benchmark actions are converted into the model's frame-aligned control representation. Translational and camera-control commands follow the action type, start time, and duration specified by each case, while intervals without an action are represented by no-input controls. Simultaneous commands are applied jointly. Videos are generated causally over successive temporal segments, with the model state carried forward within each rollout. The target rollout duration is determined by the benchmark action sequence; cases without an action sequence are generated using a no-input rollout.





\paragraph{Lingbot-World-v2} We implement the distilled causal configuration of LingBot-World-v2~\citep{gao2026infinite}, conditioned on the reference image, a textual description, and a continuous camera trajectory. Movement and rotation commands are converted into camera-to-world poses by accumulating local translation and changes in orientation over each control interval. Diagonal translation directions are normalized to maintain approximately the same movement magnitude as single-axis commands. The resulting poses and camera intrinsics are used to construct Plücker embeddings, which convey the prescribed camera motion to the model. For cases containing event instructions without discrete movement controls, the event descriptions are appended to the environment description and the prescribed camera trajectory is held constant. The assembled prompt is truncated to the supported length and encoded once for the entire rollout. Event descriptions therefore provide global conditioning rather than updates aligned to their individual execution intervals, limiting the temporal precision of language-based interventions. Videos are generated at 16 fps with an approximately 832 x 480-pixel resolution budget.

\paragraph{MatrixGame-2.0} We select the base-distilled Matrix-Game 2.0~\citep{matrixgame2} model in universal-control mode, conditioned on the reference image and a sequence of keyboard and mouse inputs. The image is center-cropped and resized to the output resolution. Movement commands are converted into four-dimensional binary vectors representing forward, backward, leftward, and rightward translation, while camera rotation is represented by a two-dimensional pitch–yaw vector. Diagonal movement activates both relevant directional components, and translation and rotation may be applied simultaneously. Each control interval is expanded into framewise inputs, with zero vectors representing idle intervals. The evaluated inference path does not use textual scene or role descriptions. Cases without discrete controls receive a zero-action sequence, and natural-language intervention instructions are not incorporated. These cases therefore produce image-conditioned rollouts without commanded actions and do not implement the requested W4 interventions. Zero control denotes the absence of commanded movement, rather than requiring the generated scene to remain stationary. Output videos have a fixed resolution of 640 x 352 pixels at 12 fps.

\paragraph{MatrixGame-3.0} We use the distilled Matrix-Game 3.0~\citep{wang2026matrixgame30realtimestreaming} model, conditioned on the reference image, the scene and role descriptions, and movement and camera controls. The image is center-cropped and resized to the output resolution. Movement commands are converted into six-dimensional keyboard vectors containing four directional components and two unused components fixed to zero. Camera rotation is represented by a pitch–yaw vector. The controls are expanded framewise, preserving diagonal movement and simultaneous translation and rotation. The action sequence is also integrated into camera poses to construct Plücker embeddings, providing geometric conditioning alongside the explicit action signals. For cases without discrete controls, zero keyboard and mouse inputs are supplied. The scene and role descriptions remain active, but natural-language event instructions are not incorporated into the text conditioning. This configuration therefore supports image- and text-conditioned scene evolution without commanded movement, but does not implement timed W4 interventions. Output videos have a fixed resolution of 1280 x 704 pixels at 16 fps.

\paragraph{Yume-1.5} We use the autoregressive Yume-1.5 model (5B, 720P)~\citep{Mao_2026_CVPR}, conditioned on the reference image, a scene and role description, and a sequence of keyboard and mouse controls. The environment and role descriptions are concatenated into a single prompt and refined once by an InternVL captioner~\citep{chen2024internvl} before generation. Unlike pose-conditioned baselines, Yume-1.5 accepts motion through a structured camera-control vocabulary expressed in language: translation commands map to W/A/S/D tokens (forward, left, back, right), with diagonal movement rendered as combined tokens (e.g.\ W+A), and camera rotation maps to mouse pan and tilt tokens (e.g.\ $\leftarrow$, $\uparrow$) with their diagonal combinations. Fixed movement- and turn-speed magnitudes accompany each token. Each control interval is emitted as one caption line and quantized to the model's fixed autoregressive chunk length (29 frames at 16 fps, $\approx$1.81\,s), so an interval of $s$ seconds spans $\max(1,\operatorname{round}(s/1.81))$ chunks. The clip is generated chunk by chunk, each conditioned on the accumulated preceding frames and the corresponding control line. Because the caption interface admits only the structured camera-control vocabulary, cases without discrete controls and natural-language event instructions cannot be expressed and are not evaluated. This configuration therefore supports image- and control-conditioned rollouts but does not implement natural-language W4 interventions. Output videos have a fixed resolution of 1280 x 704 pixels at 16 fps.

\paragraph{DreamX-World 1.0} We adopt the distilled autoregressive-forcing configuration of DreamX-World 1.0~\citep{dreamxteam2026dreamxworld10generalpurposeinteractive}, conditioned on the reference image, a textual description, and a discrete keyboard-style control sequence. The reference image is resized to the native 1280 x 704 output resolution and encoded once as the initial latent. Movement commands are converted into translations along the camera's local forward, backward, left, and right axes, while rotation commands adjust the viewing direction in pitch and yaw; diagonal movement sums the corresponding directional components, and translation and rotation may be applied simultaneously. The accumulated poses are expressed relative to the initial camera and combined with camera intrinsics to construct PRoPE conditioning that conveys the prescribed motion to the model. Each control interval is allocated a frame budget proportional to its duration, so that per-segment timing is preserved throughout the rollout. Because the control signal specifies only camera extrinsics and intrinsics, the model receives no independently commanded character state: in first-person cases forward camera motion reads as locomotion, whereas in third-person cases the subject is not driven to move with the camera. The environment and role descriptions are concatenated into a single prompt and encoded once for the entire rollout, so event instructions are not represented as temporally aligned text-conditioning changes. Cases containing event instructions without discrete movement controls are not generated, and language-based interventions are therefore not implemented. Output videos have a fixed resolution of 1280 x 704 pixels at 16 fps.

\paragraph{ABot-World} We use the distilled causal ABot-World-0-5B model~\citep{jiang2026abotworld0infiniteinteractiveworld}, conditioned on the reference image, the environment and role descriptions, and a sequence of keyboard and mouse controls. The image is resized to the output resolution and encoded as the first-frame latent. Movement commands are converted into an eight-dimensional binary key vector comprising four translation components (W/A/S/D) and four rotation components (I/J/K/L), where camera rotation is mapped from the mouse pitch and yaw directions to their corresponding keys. Diagonal movement activates both relevant translation components, and translation and rotation may be applied simultaneously; idle intervals are represented by an all-zero vector. Each control interval is expanded into a whole number of autoregressive generation blocks according to its duration, with one block nominally covering 0.5 s of output, and the model streams the video block by block using a relative-position KV cache. The generated frames are uniformly resampled to the prescribed arena time scale, so  that the clip length matches the summed control durations. The environment and role descriptions are encoded once and held fixed for the entire rollout, providing global rather than temporally aligned conditioning. For cases without discrete controls, an all-zero action sequence spanning the case duration is supplied, and natural-language event instructions are not incorporated into the text conditioning. This configuration produces image- and text-conditioned rollouts without commanded movement and does not implement the timed W4 interventions. Output videos have a fixed resolution of 832 x 480 pixels at 12 fps.

\paragraph{Cosmos3} We use the Nano variant of Cosmos3~\citep{agarwal2026cosmos} in its forward-dynamics mode, conditioned on the reference image, the environment and role descriptions, and a sequence of movement and camera controls. Rather than the model's explicit camera-pose control interface---which on this benchmark responds reliably only to in-distribution scenes and is otherwise largely ignored---we adopt the autonomous-driving (av) action domain, which produces stable motion across the full set of cases. Movement and camera commands are compiled into a per-frame relative-pose sequence in which each frame carries a three-dimensional translation and a six-dimensional rotation representation. Forward and backward commands drive translation along the viewing direction, lateral commands are rendered as curved strafes rather than pure sideways motion, and mouse controls are mapped to yaw; pitch has no analogue in this domain and is dropped. Because yaw is only rendered while the camera is translating, turning-in-place intervals are assigned a small forward speed so that rotation becomes visible. Diagonal movement combines the corresponding forward and turning components, and translation and rotation may be applied simultaneously. Each control interval is expanded into framewise poses by integrating a unicycle model over its duration, and the per-frame motion magnitudes are normalized by the frame rate so that world-space speed and total rotation are independent of the output frame rate. Long clips are produced autoregressively by splitting the rollout into fixed-length sessions and conditioning each continuation on the final frame of the preceding one. Idle intervals within an action sequence are represented by zero-motion frames. The environment and role descriptions are concatenated into a single prompt that provides global text conditioning throughout the rollout, while natural-language event instructions are not incorporated, so the configuration does not implement timed W4 interventions. Because the forward-dynamics interface conditions on numerical actions rather than text, cases specified purely through event instructions, without any movement controls, are not rendered under this configuration. The generated video is re-encoded and trimmed so that its total length matches the duration requested by the control sequence. Output videos are produced at 16 fps with a 480-pixel short-side resolution budget (e.g., 736 x 480).

\paragraph{JoyAI-Echo} We choose the Echo-WM~\citep{zhang2026echowmopenenterableomnimodal} Flash causal model, conditioned on the reference image, the environment and role descriptions, and a sequence of movement and camera controls. Control intervals are encoded as directional key combinations with associated durations, preserving simultaneous inputs and merging adjacent intervals with identical controls. These discrete commands are internally expanded into a continuous camera trajectory. The trajectory is expressed relative to the initial pose and combined with camera intrinsics to provide UCPE conditioning. The camera controller smooths transitions when controls are released, so an idle interval following movement can include residual camera motion. When discrete controls are absent, neutral inputs are supplied. Event instructions, when present, are appended to the scene and role descriptions as one global prompt; their individual execution intervals are not represented by timed text-conditioning changes. When discrete controls are present, event instructions are not appended. The evaluated configuration therefore provides global language conditioning but does not implement temporally aligned intervention prompts. Output videos have a fixed resolution of 1280 x 704 pixels at 24 fps.

\paragraph{Alaya-EVOKE}
We use Alaya-EVOKE~\citep{yin2026alayaevokelinearscalingsupervisionendless} for camera-controlled image-to-video generation, conditioned on a reference image, the environment and role descriptions, and a camera trajectory. The control sequence is converted into a camera-to- orld trajectory (\texttt{cam\_c2w}) with associated intrinsics at $30$~fps and resampled to $24$~fps by the inference engine. Translation commands are integrated in the camera's local coordinate frame, while mouse commands specify yaw and pitch changes. Inference uses the released post-distillation checkpoint with three pyramid sampling steps and no classifier-free guidance. For a case of duration $d$ seconds, the engine frame count is set to $33\lceil 24d/36\rceil$, yielding $36\lceil 24d/36\rceil-3$ output frames. Videos are generated at a resolution of $640\times384$ pixels and a frame rate of $24$~fps.

\paragraph{Lyra 2.0} We use Lyra 2.0~\citep{shen2026lyra20explorablegenerative} for trajectory-conditioned video generation from a reference image and the environment and role descriptions. Movement and camera commands are converted into a continuous sequence of camera extrinsics and intrinsics. In first-person cases, the trajectory is constructed by accumulating local translation and rotation, with diagonal translation obtained by summing the corresponding directional components. In third-person cases, the camera follows a look-at trajectory around a proxy subject position. Movement commands translate this position, while camera commands adjust the viewing direction through orbital motion. This construction specifies camera motion without supplying an independently controlled character state. Initial scene geometry is estimated from the reference image using DA3, with MoGe-based depth-scale alignment. The evaluated adaptation supplies a single scene and role description throughout generation. Output videos have a fixed resolution of 832 x 480 pixels at 16 fps.

\paragraph{SANA-WM}
We use SANA-WM \citep{zhu2026sanawmefficientminutescaleworld} for camera-controlled image-to-video generation, conditioned on a reference image, a text prompt, and a camera trajectory. The environment and role descriptions are concatenated into a single prompt and encoded with the Gemma-2-2B text encoder. Reference images are resized while preserving their aspect ratio and then center-cropped to $1280\times704$ pixels. The control sequence is converted into a per-frame camera trajectory with associated intrinsics at $16$~fps. Translation commands are integrated in the camera's local coordinate frame, with diagonal motion computed by summing the corresponding directional components; mouse commands specify yaw and pitch changes. The resulting camera parameters are supplied through the model's camera-pose embedding. Inference uses the flow-DPM solver with $60$ sampling steps, linear flow, and an inference shift of $9.8$. The LTX-2 latent refiner and action visualization overlays are disabled. For a case of duration $d$ seconds, we request $16d$ frames. Output videos have a resolution of $1280\times704$ pixels and a frame rate of $16$~fps.

\paragraph{Open-Oasis}
We use Open-Oasis~\citep{oasis2024} for autoregressive video generation, conditioned on a reference image and per-frame movement and camera controls. The control sequence is sampled at $20$\,fps and encoded as a $25$-dimensional action vector comprising binary key indicators and camera-axis values. Movement commands activate the corresponding \texttt{forward}, \texttt{back}, \texttt{left}, and \texttt{right} components, while mouse commands are mapped to \texttt{cameraX} and \texttt{cameraY} with a magnitude of $0.1$. The reference image is resized to $640\times360$ pixels. Inference is performed in FP32 with $10$ DDIM steps per generated frame. For a case of duration $d$ seconds, the output video contains $20d$ frames at a resolution of $640\times360$ pixels and a frame rate of $20$~fps.

\subsection{Detailed Metrics}

We evaluate world-model videos along four dimensions: perception and representation, consistency and state retention, causality and causal rollout, and controllable interaction. The metric names and abbreviations below follow the W1 - W4 result tables. Unless stated otherwise, normalized scores lie in $[0,1]$, reported percentage scores equal 100 times the normalized value, and higher values are better. Inapplicable cases and missing measurements are excluded from aggregation rather than assigned a zero. We denote sampled video frames by $I_t$ and use $\cosim{\mathbf{x}}{\mathbf{y}}$ for cosine similarity.

\subsubsection{Perception and Representation}

\paragraph*{\textnormal{\textbf{Imaging Quality (VQ).}}} This metric measures per-frame technical and perceptual image quality, including blur, noise, compression artifacts, exposure failures, and structural corruption. MUSIQ assigns a no-reference quality score to each sampled frame, and the video score is the average of the normalized frame scores. If MUSIQ returns $q_t\in[0,100]$, the intended score is
\begin{equation}
S_{\mathrm{VQ}}=\frac{1}{T}\sum_{t=0}^{T-1}\frac{q_t}{100},
\qquad S_{\mathrm{VQ}}^{(100)}=100S_{\mathrm{VQ}}.
\end{equation}
High VQ indicates clean and natural frames but does not imply temporal, semantic, or physical correctness.

\paragraph*{\textnormal{\textbf{Human Preference Score v3 (HPS).}}} This metric estimates broad human preference over visual appeal, composition, naturalness, and acceptability using the HPSv3 Qwen2-VL-7B reward model. For per-frame rewards $r_t$, extreme values are clipped at the first and ninety-ninth percentiles before temporal averaging:
\begin{equation}
\widetilde r_t=\min\!\left(P_{99},\max(P_1,r_t)\right),
\qquad S_{\mathrm{HPS}}=\frac{1}{T}\sum_{t=0}^{T-1}\widetilde r_t.
\end{equation}
HPS complements MUSIQ: it is closer to overall preference, whereas MUSIQ emphasizes technical quality. High HPS does not establish instruction following, causal correctness, or temporal stability.

\paragraph*{\textnormal{\textbf{Brightness Consistency (Bri).}}} This metric detects unjustified exposure or brightness-distribution drift. Each RGB frame is converted to grayscale, and the proportions of pixels in $[0,63]$, $[64,191]$, and $[192,255]$ form $\mathbf b_t\in\mathbb R^3$. Define
\begin{equation}
g_{\lambda}(x)=\frac{e^{\lambda\operatorname{clip}(x,0,1)}-1}{e^{\lambda}-1+\epsilon}.
\end{equation}
The brightness-consistency score is
\begin{equation}
S_{\mathrm{Bri}}=\frac{1}{T}\sum_{t=0}^{T-1}
g_{15}\!\left(\cosim{\mathbf b_0}{\mathbf b_t}\right),
\qquad S_{\mathrm{Bri}}^{(100)}=100S_{\mathrm{Bri}}.
\end{equation}
High values mean that dark, midtone, and bright proportions remain close to the first frame. Because the representation is a coarse histogram, it ignores spatial layout and may penalize an intentionally requested lighting transition unless that case is filtered or interpreted separately.

\paragraph*{\textnormal{\textbf{Color-Temperature Consistency (CT).}}} This metric measures stability of global hue and, by proxy, color temperature. Each frame is converted to HSV, and Hue is quantized into $[0,15]$, $[16,30]$, $[31,60]$, $[61,90]$, $[91,120]$, $[121,150]$, and $[151,179]$, producing histogram $\mathbf h_t$. The first-frame score is one; for $t\ge1$,
\begin{equation}
s_t=g_{15}\!\left(
\frac{\cosim{\mathbf h_0}{\mathbf h_t}+
\cosim{\mathbf h_{t-1}}{\mathbf h_t}}{2}
\right),
\qquad
S_{\mathrm{CT}}=\frac{1+\sum_{t=1}^{T-1}s_t}{T}.
\end{equation}
The reported percentage is $100S_{\mathrm{CT}}$. Comparing both the first and preceding frames captures gradual drift and sudden jumps. The metric is distribution-based and does not localize the source of a color change.

\paragraph*{\textnormal{\textbf{Sharpness Retention (SR).}}} This metric measures whether edge structure remains similar to the first frame while preventing persistent high-frequency noise from being rewarded as detail. For grayscale frame $I_t$, the edge descriptor is $\mathbf e_t=(\sum_p\partial_x I_t(p),\sum_p\partial_y I_t(p))$ from summed Sobel responses. The first frame scores one; later frames use $g_3(\cosim{\mathbf e_0}{\mathbf e_t})$. PIQ-BRISQUE supplies a noise score; every score at least $0.5$ increments a cumulative counter, and once the counter reaches three, the current and all subsequent similarity scores are zeroed. Thus
\begin{equation}
S_{\mathrm{SR}}=\frac{1}{T}\left[1+\sum_{t=1}^{T-1}
\mathbb 1(\text{noise gate not triggered at }t)
g_3\!\left(\cosim{\mathbf e_0}{\mathbf e_t}\right)\right],
\qquad S_{\mathrm{SR}}^{(100)}=100S_{\mathrm{SR}}.
\end{equation}
This measures retention rather than absolute sharpness and should be read with VQ if the first frame is poor.

\paragraph*{\textnormal{\textbf{Imaging Stability (IS).}}} IS summarizes temporal image stability by combining brightness consistency, color-temperature consistency, and sharpness retention. These components capture complementary forms of visual instability: exposure drift, hue or color-temperature drift, and loss of edge clarity. After the three component scores have been normalized to $[0,1]$, IS is their arithmetic mean:
\begin{equation}
S_{\mathrm{IS}}=\frac{S_{\mathrm{Bri}}+S_{\mathrm{CT}}+S_{\mathrm{SR}}}{3}.
\end{equation}
The reported percentage is $100S_{\mathrm{IS}}$. A high IS score therefore requires stability in brightness, color temperature, and sharpness.

\paragraph*{\textnormal{\textbf{Dynamic Degree (Dyn).}}} Dyn measures the amount of visible motion using RAFT optical flow. It captures whether a video contains substantial temporal change, independently of whether that motion is commanded, physically valid, or temporally smooth. Its score is obtained by aggregating adjacent-frame optical-flow magnitudes:
\begin{equation}
\mathrm{Dyn}\propto\frac{1}{T-1}\sum_{t=0}^{T-2}
\operatorname{Agg}_{p}\|\mathbf u_t(p)\|_2,
\end{equation}
where $\mathbf u_t(p)$ is the optical flow at pixel $p$ between adjacent frames. Higher Dyn indicates a greater amount of visible motion. Dyn should be interpreted together with MS: Dyn measures motion magnitude, whereas MS measures motion continuity. Flicker is excluded because temporal smoothness is represented by MS.

\paragraph*{\textnormal{\textbf{Instruction Following (IF).}}} This metric checks atomic prompt requirements, including presence, count, attributes, spatial relations, style, lighting, and viewpoint. A prompt-derived checklist decomposes the instruction into atomic items. Presence, count, and attribute items use a binary scale, while style items use a 1--5 scale. Item normalization and dimension aggregation are
\begin{equation}
\widetilde r_i=
\begin{cases}
r_i, & r_i\in\{0,1\},\\
(r_i-1)/4, & r_i\in\{1,2,3,4,5\},
\end{cases}
\qquad
S_{\mathrm{IF}}=\frac{1}{|\mathcal I|}\sum_{i\in\mathcal I}\widetilde r_i.
\end{equation}
The score is the direct mean of all applicable items, and cases with insufficient applicable evidence are excluded from aggregation. For W4, IF is evaluated only before the first intervention and focuses on the fidelity of the initial environment, subject, viewpoint, pose, and object states. If a later intervention explicitly requests an attribute to change, that instructed change does not count as an IF failure. W4 items use $v_i\in\{0,0.5,1\}$ and are averaged as
\begin{equation}
S_{\mathrm{IF}}^{\mathrm{W4}}=\frac{1}{|\mathcal I_{\mathrm{W4}}|}
\sum_{i\in\mathcal I_{\mathrm{W4}}}v_i.
\end{equation}

\subsubsection{Consistency and State Retention}

\paragraph*{\textnormal{\textbf{Background Consistency (BC).}}} This metric measures environmental stability after suppressing the moving subject. Frames are sampled at 3 fps. Subject or first-person hand regions are masked out so that foreground motion does not dominate background similarity. CLIP ViT-B/16 produces unit-normalized $\mathbf f_t$. Background consistency combines short-term similarity between adjacent frames and long-term similarity to the first frame:
\begin{equation}
s_t=\frac{\max(0,\cosim{\mathbf f_{t-1}}{\mathbf f_t})+
\max(0,\cosim{\mathbf f_0}{\mathbf f_t})}{2},
\qquad
S_{\mathrm{BC}}=\frac{1}{T-1}\sum_{t=1}^{T-1}s_t.
\end{equation}
The percentage is $100S_{\mathrm{BC}}$. The adjacent term catches abrupt changes and the first-frame term catches accumulated drift. Because this finalized version masks the subject, it is not numerically comparable to standard full-frame CLIP background consistency.

\paragraph*{\textnormal{\textbf{Object Geometry Consistency (GC).}}} This metric assesses whether 3D structure is mutually compatible across frames. DA3 depth and camera poses are used to reproject scene points across frames and measure geometric agreement. Conceptually, pixels are back-projected using estimated depth, transformed by relative camera pose, projected into another frame, and compared with target depth:
\begin{equation}
S_{\mathrm{GC}}=\operatorname{Agg}_{s,t,p}
\phi_{\mathrm{geo}}\!\left(e^{\mathrm{geo}}_{s\rightarrow t}(p)\right),
\qquad S_{\mathrm{GC}}^{(100)}=100S_{\mathrm{GC}}.
\end{equation}
Higher scores indicate that the reconstructed structure remains stable across views. This measure is based on geometric depth reprojection and should not be conflated with appearance consistency.

\paragraph*{\textnormal{\textbf{Texture Consistency (TC).}}} This metric measures whether corresponding surfaces retain appearance after geometric alignment. It shares the depth, pose, and reprojection correspondences used by GC, but compares aligned image appearance through a PSNR-like photometric measure. A representative photometric residual is
\begin{equation}
\operatorname{MSE}_{s\rightarrow t}=\frac{1}{|\Omega|}
\sum_{p\in\Omega}\left\|I_s(p)-I_t\!\left(\pi_{s\rightarrow t}(p)\right)\right\|_2^2,
\end{equation}
followed by PSNR-style aggregation. Higher values indicate that corresponding surfaces retain similar color and texture across views.

\paragraph*{\textnormal{\textbf{State Consistency (SC).}}} This metric evaluates whether object states, spatial relations, and established outcomes persist unless an instruction or valid cause changes them. Sampled frames and the scene description are evaluated with an integer rubric $r\in\{1,2,3,4,5\}$. The score is
\begin{equation}
S_{\mathrm{SC}}^{\mathrm{raw}}=r,
\qquad S_{\mathrm{SC}}^{(100)}=100\frac{r}{5}.
\end{equation}
One denotes severe or repeated inconsistencies; five denotes stable states and justified changes. This differs from WorldMark's official quantitative LPIPS+DINOv2 C1/C2 procedure, so the scores must not be mixed.

\paragraph*{\textnormal{\textbf{Subject Consistency (SuC).}}} This metric evaluates preservation of the main subject's identity and appearance. SAM2.1 masks isolate the subject; padded bounding-box crops are placed on a gray background and resized to $128\times128$. DINOv2 measures adjacent-frame identity continuity, while CLIP measures long-range similarity to the first frame. A representative aggregation is
\begin{equation}
S_{\mathrm{SuC}}=\frac{1}{T-1}\sum_{t=1}^{T-1}
\frac{\max(0,\cosim{\mathbf d_{t-1}}{\mathbf d_t})+
\max(0,\cosim{\mathbf c_0}{\mathbf c_t})}{2}.
\end{equation}
Higher values indicate stronger preservation of subject identity and appearance.

\subsubsection{Causality and Causal Rollout}

\paragraph*{\textnormal{\textbf{Physical Causality (PC).}}} This metric evaluates visible motion, contact, force, gravity, material response, and physical outcomes. It applies only when the prompt specifies an observable physical process. Each applicable physical requirement is expressed as an atomic positive assertion and scored on a 1--5 scale:
\begin{equation}
\widetilde r_i=\frac{r_i-1}{4},
\qquad
S_{\mathrm{PC}}=\frac{1}{|\mathcal P|}\sum_{i\in\mathcal P}\widetilde r_i,
\end{equation}
with two items required by default. Only applicable and visibly assessable items contribute to the average. High PC means that the observed process and outcome are physically credible, not merely that the final frame looks plausible.

\paragraph*{\textnormal{\textbf{Content Causality (CC).}}} This metric assesses logical event progression: a specified trigger should precede and produce its consequence. It applies only when the prompt or event sequence contains an explicit action-to-consequence chain. Each applicable item uses a 1--5 scale, normalized and aggregated as
\begin{equation}
\widetilde r_i=\frac{r_i-1}{4},
\qquad
S_{\mathrm{CC}}=\frac{1}{|\mathcal C|}\sum_{i\in\mathcal C}\widetilde r_i,
\end{equation}
with the same default two-item minimum. High scores require the correct trigger, order, consequence, and persistence; mere co-occurrence is insufficient. As in PC, fully invisible prerequisites may be marked inapplicable, whereas visible but incorrect behavior remains applicable and scores low.

\paragraph*{\textnormal{\textbf{Interpenetration (IP).}}} This metric checks relevant object pairs for penetration, embedding, unsupported floating, or invalid contact. It applies when the scene contains contact, attachment, holding, support, or a comparable spatial relation. Each item receives $r_i\in\{0,1\}$, where one means that the stated non-penetration condition holds:
\begin{equation}
S_{\mathrm{IP}}=\frac{1}{|\mathcal K|}\sum_{i\in\mathcal K}r_i.
\end{equation}
Inapplicable items are excluded. This is a visibility-based judgment of object interaction rather than a complete three-dimensional collision simulation.

\subsubsection{Controllable Interaction and Counterfactuals}

\paragraph*{\textnormal{\textbf{Trajectory Accuracy (TA).}}} This metric evaluates W2--W3 translation commands against DA3-estimated camera motion. W/S/A/D and diagonals define target direction $\mathbf g_k$ for action segment $k$. DA3 camera-to-world matrices provide positions $\mathbf p_t$, from which segment displacement and directional agreement are computed as
\begin{align}
\Delta\mathbf p_k&=\sum_{t\in k}(\mathbf p_{t+1}-\mathbf p_t),\\
z_k&=\mathbb 1\!\left[
\frac{|\Delta\mathbf p_k^{\top}\mathbf g_k|}
{(\|\Delta\mathbf p_k\|_2+\epsilon)(\|\mathbf g_k\|_2+\epsilon)}\ge0.5
\right],\\
S_{\mathrm{TA}}&=\frac{1}{K}\sum_{k=1}^{K}z_k.
\end{align}
The percentage is $100S_{\mathrm{TA}}$, and zero motion counts as incorrect. TA measures segment-level translational direction agreement rather than exact position, travel distance, speed, or rotation.

\paragraph*{\textnormal{\textbf{Action Execution (AE).}}} AE is reported for W2--W3 and evaluates whether commanded actions are visibly executed in the correct temporal segments. Each command is decomposed into observable completion conditions, such as the intended movement, interaction, or resulting state. An applicable condition receives $a_i\in\{0,1\}$, where one means that the action requirement is visibly satisfied. AE is the mean over all applicable action conditions:
\begin{equation}
S_{\mathrm{AE}}=\frac{1}{|\mathcal A|}\sum_{i\in\mathcal A}a_i.
\end{equation}
Missing target objects, absent actors, incorrect actions, and unmet target states count as failures rather than being excluded. In W4, event-level interaction is evaluated by IV and SF instead of TA and AE.

\paragraph*{\textnormal{\textbf{Interaction Validity (IV).}}} For W4, IV measures whether intervention proceeds through a physically and causally admissible process. Items come from case red lines and cover valid contact, support, occlusion, force, transition continuity, and absence of teleportation, tearing, deformation, or penetration. IV deliberately excludes target achievement, persistence, and untouched-content preservation to avoid double penalties with SF. With $v_i\in\{0,0.5,1\}$, IV is
\begin{equation}
\mathrm{IV}=\frac{1}{|\mathcal V|}\sum_{i\in\mathcal V}v_i.
\end{equation}

\paragraph*{\textnormal{\textbf{State Fidelity (SF).}}} For W4, SF has three subclasses: achievement asks whether the target state was established; persistence asks whether it survived until the next legitimate modification; preservation asks whether content not named by events remained unchanged. For $k\in\{A,P,R\}$, SF is
\begin{equation}
S_k=\frac{1}{|\mathcal S_k|}\sum_{i\in\mathcal S_k}v_i,
\qquad
\mathrm{SF}=\frac{1}{|\mathcal K_{\mathrm{valid}}|}
\sum_{k\in\mathcal K_{\mathrm{valid}}}S_k.
\end{equation}
This subclass-balanced aggregation prevents a large number of achievement items from overwhelming persistence or preservation.

\section{Spatial World Model Track: Detailed Metrics}
\label{supp:recon-specification}

This appendix specifies the metrics, evaluation procedures, and aggregation rules for the Spatial World Model Track. It covers 39 quantities, including the 31 retained in the main performance tables after selecting F1 composites within consistency--coverage families. Metrics are grouped by observable quality (W1), physical usability (W2), scene- and object-level consistency (W3), controlled editing (W4), and expansion and preservation (W5). Operation-specific preservation is evaluated within W4 or W5 rather than counted again under W3. W6 is unscored.

\begin{table}[htbp]
\centering\small
\caption{Metric inventory by evaluation group.}
\begin{tabular}{lll}
\toprule
Group & IDs & Number \\
\midrule
Observable Quality (W1) & M01--M07, M09--M11 & 10 \\
Physical Usability (W2) & M08, M25 & 2 \\
Scene/Place Consistency (W3) & M12--M15, M38--M39 & 6 \\
Object Consistency (W3) & M16--M24 & 9 \\
Controlled Editing (W4) & M31--M37 & 7 \\
Expansion and Preservation (W5) & M26--M30 & 5 \\
Unified World Model (W6): unscored & None & 0 \\
\bottomrule
\end{tabular}
\end{table}

\subsection{Shared Notation, Actions, and Evaluation Inputs}
\label{supp:recon-common}

Let $S$ be the reconstructed scene, $V_t$ a camera pose, and $(I_t,D_t,\alpha_t)=\mathcal R(S,V_t)$ the RGB, depth, and raw opacity outputs. Let $\Omega$ be the pixel domain, $c$ a text condition, and $\mathbf1[\cdot]$ an indicator. RGB is normalized to $[0,1]$ for error calculations. Define cosine similarity $\operatorname{cos}(x,y)=x^\top y/(\|x\|_2\|y\|_2)$ for nonzero vectors and the harmonic combination
\begin{equation}
H(p,r)=
\begin{cases}
2pr/(p+r),&p+r>0,\\
0,&p=r=0.
\end{cases}
\end{equation}
An overbar denotes the mean over the specified valid observations. For explicit formalization we write population standard deviation as
\begin{equation}
\sigma(x_1,\ldots,x_n)=
\sqrt{\frac1n\sum_{j=1}^n(x_j-\overline x)^2}.
\end{equation}
All dispersion scores use population standard deviation (ddof=0). Conditional means on empty sets are undefined. Missing targets have zero recall and zero associated composites; an unavailable evaluator is a separate error status.

\paragraph{Camera sampling for W3 consistency.}
For an interactive reconstruction, include the observer pose in the exposed state, $X_t=(S,V_t)$. A movement action changes $V_t$ even when $S$ remains static:
\begin{equation}
X_{t+1}=(S,\mathcal C(V_t,a_t,\Delta t)),\qquad
O_{t+1}=\mathcal R(S,\mathcal C(V_t,a_t,\Delta t)).
\end{equation}
For the finite WASD interface, W/S specify forward/backward translation and A/D lateral translation in the controller's local frame. A mathematical free-motion convention is
\begin{equation}
\begin{aligned}
p_{t+1}&=p_t+v\Delta t\,R_t d(a_t),\\
d(\mathrm W)&=(0,1,0)^\top,\qquad d(\mathrm S)=(0,-1,0)^\top,\\
d(\mathrm A)&=(-1,0,0)^\top,\qquad d(\mathrm D)=(1,0,0)^\top.
\end{aligned}
\end{equation}
Here $R_t$ maps the controller-local frame to world coordinates and $v=1$\,m/s is the translation speed. This specifies free-motion camera updates; collision handling is a separate controller setting. Origin panoramas and object orbits are separately prescribed inspection cameras; they are not asserted to be realizable by WASD translation alone.

The camera controller supplies observations for W3 consistency, not an additional W2 metric. A renderer following prescribed poses does not independently measure pose-estimation error, action latency, or learned physical evolution. Action timestamps, requested and realized poses, and controller settings are required to distinguish camera-interface behavior from scene limitations.

\paragraph{Coordinates, scale, and geometry.}
The initial camera anchors a right-handed frame with X right, Y forward, and Z up. If valid downward ray-hit distances are $d_{\downarrow}$, the scale convention is
\begin{equation}
s=\frac{1.6}{\operatorname{median}(d_{\downarrow})},
\qquad
s_{\mathrm{fallback}}=\frac{5}{\operatorname{percentile}_{90}(d)}.
\end{equation}
Nine downward rays are used for the floor anchor. For the fallback, render the origin ring with the provided sky mask excluded when available, pool rendered depths greater than $10^{-6}$ across frames, and set their 90th percentile to 5\,m. If no positive depth is available, scale estimation fails. These are scale conventions rather than measurements of real-world size.

Navigation preprocessing crops to [ground$-2$\,m, ground$+10$\,m], removes lifted sky geometry, and reverses all faces when less than half of surface area has normals facing the initial camera. Z-up geometry is transformed to the simulator's coordinate convention. An XY footprint $F$ is estimated from area-uniform mesh samples on a grid of size $\delta=0.05$\,m, giving $A(F)=\delta^2|F|$. For mesh footprints, we sample the surface with seed 0 and a budget of $16A_{\mathrm{proj}}/\delta^2$ points, bounded between 1,000 and 20,000,000, where $A_{\mathrm{proj}}$ is the sum of horizontally projected triangle areas. Occupied grid cells are then deduplicated. Before/after operations use one original scale, coordinate transform, and grid origin.

\paragraph{Common execution procedure.}
\begin{enumerate}[leftmargin=*,nosep]
\item Import the representation, retain the source asset, and log any geometry conversion.
\item Establish coordinates and scale, recording the selected normalization branch.
\item Cache origin views and six-face probes, action trajectories, orbit cameras, and task annotations.
\item Render RGB/depth/raw opacity; construct cleaned collision geometry and common-grid footprints.
\item Evaluate eligible tasks with frozen scorers and preserve components, counts, and failure statuses.
\end{enumerate}

\subsection{W1: Generative Construction}
\label{supp:recon-axis-a}
For ordinary origin-view scores, let $\mathcal V_0$ contain $T$ prescribed panoramic frames. Except where a specific subset is stated, score each frame and average. Resolution, field of view, color processing, background composition, scorer prompts, and checkpoints are part of the evaluation definition. These ten metrics evaluate rendered appearance and alignment with the conditioning inputs.

\paragraph{M01. \texttt{topiq\_nr} ($\uparrow$).}
TOPIQ~\citep{spatial_topiq} is a learned no-reference quality regressor. With the KonIQ checkpoint,
\begin{equation}
Q_{\mathrm{TOPIQ}}=\frac1T\sum_{t=1}^T f_{\theta,\mathrm{KonIQ}}(I_t).
\end{equation}
\emph{Procedure:} preprocess frames using the frozen checkpoint; run the regressor; average scalar outputs. Multiscale semantic features guide quality estimation, so the scorer is not reducible to a pixel-error formula. Regression outputs are not guaranteed to be bounded probabilities.

\paragraph{M02. \texttt{laion\_aes} ($\uparrow$).}
LAION-Aesthetics~\citep{spatial_laion} applies a learned head to normalized CLIP image embeddings:
\begin{equation}
Q_{\mathrm{aes}}=\frac1T\sum_t
g_\theta\!\left(\frac{\phi_I(I_t)}{\|\phi_I(I_t)\|_2}\right).
\end{equation}
\emph{Procedure:} encode each frame, normalize its embedding, apply the frozen aesthetic head, and average. Aesthetic preference complements rather than verifies scene correctness.

\paragraph{M03. \texttt{pi} ($\downarrow$).}
The perceptual index~\citep{spatial_pi} combines Ma's learned quality estimate and NIQE naturalness:
\begin{equation}
\mathrm{PI}(I)=\tfrac12(10-\mathrm{Ma}(I)+\mathrm{NIQE}(I)),
\qquad Q_{\mathrm{PI}}=\frac1T\sum_t\mathrm{PI}(I_t).
\end{equation}
At NIQE's statistical-distance stage,
\begin{equation}
\mathrm{NIQE}(I)=
\sqrt{(\mu_I-\mu_N)^\top
\left[(\Sigma_I+\Sigma_N)/2\right]^\dagger
(\mu_I-\mu_N)},
\end{equation}
where $\mu,\Sigma$ are image/reference natural-scene feature statistics and $\dagger$ denotes a pseudoinverse. \emph{Procedure:} extract NIQE features with fixed reference parameters, infer Ma, combine per image, and average. Patch extraction, scales, normalization, and the Ma checkpoint are required inputs. PI is not standalone NIQE.

\paragraph{M04. \texttt{q\_align} ($\uparrow$).}
For Q-Align~\citep{spatial_qalign}, let $\ell_k(I)$ be logits for five quality labels ordered from bad to excellent:
\begin{equation}
Q_{\mathrm{QA}}=\frac1T\sum_t\sum_{k=1}^5
k\,\frac{\exp\ell_k(I_t)}{\sum_{j=1}^5\exp\ell_j(I_t)}.
\end{equation}
\emph{Procedure:} ask the fixed image-quality question, obtain label logits, normalize over the five labels, take their score expectation, and average frames. The resulting range is $[1,5]$ under this formulation. Freeze the image-quality variant rather than switching to aesthetic or video scoring.

\paragraph{M05. \texttt{hpsv3} ($\uparrow$).}
Using the frozen text-conditioned HPSv3 preference model~\citep{ma2025hpsv3widespectrumhumanpreference},
\begin{equation}
Q_H=\frac1T\sum_t\mu_\theta(c,I_t).
\end{equation}
\emph{Procedure:} pair each frame with the prescribed evaluation text, run HPSv3, select its mean reward, and average. Rewards may be negative and are not probabilities; HPSv2 cosine scoring is not an equivalent substitution. The same frozen scorer is reused in dispersion and expansion diagnostics.

\paragraph{M06. \texttt{vlm\_score} ($\uparrow$).}
Using Qwen3-VL~\citep{spatial_qwen3vl} on 18 views, let $a_t,s_t,l_t$ be binary artifact-free, sharpness, and plausible-layout judgments:
\begin{equation}
Q_{\mathrm{VLM}}=\frac1{54}\sum_{t=1}^{18}(a_t+s_t+l_t).
\end{equation}
\emph{Procedure:} sample the prescribed 18 frames, apply the fixed three-part rubric, parse binary responses, and average judgments. Retain the three component rates. The three-part rubric is benchmark-specific.

\paragraph{M07. \texttt{hole\_rate} ($\downarrow$).}
\begin{equation}
h_t=\frac1{|\Omega|}\sum_{u\in\Omega}\mathbf1[\alpha_t(u)<0.5],
\qquad h=\frac1T\sum_t h_t.
\end{equation}
\emph{Procedure:} use renderer opacity before background compositing, threshold at 0.5, average pixels, then views. This measures missing rendered content, not mesh watertightness: an opaque incorrect wall can have zero holes, and intentional empty sky can increase the rate. The definition is benchmark-specific.

\paragraph{M09. \texttt{clip\_t} ($\uparrow$).}
We measure text--image alignment using CLIP~\citep{radford2021clip}:
\begin{equation}
C_t=\frac1T\sum_t\operatorname{cos}(\phi_I(I_t),\phi_T(c)).
\end{equation}
\emph{Procedure:} encode the full scene text and every origin view using matching CLIP towers, calculate cosine similarity, and average. This is scene semantic alignment, not a guarantee that every object or relationship is correct.

\paragraph{M10. \texttt{clip\_i} ($\uparrow$).}
For input image $I_*$ and $s_t=\operatorname{cos}(\phi_I(I_t),\phi_I(I_*))$,
\begin{equation}
C_i=\frac13\sum_{t\in\operatorname{Top3}(s)}s_t.
\end{equation}
\emph{Procedure:} encode the input and origin views, select the three most similar views, and average. Insufficient views require an explicit validity rule. Best-view selection addresses orientation mismatch but is not whole-world fidelity. 

\paragraph{M11. \texttt{object\_coverage} ($\uparrow$).}
For $J$ requested categories and six origin probes,
\begin{equation}
C_{\mathrm{obj}}=\frac1J\sum_{j=1}^J
\mathbf1[\exists v\in\{1,\ldots,6\}:\operatorname{det}(j,I_v)=1].
\end{equation}
\emph{Procedure:} test each requested category in all six views, union detections by category, and divide matched categories by $J$. Empty annotation is inapplicable; an annotated but absent target is a failure. The output measures named-category presence rather than exact instance counts.

\subsection{W2: Interactive Simulation}
\label{supp:recon-axis-d}
The two W2 metrics evaluate physical usability through connected navigation area and stable object support. Both operate on imported scene geometry: navigation is assessed with a Habitat-Sim NavMesh~\citep{spatial_habitat_navmesh}, and support is tested through external simulation.

\paragraph{M08. \texttt{navigable\_ratio} ($\uparrow$).}
Let $N_k$ denote the four-connected components of the horizontal occupancy grid obtained by projecting and rasterizing all NavMesh polygons:
\begin{equation}
R_{\mathrm{nav}}=\frac{\max_k A(N_k)}{A(F)}.
\end{equation}
\emph{Procedure:} clean/import geometry, build a NavMesh, project its polygons onto the common horizontal grid, identify four-connected grid components, and divide the largest component area by footprint area. Cells touching only at a corner are disconnected. Vertically overlapping surfaces share the same horizontal cells. A valid scene with no navigable component has numerator zero; an empty footprint is invalid.

Agent parameters are radius 0.1\,m, height 1.5\,m, maximum step 0.2\,m, and maximum slope $45^\circ$. Nominal horizontal/vertical voxel dimensions are 0.05/0.2\,m. If a NavMesh cannot be built at 0.05\,m resolution, the implementation retries at 0.10 and 0.20\,m and records the build settings. The largest component need not contain the initial observer position.

\paragraph{M25. \texttt{tabletop\_ready\_rate} ($\uparrow$).}
For scene $i$, the placement planner requests six locations on the annotated support surface. It selects spatially separated positions where the plate footprint fits and removes candidates with no underlying surface. Let $n_i$ be the number of retained placement trials and $b_{ij}$ their binary outcomes. The implemented scene score is
\begin{equation}
T_i=\begin{cases}
\displaystyle\frac1{n_i}\sum_{j=1}^{n_i}b_{ij},&n_i>0,\\
0,&n_i=0,
\end{cases}
\qquad
T=\frac1{N_{\mathrm{support}}}\sum_i T_i.
\end{equation}
Thus, scenes with six retained trials use a denominator of six; otherwise the denominator is the retained trial count. A scene with no usable mesh, scale anchor, support surface, or placement location receives zero. Missing runs remain unavailable rather than being assigned this failure score.

\emph{Procedure:} locate the annotated support, select placement locations, and independently release the same plate at each location. The plate is a cylinder of diameter 0.22\,m and thickness 0.02\,m, with density 1,800\,kg/m$^3$ and friction coefficient 0.5. The scene friction coefficient is 0.8. Each trial starts 0.01\,m above the surface and runs for 5\,s in PyBullet with gravity $-9.81$\,m/s$^2$ and timestep $1/240$\,s and 150 solver iterations per step.

A trial fails if it has initial penetration greater than 0.002\,m, if the final plate bottom is more than 0.10\,m below the support surface, or if the final plate center lies outside the support boundary with a 0.02\,m tolerance. Trials without usable collision geometry also fail. Horizontal displacement and tilt are recorded but do not determine success: placement success here means that the plate remains on the support under these rules.

\subsection{W3: State Persistence}
Fifteen consistency quantities evaluate observations across camera motion and viewpoint changes, partitioned into six scene-level (place) and nine object-level measurements.

\subsubsection{Scene-Level (Place) Metrics}
\label{supp:recon-axis-b}
The action-driven path supplies the view sequence; these scores measure the spatial validity, short-range visual continuity, and location robustness of its observations. All six quantities belong to W3 place-level consistency. M15 and M38--M39 sample additional locations rather than independently measuring action-direction error.

\paragraph{M12. \texttt{oob\_ratio} ($\downarrow$).}
Let $h_t^{(6)}$ be six-face hole rate at pose $V_t$, $\widetilde F$ the filled footprint, and $\operatorname{free}(D_t)$ the depth-based clearance predicate:
\begin{align}
o_t&=\mathbf1[h_t^{(6)}>\min(h_0^{(6)}+0.10,0.5)]
       \mathbf1[XY(V_t)\notin\widetilde F],\\
R_{\mathrm{OOB}}&=\frac1T\sum_t
\mathbf1[o_t=1\ \lor\ \neg\operatorname{free}(D_t)].
\end{align}
\emph{Procedure:} execute the prescribed translated path and render six $256\times256$ probes with $90^\circ$ fields of view at each sample. The clearance test pools pixels with rendered depth greater than $10^{-6}$ across the six faces and fails when more than half of those valid-depth pixels are closer than 0.10\,m. It is evaluated when a metric scale is available. If the filled footprint is unavailable, the outside-scene test uses the hole-rate threshold alone. The combined test is a depth and coverage proxy for invalid camera positions.

\paragraph{M13. \texttt{brightness\_consistency} ($\uparrow$).}
Let $b_t$ be a normalized grayscale histogram with bins [0,63], [64,191], and [192,255]. Define the modified-softmax transform
\begin{equation}
g_\lambda(x)=\frac{\exp(\lambda\operatorname{clip}(x,0,1))-1}{\exp(\lambda)-1+\epsilon},
\qquad \lambda=15,\quad\epsilon=10^{-8}.
\end{equation}
Histogram cosine similarity uses $(\|x\|_2+\epsilon)(\|y\|_2+\epsilon)$ in the denominator for numerical stability. Brightness consistency is
\begin{equation}
B=\frac1T\left[1+\sum_{t=1}^{T-1}
g_{15}(\operatorname{cos}(b_t,b_0))\right].
\end{equation}
\emph{Procedure:} histogram every path frame, including OOB frames; compare to the initial histogram; apply $g_{15}$; average with first-frame score one. Coarse histogram continuity can remain high despite object rearrangement.

\paragraph{M14. \texttt{color\_temperature\_constraint} ($\uparrow$).}
For normalized HSV-H histograms $u_t$ using bins 0--15, 16--30, 31--60, 61--90, 91--120, 121--150, and 151--179,
\begin{equation}
c_t=\tfrac12\{\operatorname{cos}(u_t,u_0)+
\operatorname{cos}(u_t,u_{t-1})\},\qquad
C=\frac1T\left[1+\sum_{t=1}^{T-1}g_{15}(c_t)\right].
\end{equation}
\emph{Procedure:} convert every path frame to the 0--179 hue convention, histogram, compare with first/previous views, transform, and average. RGB frames are converted with OpenCV's RGB-to-HSV conversion; all pixels contribute to the histogram without an additional saturation mask. This measures palette continuity rather than kelvin temperature.

\paragraph{M15. \texttt{place\_hpsv3\_std} ($\downarrow$).}
For mean HPSv3 values $q_k$ at sampled locations including the origin,
\begin{equation}
S_{\mathrm{place}}=\sigma(q_0,\ldots,q_K),\qquad
q_k=\frac1{|\mathcal V_k|}\sum_{V\in\mathcal V_k}\mu_\theta(c,I(V)).
\end{equation}
\emph{Procedure:} estimate median origin depth $d$ and consider four radii $d\{1.5,1.0,0.6,0.3\}$ in outside-in order. Each radius has a quota of two accepted locations at the original camera height, drawn from eight bearings spaced $45^\circ$ apart. For radius index $j\in\{0,1,2,3\}$, try bearings $j$ and $j+4$ first, then the remaining indices in ascending order until two locations pass the M12 usability test or all eight directions have been tried. No unused quota is transferred to another radius. Thus, up to eight locations are retained, although more than eight candidates may be probed. Compute HPSv3 at the accepted locations and take population standard deviation over their successfully scored ring means together with the origin. Rejected candidates are excluded from the dispersion. At least one scored additional location and a scored origin are required.

\paragraph{M38. \texttt{place\_recall} ($\uparrow$).}
The sampler has eight location slots: two at each of four radii. With $u_k$ indicating an occupied slot whose location passes the usability test, the geometric sampling recall is
\begin{equation}
r_{\mathrm{usable}}=\frac18\sum_{k=1}^{8}u_k.
\end{equation}
The current summary implementation uses successfully scored accepted locations. If $e_k$ indicates that slot $k$ also has an available HPSv3 ring score, then, for a successfully scored origin,
\begin{equation}
r_{\mathrm{place}}=\frac18\sum_{k=1}^{8}u_ke_k.
\end{equation}
The two recalls agree when all accepted locations are scored. The origin is included in M15 dispersion but is not a ninth recall slot. If the sampling or HPS section is absent or explicitly marked unavailable, recall is missing. Other failed HPS sections, including a failed origin score or no scored additional location, produce zero in the current summary implementation; this reporting behavior can therefore include evaluator failures as well as unavailable views. M39 uses this reported $r_{\mathrm{place}}$.

\paragraph{M39. \texttt{place\_hpsv3\_f1} ($\uparrow$).}
For scene $i$, combine its M15 dispersion $s_i$ with its M38 recall $r_i$:
\begin{equation}
p_i=\frac1{1+s_i},\qquad
F_{\mathrm{place},i}=H(p_i,r_i)
=\frac{2r_i}{1+r_i(1+s_i)},\qquad
F_{\mathrm{place}}=\frac1{N_{\mathrm{valid}}}\sum_iF_{\mathrm{place},i}.
\end{equation}
The expression applies when $s_i$ is defined. Zero reported recall gives a zero composite, including the failed-section cases described in M38; missing recall yields a missing composite. The summary uses $p_i=1$ if dispersion is absent, which leaves the composite zero when recall is zero. \emph{Procedure:} compute location-level mean HPS rewards, take dispersion including the origin, combine inverse dispersion with the same scene's reported recall, then average the available scene composites.

Here \emph{precision} means inverse dispersion, not true-positive precision, and the F1 label denotes a harmonic consistency--coverage score. It cannot distinguish consistently good from consistently poor images without an absolute-quality measurement. A scene with uniform observations at only a few usable locations can have low dispersion but low recall; the composite penalizes this limited coverage. Conditional components can use different valid populations, so their model-level aggregates do not establish a matched-subset quality ranking.

\paragraph{Action-path algorithm.}
\begin{enumerate}[leftmargin=*,nosep]
\item Reset to the original observer pose and load the timed action/camera manifest.
\item Apply each movement step, render its observation, and retain requested and realized poses when available.
\item Compute M12 from six-face probes and clearance. Use all path frames for M13--M14.
\item Construct the eight-location plan, retain every usability outcome, and calculate M38.
\item Score the sampled location set including the origin for M15; combine M15/M38 per scene for M39.
\item Return six metrics, sampled cameras, quality-set membership, valid counts, and execution status.
\end{enumerate}

\subsubsection{Object-Level Metrics}
\label{supp:recon-axis-c}
Use a fixed-radius 24-view orbit per annotated object, without OOB/collision-based camera filtering or radius search. Missing observations can reflect geometry, occlusion, camera placement, or segmentation. The subgroup tests W3 observation persistence and return correspondence.

\paragraph{M16. \texttt{obj\_sam\_recall} ($\uparrow$).}
With $d_t$ indicating a SAM3 target detection and $n=\sum_{t=1}^{24}d_t$,
\begin{equation}
r_s=\mathbf1[n\geq2]\,n/24.
\end{equation}
\emph{Procedure:} segment each orbit frame with SAM3~\citep{carion2025sam3segmentconcepts} using the target description, count detected views, return zero if fewer than two, otherwise divide by 24. This is detected-view coverage rather than recall against exhaustively labeled masks.

\paragraph{M17. \texttt{obj\_hpsv3\_std} ($\downarrow$).}
For detected mask $M_t$, suppress background content by replacing it with the mean target color:
\begin{equation}
\widetilde I_t(u)=
\begin{cases}
I_t(u),&u\in M_t,\\
|M_t|^{-1}\sum_{v\in M_t}I_t(v),&u\notin M_t,
\end{cases}
\quad s_o=\sigma\{\mu_\theta(c_o,\mathcal K(\widetilde I_t,M_t)):t\in\mathcal D_o\}.
\end{equation}
Here $\mathcal K$ crops to the mask bounding box with 8 pixels of padding, clipped to the image boundaries, then uses bicubic resizing to make the short side 224 pixels if it was smaller. Masks with fewer than 64 pixels are not scored; $\mathcal D_o$ contains the views with valid crops and available HPSv3 scores. \emph{Procedure:} fill the background with the mean target color, crop and resize, score with the frozen object prompt, and calculate dispersion when at least two crops have scores. Otherwise dispersion is undefined. This statistic measures variation in perceptual quality across object views.

\paragraph{M18. \texttt{obj\_hpsv3\_f1} ($\uparrow$).}
\begin{equation}
p_h=(1+s_o)^{-1},\qquad
F_H=\begin{cases}H(p_h,r_s),&r_s>0,\\0,&r_s=0.\end{cases}
\end{equation}
\emph{Procedure:} transform dispersion into inverse-dispersion quality, combine with detected-view coverage, then average object scores. The proxy $p_h$ is not true-positive precision and may reward uniformly low quality. Its scale depends on the frozen HPSv3 model.

\paragraph{M19. \texttt{obj\_track\_accuracy} ($\uparrow$).}
Rotate the orbit sequence to begin at a SAM3-visible frame and append that same frame as the 25th observation. TAPIP3D~\citep{spatial_tapip3d} tracks 64 sampled target points in persistent world coordinates:
\begin{equation}
A_o=\frac1{64}\sum_{k=1}^{64}
\mathbf1[\|\widehat X_{k,24}-\widehat X_{k,0}\|_2\leq0.01b_o].
\end{equation}
Here $b_o$ is the diagonal length of the target object's 3D bounding box. \emph{Procedure:} choose a visible start, sample points, track with RGB/depth/cameras, and compare only the endpoints. Repeating the initial image can conceal intermediate drift; this is not full-trajectory stationarity.

\paragraph{M20. \texttt{obj\_track\_f1} ($\uparrow$).}
\begin{equation}
F_T=\begin{cases}H(A_o,r_s),&r_s>0,\\0,&r_s=0.\end{cases}
\end{equation}
\emph{Procedure:} combine each object's endpoint accuracy and SAM view coverage before taking the object mean. If recall is positive but tracking accuracy is unavailable, that object's composite is excluded from the mean; zero recall contributes a zero composite even without tracking accuracy.

\paragraph{M21. \texttt{obj\_vlm\_recall} ($\uparrow$).}
Let $v_t$ indicate a VLM judgment that the camera is inside the scene and the target is present and unoccluded:
\begin{equation}
r_v=\frac1{24}\sum_{t=1}^{24}v_t.
\end{equation}
\emph{Procedure:} judge validity for every orbit view, then divide valid views by 24. Unlike SAM recall, this formulation does not impose a two-view detection minimum. Freeze the Qwen3-VL prompt/parser. The result is judged visibility, not ground-truth detector recall.

\paragraph{M22. \texttt{obj\_vlm\_artifact\_free} ($\uparrow$).}
For binary artifact-free answer $a_t$,
\begin{equation}
a_o=\frac{\sum_t v_ta_t}{\sum_t v_t}.
\end{equation}
\emph{Procedure:} on valid views, judge floaters, spikes, blur, stretching, holes, duplication, melting, and noise; average the binary artifact-free answers. With no valid view the conditional score is undefined. This measures visible reconstruction defects.

\paragraph{M23. \texttt{obj\_vlm\_completeness} ($\uparrow$).}
For binary completeness answer $c_t$,
\begin{equation}
c_o=\frac{\sum_t v_tc_t}{\sum_t v_t}.
\end{equation}
\emph{Procedure:} judge missing, torn, fragmented, or incorrectly transparent solid parts in the same valid views, then average. Empty valid sets are undefined. Completeness is a visual rubric rather than surface recall against a reference scan.

\paragraph{M24. \texttt{obj\_vlm\_f1} ($\uparrow$).}
\begin{equation}
p_v=(a_o+c_o)/2,\qquad
F_V=\begin{cases}H(p_v,r_v),&r_v>0,\\0,&r_v=0.\end{cases}
\end{equation}
\emph{Procedure:} average the two conditional quality components, combine with valid-view coverage, and average per-object composites. Retain all four VLM components to separate poor visibility from poor visible integrity. The rubric and harmonic composite are benchmark-specific.

\paragraph{Object algorithm.}
\begin{enumerate}[leftmargin=*,nosep]
\item Render the 24 fixed orbit views and apply SAM3 to every view.
\item Compute $r_s$; if positive, compute masked HPS dispersion and $F_H$, otherwise set $F_H=0$.
\item Build the repeated-start tracking sequence and compute $A_o$ where evaluable; form $F_T$ with the zero-recall rule.
\item Judge every view's validity; compute $r_v$ and conditional integrity components.
\item Set $F_V=0$ when $r_v=0$; otherwise combine components.
\item Store all nine outputs, annotations, masks, tracks, valid counts, and error states.
\end{enumerate}

\subsection{W4: Programmable Dynamics}
\label{supp:recon-axis-e}
Let subscripts 0/1 denote before/after worlds at the same pre-edit cameras. The allowed-change mask $M$ uses the after-target for addition, before-target for deletion, the before-old/after-new union for replacement, and the before/after target union for modification. The background is $B=\Omega\setminus M$. When no target mask is generated, evaluation falls back to the whole image; an empty $B$ is invalid. All seven quantities belong to W4: edit success rate, non-target preservation, and their composites. They assess a restricted intervention interface relevant to Programmable Dynamics. Explicit physical-law, causal-mechanism, behavior, and task-constraint programming require separate verification.

\paragraph{M31. Edit success rate (\texttt{edit\_execution}, $\uparrow$).}
Object edits use up to six detectable orbit views after locating the target through six origin probes; global edits use six origin views. The per-case edit-success indicator is defined as
\begin{equation}
E=\mathbf1[\exists V\in\mathcal V_E:
\operatorname{satisfies}(I_1(V),\mathrm{instruction})=1].
\end{equation}
The reported edit success rate is the mean of this binary indicator over editing cases.
\emph{Procedure:} select views by edit type, judge before/after satisfaction with Qwen3-VL, and OR the after-view answers. If no object views are found, the score is zero. Before judgments do not gate improvement, so an already-satisfied no-op can pass. Object localization uses the edited scene for additions and replacements (the new object), and the original scene for deletions and modifications.

\paragraph{M32. \texttt{rgb\_psnr\_bg} ($\uparrow$).}
\begin{align}
\mathrm{MSE}_B&=\frac1{3|B|}\sum_{u\in B}\sum_{c=1}^3
(I_1(u,c)-I_0(u,c))^2,\\
\mathrm{PSNR}_B&=\begin{cases}
99,&\mathrm{MSE}_B\leq10^{-12},\\
-10\log_{10}\mathrm{MSE}_B,&\text{otherwise}.
\end{cases}
\end{align}
\emph{Procedure:} pair RGB views and calculate background PSNR in each frame with at least 64 background pixels. For case $i$, average the valid frame PSNR values first, then compute
\begin{equation}
\mathrm{RMSE}_{B,i}=10^{-\overline{\mathrm{PSNR}}_{B,i}/20}.
\end{equation}
M36 uses this transformed mean PSNR, rather than the arithmetic mean of per-frame RMSE values.

\paragraph{M33. \texttt{lpips\_bg} ($\downarrow$).}
Let $L(u)$ be the frozen LPIPS~\citep{spatial_lpips} spatial distance map aligned to the RGB pixel grid:
\begin{equation}
\mathrm{LPIPS}_B=\frac1{|B|}\sum_{u\in B}L(u),\qquad
L=\sum_\ell\operatorname{Resize}_\ell
\left[\sum_c w_{\ell c}
(\widehat\phi_{\ell c}(I_1)-\widehat\phi_{\ell c}(I_0))^2\right].
\end{equation}
The exact layer combination and weights are those of the chosen network. \emph{Procedure:} run spatial LPIPS on paired full images, align its map with the background mask, and average background locations only. A whole-image scalar or simply blacking out the target is not equivalent. The implementation uses spatial LPIPS-VGG without mask dilation and averages its map over the background mask. Frames with fewer than 64 background pixels are skipped; valid frame scores are averaged within each case. Feature receptive fields can cross the edit boundary.

\paragraph{M34. \texttt{dinov2\_bg\_distance} ($\downarrow$).}
For corresponding DINOv2~\citep{spatial_dinov2} patch tokens $z_{0j},z_{1j}$ and background patch set $\mathcal P_B$,
\begin{equation}
D_B=\frac1{|\mathcal P_B|}\sum_{j\in\mathcal P_B}
\left[1-\operatorname{cos}(z_{0j},z_{1j})\right].
\end{equation}
\emph{Procedure:} resize both images to $518\times518$, apply ImageNet channel normalization, and extract DINOv2-base final-layer patch tokens on the $37\times37$ grid. L2-normalize each token, retain patches whose background-mask coverage exceeds 0.5, and average one minus cosine similarity over corresponding tokens. Frames with fewer than four retained patches are skipped. The case score averages valid frame scores.

\paragraph{M35. \texttt{depth\_agree\_ratio} ($\uparrow$).}
For paired frame $t$, let $G_t=\{u:\alpha_{0t}(u),\alpha_{1t}(u)\geq0.5;\ D_{0t}(u),D_{1t}(u)>0\}$. Define
\begin{equation}
\rho(d_0,d_1)=\frac{|d_1-d_0|}{\max(\min(d_0,d_1),10^{-6})}.
\end{equation}
For frames with $|G_t|\geq64$,
\begin{equation}
a_{D,t}=\frac1{|G_t|}\sum_{u\in G_t}
\mathbf1[\rho(D_{0t}(u),D_{1t}(u))\leq0.05],
\qquad A_D=\frac1{|\mathcal T_D|}\sum_{t\in\mathcal T_D}a_{D,t},
\end{equation}
where $\mathcal T_D$ contains the valid paired frames. \emph{Procedure:} render depth and opacity from identical cameras before and after editing, compute per-frame agreement over common valid geometry, and average valid frames equally. The score is unavailable if no frame qualifies. Removed geometry outside the common-valid mask does not enter this score.

\paragraph{M36. \texttt{edit\_score\_object} ($\uparrow$).}
For object-edit case $i$,
\begin{equation}
S_{o,i}=E_i\left[1-\tfrac13
(\mathrm{RMSE}_{B,i}+\mathrm{LPIPS}_{B,i}+D_{B,i})\right],
\qquad \mathrm{Score}_o=\frac1{N_o}\sum_iS_{o,i}.
\end{equation}
\emph{Procedure:} aggregate each case's paired-view components under the frozen rule, combine its edit-success indicator and errors, then average object-edit cases. Do not calculate from model-level means. The per-case raw score is not clipped and may fall outside $[0,1]$ because the component distances have different ranges. Its model-level mean is clipped during table normalization as specified in Section~\ref{sec:recon-normalized-summary}. If $E_i=0$, the joint score is zero even when preservation is unavailable; with positive $E_i$, missing preservation yields a missing joint score.

\paragraph{M37. \texttt{edit\_score\_global} ($\uparrow$).}
\begin{equation}
S_{g,i}=E_iA_{D,i},\qquad
\mathrm{Score}_g=\frac1{N_g}\sum_iS_{g,i}.
\end{equation}
\emph{Procedure:} combine the edit-success indicator and depth agreement within each global-edit case, then average global cases. This inherits M31/M35's no-op and common-valid-mask limitations. Report it separately from the object score, which uses a different population and preservation function.

\paragraph{Editing algorithm.}
\begin{enumerate}[leftmargin=*,nosep]
\item Freeze pre-edit coordinates/cameras and select views for judging edit success by operation type.
\item Judge before/after instructions and apply the any-after-view rule.
\item For object edits, build the operation-specific allowed-change mask and compute M32--M34 on its complement.
\item For global edits, evaluate M35 on paired common-valid geometry.
\item Form M36 or M37 per case; retain the edit-success indicator, preservation components, and validity counts.
\end{enumerate}
A visual material edit measures appearance control; physical changes such as altered friction or mass require simulator-based verification.

\subsection{W5: Scalable Shared World}
\label{supp:recon-axis-f}
Use common-frame before/after footprints $F_0,F_1$, with new cells $F_+=F_1\setminus F_0$. All methods use the area-weighted centroid of the newly occupied cells as the expansion sampling location. Its camera height is the expanded ground height at that location plus the original camera height. The legacy forward search for the old boundary is used only for diagnostics, not for the reported evaluation.

Fix origin views $\mathcal V_0$, full-circle views $\mathcal V_J$ at the expansion sampling location, and path views $\mathcal V_P$ from the original camera position to that location. Each before/after pair uses identical cameras. These five metrics evaluate spatial growth, rendered coverage, and view quality under expansion.

\paragraph{M26. \texttt{expand\_area\_ratio} ($\uparrow$).}
\begin{equation}
R_A=\frac{A(F_1)}{A(F_0)}.
\end{equation}
\emph{Procedure:} rasterize both footprints in the same grid, calculate areas, and divide after by before. One means unchanged total area. The ratio is not $|F_+|/|F_0|$, and loss of old cells can coexist with growth. Average per-pair ratios rather than substituting pooled areas.

\paragraph{M27. \texttt{junction\_coverage\_gain} ($\uparrow$).}
Using M07 hole rate under shared junction cameras,
\begin{equation}
G_J=\frac1{|\mathcal V_J|}\sum_{V\in\mathcal V_J}
[h_0(V)-h_1(V)].
\end{equation}
\emph{Procedure:} render paired junction panoramas, compute opacity coverage, and average after-minus-before coverage. Use the same opacity threshold of 0.5 as M07. Positive gain denotes additional observed content, not surface correctness or traversable connectivity.

\paragraph{M28. \texttt{new\_region\_hpsv3} ($\uparrow$).}
\begin{equation}
Q_{\mathrm{new}}=\frac1{|\mathcal V_J|}\sum_{V\in\mathcal V_J}
\mu_\theta(c,I_1(V)).
\end{equation}
\emph{Procedure:} score expanded-world junction views using the frozen HPSv3 prompt/checkpoint and average. This is absolute quality at the chosen proxy views rather than a uniform sample over all new cells.

\paragraph{M29. \texttt{old\_region\_hpsv3\_delta} ($\uparrow$).}
\begin{equation}
\Delta Q_{\mathrm{old}}=\frac1{|\mathcal V_0|}\sum_{V\in\mathcal V_0}
[\mu_\theta(c,I_1(V))-\mu_\theta(c,I_0(V))].
\end{equation}
\emph{Procedure:} render both worlds at the same original-origin views, score with identical settings, and average paired after-minus-before differences. Negative values indicate lower preference after expansion. Equal quality rewards alone do not prove unchanged geometry or identity.

\paragraph{M30. \texttt{connect\_hpsv3\_delta} ($\uparrow$).}
\begin{equation}
\Delta Q_{\mathrm{path}}=\frac1{|\mathcal V_P|}\sum_{V\in\mathcal V_P}
[\mu_\theta(c,I_1(V))-\mu_\theta(c,I_0(V))].
\end{equation}
\emph{Procedure:} sample 16 frames along the path from the original camera position to the center of the newly occupied region, render both worlds at the same poses, and average the after-minus-before HPSv3 differences. This measures the change in view quality along the path.

\paragraph{Expansion algorithm.}
\begin{enumerate}[leftmargin=*,nosep]
\item Reuse the original coordinate/scale transform for both worlds and rasterize common-grid footprints.
\item Locate the center of newly occupied cells and set its camera height using the expanded ground and original camera height.
\item Freeze origin, junction, and connecting-path cameras.
\item Compute M26--M27 for growth/coverage and M28--M30 for appearance/preservation.
\item Return paired components, selected locations, sample counts, and failure statuses.
\end{enumerate}
The tables report normalized expansion scores rather than raw area ratios or signed HPS changes; the mappings are given in Section~\ref{sec:recon-normalized-summary}.

\subsection{Aggregation and Reproducibility}
\label{supp:recon-audit}

\paragraph{Aggregation and missingness.}
The hierarchy is frames/trials to object/case components, then object/case composites, then model means. In general,
\begin{equation}
\overline{H(p,r)}\ne H(\overline p,\overline r),
\qquad
\overline{10^{-P/20}}\ne 10^{-\overline P/20}.
\end{equation}
Preserve metric-specific valid counts; 266 nominal snapshot scenes, 424 annotated objects, 72 support scenes, and operation subsets are not interchangeable denominators. N/A does not identify whether an interface was unsupported, untested, or invalid. Keep these statuses separate from measured zeros, and do not drop absent annotated objects as unevaluable successes.

\paragraph{Evaluation settings.}
The implementation uses population standard deviation, the fixed histogram transform in M13, four-connected navigation grids, and the placement predicate in M25. Object quality uses masked crops (M17), while object editing compares corresponding background regions (M32--M34). Expansion uses the same new-region-center sampling rule for all methods (M26--M30). Per-case reports retain component scores, valid counts, and failure statuses.

\paragraph{Interpretation.}
The reported results are descriptive model averages. The present evaluation measures scene quality and the specified spatial operations; it does not directly measure downstream agent task success.

\subsection{Normalized World-Level Summary}
\label{sec:recon-normalized-summary}

Raw metric summaries are mapped to $[0,1]$ with higher values better, then multiplied by 100 for display. The paper tables show two decimal places. Elo ratings retain their original scale. The mappings below are applied to raw model-level metric means, after the within-case and across-case aggregation described above.

\paragraph{Arena Elo with unsupported capabilities.}
\label{app:elo-calculate}
We compute Arena Elo from human pairwise comparisons across W1, W3,
W4, and W5 using Bradley--Terry maximum likelihood estimation,
with the mean rating normalized to 1000.
Since ratings based only on observed comparisons do not penalize
unsupported capabilities, we supplement human votes with deterministic
outcomes. For each prompt and model pair where at least one model
does not support the evaluated capability, we add one comparison
in each presentation order. A supported model wins against an
unsupported model; if neither model supports the capability,
the outcome is recorded as ``both bad.''
Each added comparison receives the same weight as one human vote,
and both ``both good'' and ``both bad'' outcomes contribute
half a point to each model.
The resulting ratings reflect both human preference and capability
coverage under this rule.

\paragraph{Selection and aggregation.}
We retain only F1 within each consistency--coverage family, leaving 10, 2, 7, 7, and 5 metrics for W1--W5. The omitted W3 components are location recall and its HPS consistency component, object SAM recall and its HPS consistency component, tracking accuracy, and VLM recall, artifact freedom, and completeness. W4 retains one edit-success metric, four preservation metrics, and two joint edit scores. For baseline $b$, group $w$, and the set $A_{b,w}$ of available retained metrics, let $z_{b,m}\in[0,1]$ be a normalized score. The displayed group score is
\begin{equation}
S_{b,w}=\frac{100}{|A_{b,w}|}\sum_{m\in A_{b,w}}z_{b,m}.
\end{equation}
Metrics receive equal weight, independent of their valid sample counts. FlashWorld lacks PI and therefore uses nine W1 metrics; other evaluated groups use the complete retained set. An empty group is shown as a dash, and W6 is unscored. Missing metric summaries are excluded from the group mean; the explicit failure scores described above remain included.

\paragraph{Fixed normalization rules.}
Write $C(x)=\min(1,\max(0,x))$. Table~\ref{tab:spatial-fixed-normalization} lists transformations with metric-defined parameters. A final clipping operation keeps every normalized score in $[0,1]$, including raw edit composites or regression outputs that fall outside that interval.

\begin{table}[htbp]
\centering\small
\caption{Normalization rules with fixed, metric-defined parameters. Here $x$ is a raw model-level metric summary.}
\label{tab:spatial-fixed-normalization}
\begin{tabular}{@{}p{0.46\linewidth}p{0.50\linewidth}@{}}
\toprule
Metric & Normalized score \\
\midrule
AES & $C((x-1)/9)$ \\
Q-Align & $C((x-1)/4)$ \\
CLIP-T, CLIP-I, junction coverage gain & $C((x+1)/2)$ \\
Hole rate, OOB ratio & $1-C(x)$ \\
Background DINO distance & $1-C(x/2)$ \\
Background RGB PSNR & $C(1-10^{-x/20})$ \\
Expansion area ratio & $C(1-1/x)$ for $x>0$; 0 otherwise \\
Location/object HPS standard deviation & $1/(1+\max(0,x))$ \\
Other bounded higher-is-better scores & $C(x)$ \\
\bottomrule
\end{tabular}
\end{table}

\paragraph{Frozen reference-based rules.}
PI, HPSv3, new-region HPSv3, and background LPIPS use an affine mapping with bounds $L,U$ fitted once from the 5th and 95th percentiles of a reference baseline-by-case distribution:
\begin{equation}
z=\begin{cases}
C((x-L)/(U-L)),&\text{higher raw values are better},\\
1-C((x-L)/(U-L)),&\text{lower raw values are better}.
\end{cases}
\end{equation}
Old-region and connecting-path HPSv3 changes use
\begin{equation}
z=\tfrac12+\tfrac12\tanh(x/a),
\end{equation}
where $a$ is the reference 90th percentile of the absolute change. The constants in Table~\ref{tab:spatial-frozen-normalization} are frozen in \texttt{wmbench-normalization-v1} and are not refitted when adding a method. The reference is the 2,530-row baseline-by-case summary for the \texttt{0827\_validated\_eval} evaluation set; metrics have different valid reference counts.

\begin{table}[htbp]
\centering\small
\caption{Frozen normalization constants. Affine bounds are $L,U$; signed HPSv3 changes use scale $a$.}
\label{tab:spatial-frozen-normalization}
\begin{tabular}{lrrr}
\toprule
Metric & $L$ & $U$ or $a$ & Reference count \\
\midrule
PI (lower is better) & 2.80222 & 8.728 & 1849 \\
HPSv3 & $-4.14107$ & 9.492435 & 2394 \\
New-region HPSv3 & $-8.104675$ & 9.273475 & 136 \\
Background LPIPS (lower is better) & 0.00038 & 0.32208 & 97 \\
Old-region HPSv3 change & -- & 1.4573 & 136 \\
Connecting-path HPSv3 change & -- & 5.64125 & 136 \\
\bottomrule
\end{tabular}
\end{table}

\paragraph{Reading the scores.}
The PI, hole-free, in-bounds, and background-similarity columns are higher-is-better reporting forms of the raw metrics. An expansion area score of zero includes unchanged or reduced footprint area; a positive score indicates an increased raw area ratio. Junction coverage gain and the two HPSv3 change scores have a neutral displayed value of 50: values above 50 indicate improvement, and values below 50 indicate decline. HPSv3 preservation refers to view-quality changes, not unchanged geometry or identity. W1--W5 averages summarize different metric groups and are not directly comparable measures of task difficulty.

\section{Embodied World Model Track: Detailed Metrics}
\label{app:embodied-metrics}

This appendix gives the detailed definitions and implementation specifications
for the metrics summarized in Section~\ref{sec:embodied-metrics}. The scoring
axis contains 17 metrics: four perception metrics, five consistency metrics,
three causality metrics, and five controllability metrics. All metric scores
are normalized to $[0,100]$, with higher values indicating better performance.

\subsection{Metric Inventory and Applicability}

Table~\ref{tab:embodied-metric-levels} summarizes the evidence used by each
metric and its applicability across W2--W4.

\begin{table}[t]
    \centering\footnotesize
    \setlength{\tabcolsep}{3pt}
    \renewcommand{\arraystretch}{1.05}
    \caption{The 17 embodied metrics and their applicability across W2--W4.
    The Evidence column summarizes the information used by each metric; a check
    mark indicates that the metric contributes to the corresponding level.}
    \label{tab:embodied-metric-levels}
    \begin{tabularx}{\linewidth}{@{}>{\raggedright\arraybackslash}p{0.115\linewidth}>{\raggedright\arraybackslash}p{0.06\linewidth}>{\raggedright\arraybackslash}Xccc@{}}
        \toprule
        Dimension & Metric & Evidence & W2 & W3 & W4 \\
        \midrule
        \multirow{4}{=}{Perception}
        & SF & Persistent environment facts remain visible and recognizable. & \checkmark & \checkmark & \checkmark \\
        & SuF & Robot category, arm and gripper configuration, appearance, and viewpoint. & \checkmark & \checkmark & \checkmark \\
        & PQ & Frame-quality scores from MUSIQ-SPAQ. & \checkmark & \checkmark & \checkmark \\
        & SR & Tenengrad gradient-energy retention relative to the first frame. & \checkmark & \checkmark & \checkmark \\
        \midrule
        \multirow{5}{=}{Consistency}
        & SAC & DINOv3 and CLIP features extracted from SAM3-tracked robot-subject masks. & \checkmark & \checkmark & \checkmark \\
        & GC & Normalized Chamfer residual over static anchors with DA3 depth and pose. & \checkmark & \checkmark & \checkmark \\
        & SSC & Frozen scene and state assertions at fixed temporal checkpoints. & \checkmark & \checkmark & \checkmark \\
        & TC & Motion-compensated RGB residual under Farneback optical flow. & \checkmark & \checkmark & \checkmark \\
        & MS & AMT-S intermediate-frame prediction error. & \checkmark & \checkmark & \checkmark \\
        \midrule
        \multirow{3}{=}{Causality}
        & PP & Support, contact, collision, gravity, friction, and trajectory continuity. & \checkmark & \checkmark & \checkmark \\
        & TCO & Temporal precedence of specified causal edges or event transitions. & \checkmark & \checkmark & \checkmark \\
        & CTV & Visible state change supported by its specified trigger. & \checkmark & \checkmark & \checkmark \\
        \midrule
        \multirow{5}{=}{Controllability}
        & GSA & Robot camera viewpoint, base pose, or target-entity position, pose, or property at the requested endpoint. & \checkmark & \checkmark & \checkmark \\
        & AAC & Occurrence of the specified atomic action by the correct subject on the correct target. & \checkmark & \checkmark & \\
        & FIS & Task-specific functional effect: prescribed robot motion and stopping, or target displacement, transfer, or state response during manipulation. & \checkmark & \checkmark & \checkmark \\
        & MAO & Ordered stages appear in sequence with the required transitions. & & \checkmark & \\
        & CBF & Branch conditions, branch outcomes, and the required inter-branch difference. & & & \checkmark \\
        \bottomrule
    \end{tabularx}
\end{table}

\subsection{Perception}

\paragraph{Scene Fidelity (SF).} SF measures whether the key scene facts
established by the reference image remain visible and recognizable throughout
the generated rollout. The reference image defines the scene assertions,
including the presence, identity, color, and spatial arrangement of relevant
entities. The reference image and sampled generated frames are jointly
provided to the VLM in one call. The VLM returns one binary judgment for each
assertion, and the resulting judgments are combined using the predefined
importance weights for the applicable assertions.

\paragraph{Subject Fidelity (SuF).} SuF measures whether the robot subject
specified by the reference image remains recognizable with the required
identity, appearance, and structural attributes throughout the generated
rollout. The subject assertions cover the robot category, arm and gripper
configuration, appearance, viewpoint, and other task-relevant attributes. The
reference image and sampled generated frames are jointly provided to the VLM
in one call. The resulting judgments are combined using the predefined
importance weights for the applicable assertions.

\paragraph{Perceptual Quality (PQ).} PQ measures the frame-level visual quality
of the generated rollout. We adopt the MUSIQ image-quality model
~\cite{ke2021musiq} with its SPAQ configuration to evaluate each sampled frame.
Let $\{I_t\}_{t=1}^{T}$ denote the video frames selected under the common
temporal sampling protocol. The model produces a quality score $q_t$ for each
sampled frame, and the video-level score is computed as
\begin{equation}
    \mathrm{PQ}_i=\frac{1}{T}\sum_{t=1}^{T}q_t,
\end{equation}
with the calibrated output clipped to $[0,100]$.

\paragraph{Sharpness Retention (SR).} SR is implemented as a Tenengrad-based
variant of the sharpness-retention metric used in iWorld-Bench~\cite{xu2026worldroambenchopenworldbenchmarklonghorizon}.
For each sampled frame $I_t$, we compute the Tenengrad focus measure
~\cite{krotkov1989active}
\begin{equation}
    E_t=\operatorname{mean}\!\left(G_{x,t}^{2}+G_{y,t}^{2}\right),
\end{equation}
where $G_{x,t}$ and $G_{y,t}$ are the $3\times3$ Sobel derivatives of the
grayscale frame. Taking the first generated frame as the sharpness reference,
the per-frame retention is
\begin{equation}
    r_t=\min\!\left(1,\frac{E_t}{E_1}\right),
\end{equation}
and the video-level score is
\begin{equation}
    \mathrm{SR}_i=100\frac{1}{T}\sum_{t=1}^{T}r_t.
\end{equation}

\subsection{Consistency}

\paragraph{Subject Appearance Consistency (SAC).} SAC measures the temporal
stability of the robot subject's appearance and visual representation over the
generated rollout. SAM3~\cite{carion2025sam3segmentconcepts} tracks the fixed
subject prompt on the sampled sequence. For each frame with a valid mask,
DINOv3~\cite{simeoni2025dinov3} features $d_t$ and CLIP~\cite{radford2021clip}
features $c_t$ are extracted from the masked subject crop. Let $\mathcal{V}_i$
denote the set of sampled frames in which a valid subject mask is obtained,
and define $t_0=\min\mathcal{V}_i$ as the index of the first such frame. The
feature similarities are computed as
\begin{equation}
    \tilde d_t=\frac{1+\cos(d_t,d_{t_0})}{2},
    \qquad
    \tilde c_t=\frac{1+\cos(c_t,c_{t_0})}{2}.
\end{equation}
The DINOv3 and CLIP similarities are aggregated using a two-sided 10\%
trimmed mean, which removes the lowest and highest 10\% of valid-frame scores
before averaging. Let $\rho_i=|\mathcal{V}_i|/T$ denote the proportion of
sampled frames with a valid subject mask, where $T$ is the total number of
sampled frames used for the rollout. Because feature similarities are defined
only on $\mathcal{V}_i$, multiplying by $\rho_i$ lowers the SAC score when the
subject cannot be localized in a substantial part of the rollout. This temporal
mask coverage is distinct from SuF, which evaluates whether the robot subject's
prescribed attributes are visually recognizable and satisfy the corresponding
assertions. The final score is
\begin{equation}
    \mathrm{SAC}_i=100\rho_i\left[
    \frac{1}{2}\operatorname{TrimMean}_{0.1}
    \left(\{\tilde d_t\}_{t\in\mathcal{V}_i}\right)
    +\frac{1}{2}\operatorname{TrimMean}_{0.1}
    \left(\{\tilde c_t\}_{t\in\mathcal{V}_i}\right)
    \right].
\end{equation}

\paragraph{Geometric Consistency (GC).} GC measures the temporal stability of
static scene geometry while excluding entities whose geometry is expected to
change under the prescribed action. Static-anchor masks are generated by SAM3
from anchor prompts specified by the task configuration or derived from
predefined static scene categories. Entities affected by the prescribed action
are excluded from the anchor masks. For each sampled frame with a valid static
anchor, DA3~\cite{depthanything3} provides depth and camera pose, which are
used to back-project the masked region into a world-coordinate point cloud
$P_t$. Each point cloud is compared with the first valid point cloud using a
symmetric Chamfer distance~\cite{barrow1977parametric} normalized by the
spatial extent of the reference cloud. After removing the largest 10\% of
frame-level residuals, let $\bar e$ denote the mean of the remaining normalized
residuals. Let $\rho_g$ denote the proportion of sampled frames with valid
static-anchor point clouds. The score is
\begin{equation}
    \mathrm{GC}_i=100\exp(-\bar e/0.05)\rho_g.
\end{equation}

\paragraph{Scene--State Consistency (SSC).} SSC evaluates the persistence of
scene and object states at fixed temporal checkpoints sampled from the
normalized rollout. Predefined task-specific scene and state assertions are
evaluated by the VLM at each checkpoint, producing a binary judgment
$y_{j,k}\in\{0,1\}$ for assertion $j$ at checkpoint $k$. Scene assertions
describe persistent environmental facts, such as the presence, identity, and
spatial arrangement of relevant entities. State assertions describe the
temporal states of the robot and manipulated objects, including their
configuration, placement, and other task-relevant properties. They encode
persistent or invariant properties while allowing the changes explicitly
prescribed by the action prompt. Let $S_{\mathrm{scene}}$ and
$S_{\mathrm{state}}$ denote the corresponding weighted means over assertions
and checkpoints, each normalized to $[0,100]$. The SSC score is
\begin{equation}
    \mathrm{SSC}_i=
    \frac{1}{2}S_{\mathrm{scene}}+
    \frac{1}{2}S_{\mathrm{state}}.
\end{equation}

\paragraph{Temporal Coherence (TC).} TC measures low-level visual continuity
between adjacent video frames after motion compensation. Under the common
temporal sampling protocol, Farneback optical flow~\cite{farneback2003two} is
estimated for each adjacent pair of sampled frames and used to compensate for
the estimated image motion. The mean RGB absolute error over valid pixels after
compensation is denoted by $e_t$, and its average over all adjacent frame pairs
by $\bar e$. Using a fixed decay scale $\tau_{\mathrm{TC}}$ shared across all
models, the score is defined as
\begin{equation}
    \mathrm{TC}_i =
    100\exp\left(-\frac{\bar e}{\tau_{\mathrm{TC}}}\right).
\end{equation}
TC captures residual visual changes after motion compensation, including
inter-frame flicker and appearance changes that cannot be explained by the
estimated motion.

\paragraph{Motion Smoothness (MS).} MS uses AMT-S~\cite{li2023amt} to predict
each odd-indexed sampled frame $\hat I_{2k+1}$ from its neighboring even-indexed
frames $I_{2k}$ and $I_{2k+2}$. The normalized RGB mean absolute error between
the predicted and observed middle frames is averaged as $\bar e$. Using a fixed
decay scale $\tau_{\mathrm{MS}}$, the score is
\begin{equation}
    \mathrm{MS}_i=
    100\exp\left(-\frac{\bar e}{\tau_{\mathrm{MS}}}\right).
\end{equation}
MS quantifies local temporal continuity through the consistency between
predicted and observed intermediate frames.

\subsection{Causality}

\paragraph{Physical Plausibility (PP).} PP evaluates whether visible motion,
contact, and object interactions satisfy the physical constraints applicable to
each task. The case-specific assertions may concern support and contact
relations, collision and joint constraints, gravity, friction, material
behavior, mass and inertia, or trajectory continuity, depending on the task
condition. The VLM evaluates each assertion over the ordered video, and the
score is computed as their weighted mean.

\paragraph{Temporal and Causal Order (TCO).} TCO measures whether the visible
events linked by each task-defined cause--effect relation occur in the required
temporal order. For each relation, the VLM judges whether the specified cause
precedes the specified effect in the ordered rollout. The score is computed as
the weighted mean of these binary judgments. Typical relations include grasp
before transport, contact before object displacement, and release before final
placement.

\paragraph{Causal Trigger Validity (CTV).} CTV evaluates whether each salient
visible state change has the specified causal trigger. For each assertion, the
VLM checks whether the trigger and the corresponding response are both visible,
and whether the response can be attributed to that trigger. CTV is computed as
the weighted mean of these binary judgments. Representative relations include
object displacement triggered by contact and stable placement triggered by
release.

\subsection{Controllability}

Controllability measures whether the generated rollout realizes the action
condition specified by the task. Its assertions describe the acting subject,
target, action type, direction, temporal placement, and the requested
interaction or state outcome.

\paragraph{Goal-State Achievement (GSA).} GSA evaluates whether the requested
final state is reached at the end of the action window. Depending on the task,
endpoint assertions concern the robot's first-person camera viewpoint or base
pose, or the target entity's final position, pose, or task-relevant property.
Its score is the weighted mean of the applicable endpoint assertions.

\paragraph{Atomic Action Compliance (AAC).} AAC evaluates whether the
specified subject visibly performs the requested atomic action on the correct
target. Its assertions describe the occurrence of the action event, including
the actor, target, action type, and, when specified, its direction and temporal
placement. The score is the weighted mean of the applicable binary action
assertions.

\paragraph{Functional Interaction Success (FIS).} FIS evaluates whether the
requested action produces the required observable functional outcome. For
movement tasks, its assertions concern the prescribed path and direction,
required spatial clearance, and stopping behavior. For manipulation tasks,
they concern the target entity's required displacement, transfer, state change,
or device and environment response. The score is the weighted mean of the
applicable binary functional assertions.

\paragraph{Multi-Action Ordering (MAO).} MAO applies to multi-step tasks and
evaluates whether the prescribed atomic actions and state transitions occur in
the required order and at the expected temporal positions. Its score is the
weighted mean of the corresponding sequence assertions.

\paragraph{Condition-Branch Fidelity (CBF).} CBF applies to paired branches
that share the same initial state. For each branch, the VLM evaluates whether
the specified branch action or condition and the corresponding expected outcome
are visible, and then evaluates whether the two branch outcomes exhibit the
required difference. The resulting judgments are averaged to obtain one group
score, and group scores are averaged across W4 groups. Action-condition variants
may differ in motion direction, displacement magnitude, target, or action type.
Physical-rule variants may change gravity, friction, mass or inertia, collision
constraints, or material rigidity; for example, the same lift-and-release action
may cause an object to fall under normal gravity but move upward under vertically
upward gravity.

\subsection{Metric Scoring and Overall Aggregation}
\label{sec:embodied-aggregation}

The preceding subsections define the score of each metric. For each evaluated
level, the applicable metrics within each dimension are averaged to obtain the
perception, consistency, causality, and controllability scores. Metrics that do
not apply to a given level or task are omitted from the corresponding average.
For assertion-based metrics, judgments are combined using the predefined
importance weights, and inapplicable assertions are excluded. The overall score
is then computed as the weighted combination
\begin{equation}
    S_{\mathrm{overall}} = 0.20S_{\mathrm{perc}} +
    0.20S_{\mathrm{cons}} + 0.30S_{\mathrm{caus}} +
    0.30S_{\mathrm{ctrl}}.
\end{equation}

\label{sec:appendix}
\end{document}